\documentclass{article}

\PassOptionsToPackage{numbers}{natbib}

\usepackage[final]{neurips_2021}

\usepackage[utf8]{inputenc} %
\usepackage[T1]{fontenc}    %
\usepackage{hyperref}       %
\usepackage{url}            %
\usepackage{booktabs}       %
\usepackage{amsfonts}       %
\usepackage{nicefrac}       %
\usepackage{microtype}      %

\usepackage{amsmath, amsthm, bbm, mathtools, commath}
\usepackage{datetime}
\usepackage{verbatim}
\usepackage[font=small]{subfig}
\usepackage{colortbl}
\usepackage{arydshln} %
\usepackage[dvipsnames]{xcolor}

\newcommand{\RR}{\mathbb R}
\newcommand{\mc}[1]{\mathcal{#1}}

\newcommand{\mrm}[1]{\mathrm{#1}}

\newcommand{\mbf}{\mathbf}

\newcommand{\bestcell}{\cellcolor{blue!25}}
\newcommand{\myparagraph}[1]{\noindent \textbf{#1}}
\newcommand{\printfnsymbol}[1]{%
  \textsuperscript{\@fnsymbol{#1}}%
}
\makeatother

\newcommand{\A}{\mbf{A}}
\newcommand{\X}{\mbf{X}}
\newcommand{\W}{\mbf{W}}
\newcommand{\Y}{\mbf{Y}}
\newcommand{\hA}{\mbf{h}_\A}
\newcommand{\hX}{\mbf{h}_\X}

\newcommand{\std}[1]{\pm #1}

\title{Large Scale Learning on Non-Homophilous Graphs: New Benchmarks and Strong Simple Methods}

\author{%
  Derek Lim\thanks{Equal contribution. }\\
  Cornell University\\
  \texttt{dl772@cornell.edu} \\
  \And
  Felix Hohne$^{*}$
 \\
  Cornell University \\
  \texttt{fmh42@cornell.edu} \\
  \And 
  Xiuyu Li$^{*}$
  \\ Cornell University 
  \\ \texttt{xl289@cornell.edu}
  \AND 
  Sijia Linda Huang \\
  Cornell University \\
  \texttt{sh837@cornell.edu}
  
  \And 
  Vaishnavi Gupta 
  \\ 
  Cornell University 
  \\ \texttt{vg222@cornell.edu}
  \And 
  Omkar Bhalerao
  \\ Cornell University \\
  \texttt{opb7@cornell.edu}
  \And 
  Ser-Nam Lim 
  \\ Facebook AI 
  \\ \texttt{sernam@gmail.com}
}

\begin{document}

\maketitle

\author{%
  David S.~Hippocampus\thanks{Use footnote for providing further information
    about author (webpage, alternative address)---\emph{not} for acknowledging
    funding agencies.} \\
  Department of Computer Science\\
  Cranberry-Lemon University\\
  Pittsburgh, PA 15213 \\
  \texttt{hippo@cs.cranberry-lemon.edu} \\
}

\maketitle

\begin{abstract}
  Many widely used datasets for graph machine learning tasks have generally been homophilous, where nodes with similar labels connect to each other. Recently, new Graph Neural Networks (GNNs) have been developed that move beyond the homophily regime; however, their evaluation has often been conducted on small graphs with limited application domains. We collect and introduce diverse non-homophilous datasets from a variety of application areas that have up to 384x more nodes and 1398x more edges than prior datasets. We further show that existing scalable graph learning and graph minibatching techniques lead to performance degradation on these non-homophilous datasets, thus highlighting the need for further work on scalable non-homophilous methods. To address these concerns, we introduce LINKX --- a strong simple method that admits straightforward minibatch training and inference. Extensive experimental results with representative simple methods and GNNs across our proposed datasets show that LINKX achieves state-of-the-art performance for learning on non-homophilous graphs. Our codes and data are available at \url{https://github.com/CUAI/Non-Homophily-Large-Scale}.
\end{abstract}

\section{Introduction}

Graph learning methods generate predictions by leveraging complex inductive biases captured in the topology of the graph \cite{battaglia2018relational}. A large volume of work in this area, including graph neural networks (GNNs), exploits \textit{homophily} as a strong inductive bias, where connected nodes tend to be similar to each other in terms of labels \cite{mcpherson2001birds, altenburger2018monophily}. Such assumptions of homophily, however, do not always hold true. For example, malicious node detection, a key application of graph machine learning, is known to be non-homophilous in many settings \cite{pandit2007netprobe, chau2006detecting, gatterbauer2014semi, breuer2020friend}.

Further, while new GNNs that work better in these non-homophilous settings have been developed \cite{zhu2020beyond, nonlocal, zhu2020graph, chien2021adaptive, chen2020simple, yan2021two,  kim2021how, jin2021node, bo2021beyond, nt2020stacked}, their evaluation is limited to a few graph datasets used by \citet{pei2019geom} (collected by \cite{rozemberczki2019multi,tang2009social, mitchell1997web}) that have certain undesirable properties such as small size, narrow range of application areas, and high variance between different train/test splits \cite{zhu2020beyond}.
Consequently, method scalability has not been thoroughly studied in non-homophilous graph learning. In fact, many non-homophilous techniques frequently require more parameters and computational resources~\cite{zhu2020beyond, abu2019mixhop, chien2021adaptive}, which is neither evident nor detrimental when they are evaluated on very small datasets. Even though scalable graph learning techniques do exist, these methods generally cannot be directly applied to the non-homophilous setting, as they oftentimes assume homophily in their construction \cite{wu2019simplifying, huang2021combining, deng2020graphzoom, bojchevski2020scaling}. 

Non-homophily in graphs also degrades proven graph learning techniques that have been instrumental to strong performance in scalable graph learning.
For instance, label propagation, personalized PageRank, and low-pass graph filtering have been used for scalable graph representation learning models, but these methods all assume homophily \cite{wu2019simplifying, huang2021combining, deng2020graphzoom, bojchevski2020scaling}.
Moreover, we give empirical evidence that existing minibatching techniques in graph learning \cite{chiang2019cluster, zeng2019graphsaint} significantly degrade performance in non-homophilous settings.
In response, we develop a novel model, LINKX, that addresses these concerns; LINKX outperforms existing graph learning methods on large-scale non-homophilous datasets and admits a simple minibatching procedure that maintains strong performance.

To summarize, we demonstrate three key areas of deficiency as mentioned above, namely: (1) that there is a lack of large, high-quality datasets covering different non-homophilous applications, (2) that current graph minibatching techniques and scalable methods do not work well in non-homophilous settings, and (3) that prior non-homophilous methods are not scalable. To these ends, this paper makes the following contributions: 

\textbf{Dataset Collection and Benchmarking.} We collect a diverse series of large, non-homophilous graph datasets and define new node features and tasks for classification. These datasets are substantially larger than previous non-homophilous datasets, span wider application areas, and capture different types of complex label-topology relationships. 
With these proposed datasets, we conduct extensive experiments with 14 graph learning methods and 3 graph minibatching techniques that are broadly representative of the graph machine learning model space.

\textbf{Analyzing Scalable Methods and Minibatching.}
We analyze current graph minibatching techniques like GraphSAINT \cite{zeng2019graphsaint} in non-homophilous settings, showing that they substantially degrade performance in experiments.
Also, we show empirically that scalable methods for graph learning like SGC and C\&S \cite{wu2019simplifying, huang2021combining} do not perform well in non-homophilous settings --- even though they achieve state-of-the-art results on many homophilous graph benchmarks. Finally, we demonstrate that existing non-homophilous methods often suffer from issues with scalability and performance in large non-homophilous graphs, in large part due to a lack of study of large-scale non-homophilous graph learning.

\begin{figure}[ht]
    \centering
    \includegraphics[width=.98\textwidth]{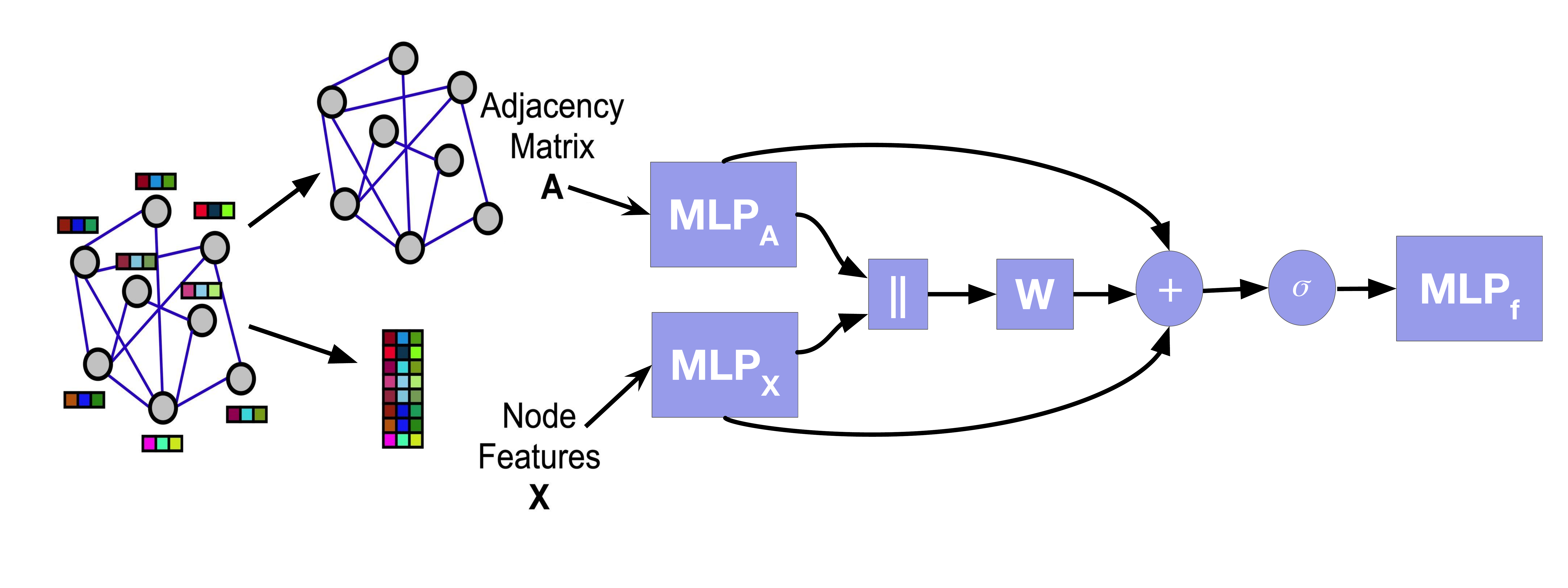}
    \caption{Our model LINKX separately embeds node features and adjacency information with MLPs, combines the embeddings together by concatenation, then uses a final MLP to generate predictions.}
    \label{fig:linkx}
    \vspace{-10pt}
\end{figure}

\textbf{LINKX: a strong, simple method.} We propose a simple method LINKX that achieves excellent results for non-homophilous graphs while overcoming the above-mentioned minibatching issues. LINKX works by separately embedding the adjacency $\A$ and node features $\X$, then combining them with multilayer perceptrons and simple transformations, as illustrated in Figure~\ref{fig:linkx}. It generalizes node feature MLP and LINK regression \cite{zheleva2009to}, two baselines that often work well on non-homophilous graphs. This method is simple to train and evaluate in a minibatched fashion, and does not face the performance degradation that other methods do in the minibatch setting. We develop the model and give more details in Section~\ref{sec:linkx}.

\section{Prior Work}

\myparagraph{Graph Representation Learning.} Graph neural networks \cite{hamilton2017inductive, kipf2017semi, velivckovic2018graph} have demonstrated their utility on a variety of graph machine learning tasks. Most GNNs are constructed by stacking layers that propagate transformed node features, which are then aggregated via different mechanisms. The neighborhood aggregation used in many existing GNNs implicitly leverage homophily, so they often fail to generalize on non-homophilous graphs \cite{zhu2020beyond, balcilar2021analyzing}. Indeed, a wide range of GNNs operate as low-pass graph filters \cite{nt2019revisiting, wu2019simplifying, balcilar2021analyzing} that smooth features over the graph topology, which produces similar representations and thus similar predictions for neighboring nodes.

\myparagraph{Scalable methods.} A variety of scalable graph learning methods have been developed for efficient computation in larger datasets \cite{zeng2019graphsaint, chiang2019cluster, ying2018graph, hamilton2017inductive, wu2019simplifying, huang2021combining, deng2020graphzoom, bojchevski2020scaling}. Many of these methods explicitly make use of an assumption of homophily in the data \cite{wu2019simplifying, huang2021combining, deng2020graphzoom, bojchevski2020scaling}.  By leveraging this assumption, several simple, inexpensive models are able to achieve state-of-the-art performance on homophilic datasets \cite{wu2019simplifying, huang2021combining}. However, these methods are unable to achieve comparable performance in non-homophilous settings, as we show empirically in Section~\ref{sec:experiments}.

\myparagraph{Graph sampling.} As node representations depend on other nodes in the graph, there are no simple minibatching techniques in graph learning as there are for i.i.d. data. To scale to large graphs, one line of work samples nodes that are used in each layer of a graph neural network \cite{hamilton2017inductive, ying2018graph, chen2018fastgcn}. Another family of methods samples subgraphs of an input graph, then passes each subgraph through a GNN to make a prediction for each node of the subgraph \cite{chiang2019cluster, zeng2019accurate, zeng2019graphsaint}. While these methods are useful for scalable graph learning, we show that they substantially degrade performance in our non-homphilous experiments (see Section~\ref{sec:experiments}).

\myparagraph{Non-Homophilous methods.} Various GNNs have been proposed to achieve higher performance in low-homophily settings \cite{zhu2020beyond, nonlocal, zhu2020graph, chien2021adaptive, chen2020simple, yan2021two,  kim2021how, jin2021node}. Geom-GCN \cite{pei2019geom} introduces a geometric aggregation scheme, MixHop \cite{abu2019mixhop} proposes a graph convolutional layer that mixes powers of the adjacency matrix, GPR-GNN \cite{chien2021adaptive} features learnable weights that can be positive and negative in feature propagation, GCNII \cite{chen2020simple} allows deep graph convolutional networks with relieved oversmoothing, which empirically performs better in non-homophilous settings, and H\textsubscript{2}GCN \cite{zhu2020beyond} shows that separation of ego and neighbor embeddings, aggregation in higher-order neighborhoods, and the combination of intermediate representations improves GNN performance in low-homophily. 

There are several recurring design decisions across these methods that appear to strengthen performance in non-homophilous settings: using higher-order neighborhoods, decoupling neighbor information from ego information, and combining graph information at different scales \cite{zhu2020beyond}. Many of these design choices require additional overhead (see Section~\ref{sec:complexity}), thus reducing their scalability.

\myparagraph{Datasets.} The widely used citation networks Cora, Citeseer, and Pubmed \cite{sen2008collective, yang2016revisiting} are highly homophilous (see Appendix~\ref{sec:appendix_measures}) \cite{zhu2020beyond}. Recently, the Open Graph Benchmark \cite{hu2020open} has provided a series of datasets and leaderboards that improve the quality of evaluation in graph representation learning; however, most of the node classification datasets tend to be homophilous, as noted in past work \cite{zhu2020beyond} and expanded upon in Appendix~\ref{sec:homophilous_stats}. A comparable set of high-quality benchmarks to evaluate non-homophilous methods does not currently exist. 

\section{Datasets for Non-Homophilous Graph Learning}

\subsection{Currently Used Datasets}

The most widely used datasets to evaluate non-homophilous graph representation learning methods were used by \citet{pei2019geom} (and collected by \cite{rozemberczki2019multi,tang2009social, mitchell1997web}); see our Table~\ref{tab:geom_gcn} for statistics. However, these datasets have fundamental issues. First, they are very small --- the Cornell, Texas, and Wisconsin datasets have between 180-250 nodes, and the largest dataset Actor has 7,600 nodes. In analogy to certain pitfalls of graph neural network evaluation on small (homophilic) datasets discussed in \cite{shchur2018pitfalls}, evaluation on the datasets of \citet{pei2019geom} is plagued by high variance across different train/test splits (see results in \cite{zhu2020beyond}). The small size of these datasets may tend to create models that are more prone to overfitting \cite{dwivedi2020benchmarking}, which prevents the scaling up of GNNs designed for non-homophilous settings. %

\citet{peel2017graph} also studies node classification on network datasets with various types of relationships between edges and labels. However, they only study methods that act on graph topology, and thus their datasets do not necessarily have node features. We take inspiration from their work, by testing on Pokec and Facebook networks with node features that we define, and by introducing other year-prediction tasks on citation networks that have node features.

\subsection{An Improved Homophily Measure}\label{sec:new_measure}

Various metrics have been proposed to measure the homophily of a graph. However, these metrics are sensitive to the number of classes and the number of nodes in each class. Let $G = (V, E)$ be a graph with $n$ nodes, none of which are isolated. Further let each node $u \in V$ have a class label $k_u \in \{0, 1, \ldots, C-1\}$ for some number of classes $C$, and denote by $C_k$ the set of nodes in class $k$. The edge homophily \cite{zhu2020beyond} is the proportion of edges that connect two nodes of the same class:
\begin{equation}\label{eq:edge_hom}
    h = \frac{| \{(u,v) \in E : k_u = k_v \} |}{|E|}.
\end{equation}
Another related measure is what we call the node homophily \cite{pei2019geom}, defined as $\frac{1}{|V|} \sum_{u \in V} \frac{d_u^{(k_u)}}{d_u}$,
in which $d_u$ is the number of neighbors of node $u$, and $d_u^{(k_u)}$ is the number of neighbors of $u$ that have the same class label. We focus on the edge homophily \eqref{eq:edge_hom} in this work, but find that node homophily tends to have similar qualitative behavior in experiments. 

The sensitivity of edge homophily to the number of classes and size of each class limits its utility.
We consider a null model for graphs in which the graph topology is independent of the labels; suppose that nodes with corresponding labels are fixed, and include edges uniformly at random in the graph that are independent of node labels.
Under this null model, a node $u \in V$ would be expected to have $d_u^{(k_u)}/d_u \approx |C_{k_u}|/n$ as the proportion of nodes of the same class that they connect to \cite{altenburger2018monophily}. For a dataset with $C$ balanced classes, we would thus expect the edge homophily to be around $\frac{1}{C}$, so the interpretation of the measure depends on the number of classes. Also, if classes are imbalanced, then the edge homophily may be misleadingly large. For instance, if 99\% of nodes were of one class, then most edges would likely be within that same class, so the edge homophily would be high, even when the graph is generated from the null model where labels are independent of graph topology. Thus, the edge homophily does not capture deviation of the label distribution from the null model.

We introduce a metric that better captures the presence or absence of homophily. Unlike the edge homophily, our metric measures excess homophily that is not expected from the above null model where edges are randomly wired. Our metric does not distinguish between different non-homophilous settings (such as heterophily or independent edges); we believe that there are too many degrees of freedom in non-homophilous settings for a single scalar quantity to be able to distinguish them all.  Our measure is given as:
\begin{equation}\label{eq:our_measure}
\hat h = \frac{1}{C-1} \sum_{k=0}^{C-1} \left[h_k - \frac{|C_k|}{n} \right]_+,
\end{equation}
where $[a]_+ = \max(a, 0)$, and $h_k$ is the class-wise homophily metric
\begin{equation}
h_k = \frac{\sum_{u \in C_k} d_u^{(k_u)}}{\sum_{u \in C_k} d_u}.
\end{equation}
Note that $\hat h \in [0,1]$, with a fully homophilous graph (in which every node is only connected to nodes of the same class) having $\hat h = 1$. Since each class-wise homophily metric $h_k$ only contributes positive deviations from the null expected proportion $|C_k|/n$, the class-imbalance problem is substantially mitigated. Also, graphs in which edges are independent of node labels are expected to have $\hat h \approx 0$, for any number of classes. Our measure $\hat h$ measures presence of homophily, but does not distinguish between the many types of possibly non-homophilous relationships. This is reasonable given the diversity of non-homophilous relationships. For example, non-homophily can imply independence of edges and classes, extreme heterophily, connections only among subsets of classes, or certain chemically / biologically determined relationships. Indeed, these relationships are very different, and are better captured by more than one scalar quantity, such as the compatibility matrices presented in the appendix. Further discussion is given in Appendix~\ref{sec:appendix_measures}.

\subsection{Proposed Datasets}

Here, we detail the non-homophilous datasets that we propose for graph machine learning evaluation.  
Our datasets and tasks span diverse application areas. \textbf{Penn94}~\cite{traud2012social},  \textbf{Pokec}~\cite{snapnets}, \textbf{genius}~\cite{lim2021expertise}, and \textbf{twitch-gamers}~\cite{rozemberczki2021twitch} are online social networks, where the task is to predict reported gender, certain account labels, or use of explicit content on user accounts. For the citation networks \textbf{arXiv-year}~\cite{hu2020open} and \textbf{snap-patents} \cite{leskovec2005graphs, snapnets} the goal is to predict year of paper publication or the year that a patent is granted. The dataset \textbf{wiki} consists of Wikipedia articles, where the goal is to predict total page views of each article. Detailed descriptions about the graph structure, node features, node labels, and licenses of each dataset are given in Appendix~\ref{sec:dataset_properties}.

 Most of these datasets have been used for evaluation of graph machine learning models in past work; we make adjustments such as modifying node labels and adding node features that allow for evaluation of GNNs in non-homophilous settings. We define node features for Pokec, genius, and snap-patents, and we also define node labels for arXiv-year, snap-patents, and genius. Additionally, we crawl and clean the large-scale wiki dataset --- a new Wikipedia dataset where the task is to predict page views, which is non-homophilous with respect to the graph of articles connected by links between articles (see Appendix~\ref{sec:wiki}). This wiki dataset has 1,925,342 nodes and 303,434,860 edges, so training and inference require scalable algorithms.

Basic dataset statistics are given in Table~\ref{tab:data_stats}. Note the substantial difference between the size of our datasets and those of \citet{pei2019geom} in Table~\ref{tab:geom_gcn}; our datasets have up to  384x more nodes and 1398x more edges. 
The homophily measures along with the lower empirical performance of homophily-assuming models (Section~\ref{sec:experiments}) and examination of compatibility matrices (Appendix~\ref{sec:appendix_measures}) show that our datasets are indeed non-homophilous. 
As there is little study in large-scale non-homophilous graph learning, our proposed large datasets strongly motivate the need for developing a new, scalable approach that can accurately learn on non-homophilous graphs.

\begin{table*}[t]
    \centering
    \caption{Statistics for previously used datasets from \citet{pei2019geom} (collected by \cite{rozemberczki2019multi, tang2009social, mitchell1997web}). \#C is the number of node classes. The highest number of nodes or edges overall are bolded.}
    {\footnotesize
    \begin{tabular}{crrrrrrr}
    \toprule
    Dataset & \# Nodes & \# Edges &  \# Feat. & \# C & Context & Edge hom. & $\hat h$ (ours) \\
    \midrule
         Chameleon & 2,277 & 36,101 & 2,325 & 5 & Wiki pages & .23 & .062\\
         Cornell & 183 & 295 & 1,703 & 5 & Web pages & .30 & .047\\
         Actor & \textbf{7,600} & 29,926 & 931 & 5 & Actors in movies  & .22 & .011\\
         Squirrel & 5,201 & \textbf{216,933} & 2,089 & 5 & Wiki pages & .22 & .025\\
         Texas  & 183 & 309 & 1,703 & 5 & Web pages & .11 & .001\\
         Wisconsin & 251 & 499 & 1,703 & 5 & Web pages & .21 & .094\\
    \bottomrule
    \end{tabular}
    }
    \vspace{-5pt}
    \label{tab:geom_gcn}
\end{table*}

\begin{table}[ht]
    \centering
    \caption{Statistics of our proposed non-homophilous graph datasets. \# C is the number of distinct node classes. Note that our datasets come from more diverse applications areas and are much larger than those shown in Table~\ref{tab:geom_gcn}, with up to 384x more nodes and 1398x more edges.}
    \label{tab:data_stats}
    {\footnotesize
    \begin{tabular}{crrrrrrrr}
    \toprule
    Dataset &  \# Nodes & \# Edges &  \# Feat. & \# C & Class types & Edge hom. & $\hat h$ (ours) \\
    \midrule
         Penn94 & 41,554 & 1,362,229  & 5 & 2 & gender & .470  & .046  \\
         pokec  & 1,632,803 & 30,622,564 & 65  &  2 & gender & .445 & .000 \\
         arXiv-year  & 169,343 & 1,166,243 & 128 & 5 & pub year & .222 & .272  \\
         snap-patents  & \textbf{2,923,922} & 13,975,788 & 269 & 5 & time granted & .073 & .100  \\
         genius  & 421,961 & 984,979 & 12 & 2 & marked act. & .618  & .080  \\
         twitch-gamers  & 168,114 & 6,797,557 & 7 & 2 & mature content & .545 & .090  \\
		 wiki  & 1,925,342 & \textbf{303,434,860} & 600 & 5 & views & .389 & .107 \\
    \bottomrule
    \end{tabular}
    }
    \vspace{0pt}
\end{table}

\section{LINKX: A New Scalable Model}\label{sec:linkx}

In this section, we introduce our novel model, LINKX, for scalable node classification in non-homophilous settings. LINKX is built out of multilayer perceptrons (MLPs) and linear transformations, thus making it simple and scalable. It also admits simple row-wise minibatching procedures that allow it to perform well on large non-homophilous graphs. 
As a result, LINKX is able to circumvent aforementioned issues of graph minibatching and non-homophilous GNNs in large-scale non-homophilous settings.

\subsection{Motivation from two simple baselines}

Here, we detail two simple baselines for node classification that we build on to develop LINKX. 

\myparagraph{MLP on node features.} A na\"ive method for node classification is to ignore the graph topology and simply train an MLP on node features. 
For the same reason that the graph topology has more complicated relationships with label distributions in non-homophilous graphs, many GNNs are not able to effectively leverage the graph topology in these settings. Thus, MLPs can actually perform comparatively well on non-homophilous graphs --- achieving higher or approximately equal performance to various GNNs~\cite{zhu2020beyond}.

\myparagraph{LINK regression on graph topology.} On the other extreme, there is LINK~\cite{zheleva2009to} --- a simple baseline that only utilizes graph topology. In particular, we consider LINK regression, which trains a logistic regression model in which each node's features are taken from a column of the adjacency matrix. Letting $\A \in \{0,1\}^{n \times n}$ be the binary adjacency matrix of the graph, and $\W \in \RR^{c \times n}$ be a learned weight matrix, LINK computes class probabilities as
\begin{equation}
    \Y = \mrm{softmax}(\W\A).
\end{equation}
Let $u \in \{1, \ldots, n\}$ be a specific node, and let $k \in \{1, \ldots, c \}$ be a specific class. Then, expanding the matrix multiplication, the log-odds of node $u$ belonging to class $k$ is given by
\begin{equation}
    (\W \A)_{ku} = \sum_{v \in \mc N(u)} \W_{kv},
\end{equation}
where $\mc N(u)$ contains the 1-hop neighbors of $u$. In other words, the logit is given by the sum of weights $\W_{kv}$ across the 1-hop neighbors of $u$. If a specific node $v$ has many neighbors of class $k$, then $\W_{kv}$ is probably large, as we would expect with a high probability that any neighbor of $v$ is of class $k$. 
In this sense, LINK is like a 2-hop method: for a given node $u$, the probability of being in a given class is related to the class memberships of $u$'s 2-hop neighbors in $\mc N(v)$ for each neighbor $v \in \mc N(u)$. Related interpretations of LINK as a method acting on 2-hop paths between nodes are given by \citet{altenburger2018monophily}.

Though it is simple and has been overlooked in the recent non-homophilous GNN literature, LINK has been found to perform well in certain node classification tasks like gender prediction in social networks~\cite{altenburger2018monophily, altenburger2018node}. A major reason why LINK does well in many settings is exactly because it acts as a 2-hop method. For example, while 1-hop neighbors are often not so informative for gender prediction in social networks due to lack of homophily, 2-hop neighbors are very informative due to so-called ``monophily,'' whereby many nodes have extreme preferences for connecting to a certain class~\cite{altenburger2018monophily}. Beyond just gender prediction, we show in Section~\ref{sec:experiments} that LINK empirically outperforms many models across the various application areas of the non-homophilous datasets we propose.

\subsection{LINKX}\label{sec:linkx_design}

We combine these two simple baselines through simple linear transformations and component-wise nonlinearities. Let $\X \in \RR^{D \times n}$ denote the matrix of node features with input dimension $D$, and let $[\mbf{h}_1 ; \mbf{h}_2]$ denote concatenation of vectors $\mbf{h}_1$ and $\mbf{h}_2$. Then our model outputs predictions $\Y$ through the following mapping:
\begin{align}
    \hA & = \mrm{MLP}_\A(\A) \in \RR^{d \times n}\\
    \hX & = \mrm{MLP}_\X(\X) \in \RR^{d \times n}\\
    \mbf{Y} & = \mrm{MLP}_f\left(\sigma\left(\W[\hA; \hX] + \hA + \hX \right)\right),
\end{align}
in which $d$ is the hidden dimension, $\W \in \RR^{d \times 2d}$ is a weight matrix, and $\sigma$ is a component-wise nonlinearity (which we take to be $\mrm{ReLU}$). We call our model LINKX, as it extends LINK with node feature information from the matrix $\X$. A diagram of LINKX is given in Figure~\ref{fig:linkx}.

First, LINKX computes hidden representations $\hA$ of the adjacency (extending LINK) and $\hX$ of the feature matrix (as in node-feature MLPs). Then it combines these hidden representations through a linear transform $\W$ of their concatenation, with skip connections that add back in $\hA$ and $\hX$ to better preserve pure adjacency or node feature information. Finally, it puts this combined representation through a non-linearity and another MLP to make a prediction.

\myparagraph{Separating then mixing adjacency and feature information.} LINKX separately embeds the adjacency $\A$ to $\hA$ and the features $\X$ into $\hX$ before mixing them for a few reasons. First, we note that this design is reminiscent of fusion architectures in multimodal networks, where data from different modalities are processed and combined in a neural network \cite{gadzicki2020early, zeng2019deep}. In our setting, we can view adjacency information and node feature information as separate modalities. Since node feature MLPs and LINK do well independently on different datasets, this allows us to preserve their individual performance if needed. Ignoring $\hX$ information is similar to just using LINK, and ignoring $\hA$ information is just using an node feature MLP. Still, to preserve the ability to just learn a similar mapping to LINK or to a node feature MLP, we find that having the additive skip connections helps to get performance at least as good as either baseline. Our initial empirical results showed that simply concatenating adjacency and node features as input to a network does worse overall empirically (see Appendix~\ref{sec:linkx_ablation}). 

There are also computational benefits to our design choices.
Embedding $\A$ is beneficial for depth as
adding more layers to the MLPs only gives an $\mc O(d^2)$ cost --- depending only on the hidden dimension $d$ --- and thus does not scale in the number of edges $|E|$ as when adding layers to message-passing GNNs. This is because the graph information in $\A$ is already compressed to hidden feature vectors after the first linear mapping of $\mrm{MLP}_\A$, and we do not need to propagate along the graph in later steps.
Moreover, this enables a sparse-dense matrix product to compute the first linear mapping of $\mrm{MLP}_\A$ on $\A$, which greatly increases efficiency as $\A$ is typically very sparse for real-world graphs. Separate embeddings are key here, as this would not be possible if we for instance concatenated $\A$ and $\X$ when $\X$ is large and dense.

\myparagraph{Simple minibatching.} Message-passing GNNs must take graph topology into account when minibatching with techniques such as neighbor sampling, subgraph sampling, or graph partitioning. However, LINKX does not require this, as it utilizes graph information solely through defining adjacency matrix columns as features. Thus, we can train LINKX with standard stochastic gradient descent variants by taking i.i.d. samples of nodes along with the corresponding columns of the adjacency and feature matrix as features.
This is much simpler than the graph minibatching procedures for message-passing GNNs, which require specific hyperparameter choices, have to avoid exponential blowup of number of neighbors per layer, and are generally more complex to implement \cite{zeng2019graphsaint}.
In Section~\ref{sec:mini_results}, we use the simple LINKX minibatching procedure for large-scale experiments that show that LINKX with this minibatching style outperforms GNNs with graph minibatching methods. This is especially important on the scale of the wiki dataset, where none of our tested methods --- other than MLP --- is capable of running on a Titan RTX GPU with 24 GB GPU RAM (see Section~\ref{sec:experiments}).

\subsection{Complexity Analysis}\label{sec:complexity}
Using the above notation, a forward pass of LINKX has a time complexity of $\mc O\left(d|E| + nd^2L\right)$, in which $d$ is the hidden dimension (which we assume to be on the same order as the input feature dimension $D$), $L$ is the number of layers, $n$ is the number of nodes, and $|E|$ is the number of edges. We require a $\mc O(d|E|)$ cost for the first linear mapping of $\A$  and a $\mc O(d^2)$ cost per layer for MLP operations on hidden features, for $L$ total layers and each of $n$ nodes.

As mentioned above, message passing GNNs have to propagate using the adjacency in each layer, so they have an $L |E|$ term in the complexity. For instance, an $L$-layer GCN~\cite{kipf2017semi} with $d$ hidden dimensions has $\mc O(d L |E| + nd^2 L)$ complexity, as it costs $\mc O(d|E|)$ to propagate features in each layer, and $\mc O(nd^2)$ to multiply by the weight matrix in each layer.

Non-homophilous methods often make modifications to standard architectures that increase computational cost, such as using higher-order neighborhoods or using additional hidden embeddings~\cite{zhu2020beyond}. For instance, the complexity of MixHop~\cite{abu2019mixhop} is $\mc O(K(dL|E| + nd^2 L))$, which has an extra factor $K$ that is the number of adjacency powers to propagate with. The complexity of GCNII~\cite{chen2020simple} is asymptotically the same as that of GCN, but in practice it requires more computations per layer due to residual connections and linear combinations, and it also often achieves best performance with a large number of layers $L$. H\textsubscript{2}GCN~\cite{zhu2020beyond} is significantly more expensive due to its usage of strict two-hop neighborhoods, which requires it to form the squared adjacency $\A^2$. This makes the memory requirements intractable even for medium sized graphs (see Section~\ref{sec:experiments}).

\section{Experiments}\label{sec:experiments}

We conduct two sets of experiments for node classification on our proposed non-homophilous datasets. One set of experiments does full batch gradient descent training for all applicable methods. This of course limits the size of each model, as the large datasets require substantial GPU memory to train on. Our other set of experiments uses minibatching methods.
As all graph-based methods run out of memory on the wiki dataset, even on 24 GB GPUs, we only include wiki results in the minibatching section.
In all settings, our LINKX model matches or outperforms other methods.

\begin{table}[ht]
    \vspace{-5pt}
    \centering
    \caption{Experimental results. Test accuracy is displayed for most datasets, while genius displays test ROC AUC. Standard deviations are over 5 train/val/test splits. The three best results per dataset are highlighted. (M) denotes some  (or all) hyperparameter settings run out of memory.  }
    \label{tab:results}
    {\tiny
    \begin{tabular}{lllllllll}
    \toprule
	 & Penn94 &  pokec &   arXiv-year & snap-patents & genius & twitch-gamers \\
    \midrule
		MLP & $73.61 \std{0.40}$  & $62.37\std{0.02}$ & $36.70\std{0.21}$ & $31.34\std{0.05}$ & $86.68 \std{0.09}$ & $60.92\std{0.07}$ \\    
     \hdashline
     L Prop 1-hop & $63.21 \std{0.39}$ & $53.09\std{0.05}$  & $43.42\std{0.17}$ & $30.28\std{0.09}$ & $66.02\std{0.16}$ & $62.77\std{0.24}$  \\ 
     L Prop 2-hop & $74.13 \std{0.46} $ & $76.76\std{0.03}$  & $46.07\std{0.15}$ & $38.61\std{0.07}$ & $67.04\std{0.20}$ &  $63.88\std{0.24}$ \\ 
     LINK  & $80.79 \std{0.49}$ & $80.54\std{0.03}$  & \bestcell $53.97\std{0.18}$ & \bestcell $60.39\std{0.07}$ & $73.56\std{0.14}$ & $64.85\std{0.21}$ \\    
     \hdashline
     SGC 1-hop & $66.79 \std{0.27} $ & $53.61\std{0.17}$  &  $32.83\std{0.13}$ & $30.31\std{0.06}$   & $82.36 \std{0.37}$  & $58.97\std{0.19}$ \\    
     SGC 2-hop & $76.09 \std{0.45}$ & $62.81\std{1.42}$  & $32.27\std{0.06}$ & $29.09\std{0.09}$ & $82.10 \std{0.14}$ & $59.94\std{0.21}$  \\    
     C\&S 1-hop & $74.28 \std{1.19}$  & $62.35\std{0.06}$  & $44.51\std{0.16}$ & $35.55\std{0.05}$ & $82.93 \std{0.15}$ & $64.86 \std{0.27}$ \\    
     C\&S 2-hop & $78.40 \std{3.12} $ &  \bestcell $81.69\std{0.09}$  & $49.78\std{0.26}$ & $49.08\std{0.04}$ & $84.94 \std{0.49}$ & \bestcell $65.02 \std{0.16}$ \\    
     \hdashline
     GCN & $82.47 \std{0.27}$ & $75.45\std{0.17}$  & $46.02\std{0.26}$ & $45.65\std{0.04}$ & $87.42 \std{0.37}$ & $62.18\std{0.26}$ \\
     GAT & $81.53 \std{0.55}$ & $71.77\std{6.18}$ (M)  & $46.05 \std{0.51}$ & $45.37\std{0.44}$ (M)  & $55.80 \std{0.87}$
 & $59.89\std{4.12}$ \\
     GCNJK & $ 81.63 \std{0.54} $ & $77.00\std{0.14}$  &  $46.28\std{0.29}$ &  $46.88\std{0.13}$  & $89.30 \std{0.19}$ & $63.45\std{0.22}$ \\
     GATJK & $80.69 \std{0.36}$ & $71.19\std{6.96}$ (M)  & $45.80 \std{0.72}$ &  $44.78\std{0.50}$ & $56.70 \std{2.07}$ & $59.98\std{2.87}$\\
     APPNP & $74.33 \std{0.38}$  & $62.58\std{0.08}$  & $38.15\std{0.26}$  & $32.19\std{0.07}$ & $85.36 \std{0.62}$ & $60.97\std{0.10}$\\
     \hdashline
     H\textsubscript{2}GCN & (M) & (M)  & $49.09\std{0.10}$ & (M) & (M) & (M) \\
     MixHop & \bestcell $83.47 \std{0.71}$ & \bestcell $81.07\std{0.16}$  & \bestcell $51.81\std{0.17}$ & \bestcell $52.16\std{0.09}$ (M) & \bestcell $90.58 \std{0.16}$ & \bestcell $65.64\std{0.27}$  \\
     GPR-GNN & $81.38 \std{0.16}$  & $78.83\std{0.05}$ &  $45.07\std{0.21}$  & $40.19\std{0.03}$   &  $90.05 \std{0.31}$ & $61.89\std{0.29}$\\
	 GCNII & \bestcell $82.92 \std{0.59}$ & $78.94 \std{0.11}$ (M)  & $47.21 \std{0.28}$ & $37.88 \std{0.69}$ (M)  & \bestcell $90.24 \std{0.09}$ & $63.39 \std{0.61}$\\
	 \midrule
	 LINKX &  \bestcell $84.71 \std{0.52}$ & \bestcell $82.04\std{0.07}$  &  \bestcell $56.00 \std{1.34}$ & \bestcell $61.95 \std{0.12}$ & \bestcell $90.77 \std{0.27}$ & \bestcell $66.06\std{0.19}$\\
    \bottomrule
    \end{tabular}
    }
    \vspace{-10pt}
\end{table}

\subsection{Experimental Setup}\label{sec:experimental_setup}

\myparagraph{Methods.} We include both methods that are graph-agnostic and node-feature-agnostic as simple baselines. The node-feature-agnostic models of two-hop label propagation \cite{peel2017graph} and LINK (logistic regression on the adjacency matrix) \cite{zheleva2009to} have been found to perform well in various non-homophilous settings, but they have often been overlooked by recent graph representation learning work. Also, we include SGC \cite{wu2019simplifying} and C\&S \cite{huang2021combining} as simple, scalable methods that perform well on homophilic datasets. We include a two-hop propagation variant of C\&S in analogy with two-step label propagation. In addition to representative general GNNs, we also include GNNs recently proposed for non-homophilous settings. The full list of methods is: \textbf{Only node features:} MLP \cite{goodfellow2016deep}.
\textbf{Only graph topology:} label propagation (standard and two-hop) \cite{zhou2004learning, peel2017graph}, LINK \cite{zheleva2009to}.
\textbf{Simple methods:} SGC \cite{wu2019simplifying}, C\&S \cite{huang2021combining} and their two-hop variants.
\textbf{General GNNs:} GCN \cite{kipf2017semi}, GAT \cite{velivckovic2018graph}, jumping knowledge networks (GCNJK, GATJK) \cite{xu2018representation}, and APPNP \cite{klicpera2019predict}.
\textbf{Non-homophilous methods:} H\textsubscript{2}GCN \cite{zhu2020beyond}, MixHop \cite{abu2019mixhop}, GPR-GNN \cite{chien2021adaptive}, GCNII \cite{chen2020simple}, and LINKX (ours).

\myparagraph{Minibatching methods.} We also evaluate GNNs with various minibatching methods. We take GCNJK~\cite{xu2018representation} and MixHop~\cite{abu2019mixhop} as our base models for evaluation, as they are representative of many GNN design choices and MixHop performs very well in full batch training. As other minibatching methods are trickier to make work with these models, we use the Cluster-GCN~\cite{chiang2019cluster} and GraphSAINT~\cite{zeng2019graphsaint} minibatching methods, which sample subgraphs. We include both the node based sampling and random walk based sampling variants of GraphSAINT. We compare these GNNs with MLP, LINK, and our LINKX, which use simple i.i.d. node minibatching.

\myparagraph{Training and evaluation.} Following other works in non-homophilous graph learning evaluation, we take a high proportion of training nodes \cite{zhu2020beyond, pei2019geom, yan2021two}; we run each method on the same five random 50/25/25 train/val/test splits for each dataset. All methods requiring gradient-based optimization are run for 500 epochs, with test performance reported for the learned parameters of highest validation performance. We use ROC-AUC as the metric for the class-imbalanced genius dataset (about 80\% of nodes are in the majority class), as it is less sensitive to class-imbalance than accuracy.  For other datasets, we use classification accuracy as the metric. Further experimental details are in Appendix \ref{sec:exp_details}.

\subsection{Full-Batch Results}

Table~\ref{tab:results} lists the results of each method across the datasets that we propose.
Our datasets reveal several important properties of non-homophilous node classification.  Firstly, the stability of performance across runs is better for our datasets than those of \citet{pei2019geom} (see \cite{zhu2020beyond} results). Secondly, as suggested by prior theory and experiments \cite{zhu2020beyond, abu2019mixhop, chien2021adaptive}, the non-homophilous GNNs usually do well --- though not necessarily on every dataset.

The core assumption of homophily in SGC and C\&S that enables them to be simple and efficient does not hold on these non-homophilous datasets, and thus the performance of these methods is typically relatively low. Still, as expected, two-hop variants generally improve upon their one-hop counter-parts in these low-homophily settings. 

One consequence of using larger datasets for benchmarks is that the tradeoff between scalability and learning performance of non-homophilous methods has become starker, with some methods facing memory issues. This tradeoff is especially important to consider in light of the fact that many scalable graph learning methods rely on implicit or explicit homophily assumptions \cite{wu2019simplifying, huang2021combining, deng2020graphzoom, bojchevski2020scaling}, and thus face issues when used in non-homophilous settings.

Finally, LINKX achieves superior performance on all datasets, taking advantage of LINK's power, while also being able to utilize node features where they provide additional information.

\subsection{Minibatching Results}\label{sec:mini_results}
\begin{table}[ht]
    \vspace{-5pt}
    \centering
	\caption{Minibatching results on our proposed datasets. $\dagger$ denotes that 10 random partitions of the graphs are used for testing GraphSAINT sampling. (T) denotes that five runs takes $\geq 48$ hours for a single hyperparameter setting. Best results up to a standard deviation are highlighted.}
	\label{tab:mini_results}
    {\tiny
    \begin{tabular}{lllllllll}
    \toprule
	 & Penn94 & pokec $\dagger$ & arXiv-year & snap-patents $\dagger$ & genius  & twitch-gamers $\dagger$ & wiki  $\dagger$ \\
    \midrule
    MLP Minibatch & 74.24$\pm$0.55  & 62.14$\pm$0.05  & 36.89$\pm$0.11 & 22.96$\pm$0.81 & 82.35$\pm$0.38 & 61.01$\pm$0.06 & 37.38$\pm$0.21 \\
    LINK Minibatch &  81.61$\pm$0.34 &  \bestcell 81.15$\pm$0.25 &  \bestcell 53.76$\pm$0.28 &  45.65$\pm$8.25 & 80.95$\pm$0.07 &  64.38$\pm$0.26 & 57.11$\pm$0.26 \\
	GCNJK-Cluster & 69.99$\pm$0.85 & 72.67$\pm$0.05 & 44.05$\pm$0.11 & 37.62$\pm$0.31 & 83.04$\pm$0.56 & 61.15$\pm$0.16 & (T) \\

	GCNJK-SAINT-Node & 72.80$\pm$0.43 & 63.68$\pm$0.06 & 44.30$\pm$0.22 & 26.97$\pm$0.10 & 80.96$\pm$0.09 & 59.50$\pm$0.35 & 44.86$\pm$0.19 \\
	GCNJK-SAINT-RW & 72.29$\pm$0.49 & 65.00$\pm$0.11 & 47.40$\pm$0.17 & 33.05$\pm$0.06 & 81.04$\pm$0.14 & 59.82$\pm$0.27 & 47.39$\pm$0.19 \\
	MixHop-Cluster & 75.79$\pm$0.44 & 76.67$\pm$0.07 & 48.41$\pm$0.31 & 46.82$\pm$0.11 & 81.12$\pm$0.10 & 62.95$\pm$0.08 & (T) \\
	MixHop-SAINT-Node & 75.61$\pm$0.55 & 66.42$\pm$0.06 & 44.84$\pm$0.18 & 27.45$\pm$0.11 & 81.06$\pm$0.08 & 59.58$\pm$0.27 & 47.39$\pm$0.18\\
	MixHop-SAINT-RW & 76.38$\pm$0.50 & 67.92$\pm$0.06 & 50.55$\pm$0.20 & 34.21$\pm$0.07 & 82.25$\pm$0.78 & 60.39$\pm$0.16 & 49.15$\pm$0.26 \\
	 \midrule
	LINKX Minibatch & \bestcell 84.50$\pm$0.65 & \bestcell 81.27$\pm$0.38 & \bestcell 53.74$\pm$0.27 & \bestcell 60.27$\pm$0.29 & \bestcell 85.81$\pm$0.10 & \bestcell 65.84$\pm$0.19 & \bestcell 59.80$\pm$0.41 \\
	 \bottomrule
    \end{tabular}
    }
    \vspace{-10pt}
\end{table}

Our experimental results for minibatched methods on our proposed datasets are in Table~\ref{tab:mini_results}. Since GraphSAINT does not partition the nodes of the graph into subgraphs that cover all nodes, we test on the full input graph for the smaller datasets and uniformly random partitions of the graph into 10 induced subgraphs for the larger datasets.

First, we note that both Cluster-GCN and GraphSAINT sampling lead to performance degradation for these methods on our proposed non-homophilous datasets. When compared to the full-batch training results of the previous section, classification accuracy is typically substantially lower. Further experiments in Appendix~\ref{sec:appendix_minibatch_exp} give evidence that the performance degradation is often more substantial in non-homophilous settings, and provides possible explanations for why this may be the case.

On the other hand, LINKX does not suffer much performance degradation with the simple i.i.d. node minibatching technique. In fact, it matches or outperforms all methods in this setting, often by a wide margin. Though LINK performs on par with LINKX in arXiv-year and pokec, our LINKX model significantly outperforms it on other datasets, again due to LINKX's ability to integrate node feature information. We again stress that the LINKX minibatching is very simple to implement, yet it still substantially outperforms other methods. Consequently, LINKX is generally well-suited for scalable node classification across a broad range of non-homophilous settings, surpassing even specially designed non-homophilous GNNs with current graph minibatching techniques.

\section{Discussion and Conclusion}\label{sec:conclusion}

In this paper, we propose new, high-quality non-homophilous graph learning datasets, and we benchmark simple baselines and representative graph representation learning methods across these datasets. Further, we develop LINKX: a strong, simple, and scalable method for non-homophilous classification. Our experiments show that LINKX significantly outperforms other methods on our proposed datasets, thus providing one powerful method in the underexplored area of scalable learning on non-homophilous graphs.
We hope that our contributions will provide researchers with new avenues of research in learning on non-homophilous graphs, along with better tools to test models and evaluate utility of new techniques.

While we do find utility in our proposed datasets and LINKX model, this work is somewhat limited by only focusing on transductive node classification. This setting is the most natural for studying performance in the absence of homophily, since here we define homophily in terms of the node labels, and previous non-homophilous GNN work using the \citet{pei2019geom} data also studies this setting exclusively \cite{zhu2020beyond, chien2021adaptive}. Using other Facebook 100 datasets besides Penn94 \cite{traud2012social} would allow for inductive node classification, but LINKX does not directly generalize to this setting. Our proposed datasets and model LINKX could be used for link prediction, but this is left for future work.

\myparagraph{Broader Impact.} 
Fundamental research in graph learning on non-homophilous graphs has the potential for positive societal benefit. As a major application, it enables malicious node detection techniques in social networks and transaction networks that are not fooled by fraudsters’ connections to legitimate users and customers. This is a widely studied task, and past works have noted that non-homophilous structures are present in many such networks \cite{breuer2020friend, gatterbauer2014semi, pandit2007netprobe}. We hope that this paper provides insight on the homophily limitations of existing scalable graph learning models and help researchers design scalable models that continue to work well in the non-homophilous regime, thus improving the quality of node classification on graphs more broadly. As our proposed datasets have diverse structures and our model performs well across all of these datasets, the potential for future application of our work to important non-homophilous tasks is high.

Nevertheless, our work could also have potential for different types of negative social consequences. Nefarious behavior by key actors could be one source of such consequences. 
Nonetheless, we expect that the actors that can make use of large-scale social networks for gender prediction as studied in our work are limited in number. Actors with both the capability and incentive to perform such operations probably mostly consist of entities with access to large social network data such as social media companies or government actors with auxiliary networks \cite{Narayanan2009}. 
Smaller actors can perform certain attacks, but this may be made more difficult by resource requirements such as the need for certain external information \cite{Narayanan2009} or the ability to add nodes and edges before an anonymized version of a social network is released \cite{Backstrom2007WhereforeAT}. Furthermore, additional actors could make use of deanonymization attacks \cite{Hay2008Resisting, Narayanan2008, Narayanan2009} to reveal user identities in supposedly anonymized datasets.

Also, accidental consequences and implicit biases are a potential issue, even if the applications of the learning algorithms are benign and intended to benefit society \cite{mehrabi2019survey}. Performance of algorithms may vary substantially between intersectional subgroups of subjects — as in the case of vision-based gender predictors \cite{buolamwini18gender} (and some have questioned the propriety of vision-based gender classifiers altogether). Thus, there may be disparate effects on different populations, so care should be taken to understand the impact of those differences across subgroups. Moreover, large datasets require computing resources, so projects can only be pursued by large entities at the possible expense of the individual and smaller research groups \cite{birhane2021values}. This is alleviated by the fact that our experiments are each run on one GPU, and hence have significantly less GPU computing requirements than much current deep learning research. Thus, smaller research groups and independent researchers should find our work beneficial, and should be able to build on it.

Finally, the nature of collection of online user information also comes with notable ethical concerns. Common notice-and-consent policies are often ineffective in actually protecting user privacy \cite{Nissenbaum2011ACA}. Indeed, users may not actually have much choice in using certain platforms or sharing data due to social or economic reasons. Also, users are generally unable to fully read and understand all of the different privacy policies that they come across, and may not understand the implications of having their data available for long periods of time to entities with powerful inference algorithms. Furthermore, people may rely on obscurity for privacy \cite{CaseOnlineObscurity}, but this assumption may be ignored in courts of law, and it may be directly broken when data leaks or is released in aggregated form without sufficient privacy protections. Overall, while we believe that our work will benefit machine learning research and enable positive applications, we must still be aware of possible negative consequences.

\subsubsection*{Acknowledgements}

We thank Abhay Singh, Austin Benson, and Horace He for insightful discussions.
We also thank the rest of Cornell University Artificial Intelligence
for their support and discussion. We thank Facebook AI for funding equipment that made this work possible.

\bibliography{refs}
\bibliographystyle{plainnat}

\section*{Checklist}

\begin{enumerate}

\item For all authors...
\begin{enumerate}
  \item Do the main claims made in the abstract and introduction accurately reflect the paper's contributions and scope?
    \answerYes{We have supported the claims in the main paper and appendix.}
  \item Did you describe the limitations of your work?
    \answerYes{See discussion / conclusion Section~\ref{sec:conclusion}.}
  \item Did you discuss any potential negative societal impacts of your work?
    \answerYes{See discussion / conclusion Section~\ref{sec:conclusion}.}
  \item Have you read the ethics review guidelines and ensured that your paper conforms to them?
    \answerYes{}
\end{enumerate}

\item If you are including theoretical results...
\begin{enumerate}
  \item Did you state the full set of assumptions of all theoretical results?
    \answerNA{}
	\item Did you include complete proofs of all theoretical results?
    \answerNA{}
\end{enumerate}

\item If you ran experiments...
\begin{enumerate}
  \item Did you include the code, data, and instructions needed to reproduce the main experimental results (either in the supplemental material or as a URL)?
    \answerYes{We include our proposed datasets and our codes for reproducing the experimental results in the supplemental material.}
  \item Did you specify all the training details (e.g., data splits, hyperparameters, how they were chosen)?
    \answerYes{See Section~\ref{sec:experimental_setup} and Appendix Section~\ref{sec:exp_details}.}
	\item Did you report error bars (e.g., with respect to the random seed after running experiments multiple times)?
    \answerYes{We include standard deviation of performance metrics across multiple runs in Section~\ref{sec:experiments}.}
	\item Did you include the total amount of compute and the type of resources used (e.g., type of GPUs, internal cluster, or cloud provider)?
    \answerYes{We provide GPU information in Appendix Section~\ref{sec:exp_details}.}
\end{enumerate}

\item If you are using existing assets (e.g., code, data, models) or curating/releasing new assets...
\begin{enumerate}
  \item If your work uses existing assets, did you cite the creators?
    \answerYes{We cite the creators of existing datasets and methods in the main paper. Also, in Appendix~\ref{sec:exp_details}, we cite codes that we used.}
  \item Did you mention the license of the assets?
    \answerYes{In Appendix~\ref{sec:further_dataset}, we note the licenses of datasets, if provided in published work. Also, we include the license of open-source codes we build off of in Appendix~\ref{sec:exp_details}. }
  \item Did you include any new assets either in the supplemental material or as a URL?
    \answerYes{We include our proposed datasets along with codes for reproducing our results in the supplemental material.}
  \item Did you discuss whether and how consent was obtained from people whose data you're using/curating?
    \answerYes{We discuss this in Appendix~\ref{sec:further_dataset}. Most of the datasets were constructed and presented in previously published academic work. This mostly does not apply to the wiki dataset that we collect, as it is encyclopedic in nature.}
  \item Did you discuss whether the data you are using/curating contains personally identifiable information or offensive content?
    \answerYes{In Appendix~\ref{sec:further_dataset}, we discuss this. Our datasets are primarily numerical, so they do not contain offensive content. To the best of our knowledge, our datasets do not contain personally identifiable content. }
\end{enumerate}

\item If you used crowdsourcing or conducted research with human subjects...
\begin{enumerate}
  \item Did you include the full text of instructions given to participants and screenshots, if applicable?
    \answerNA{}
  \item Did you describe any potential participant risks, with links to Institutional Review Board (IRB) approvals, if applicable?
    \answerNA{}
  \item Did you include the estimated hourly wage paid to participants and the total amount spent on participant compensation?
    \answerNA{}
\end{enumerate}

\end{enumerate}

\clearpage

\appendix

\section{Compatibility Matrices and Statistics}\label{sec:appendix_measures}

\subsection{Measuring Homophily}\label{sec:measure}

\begin{figure}
    \centering
    \begin{tabular}{cccc}
    \subfloat[$h=1$, \  $\hat h = 1$]{\includegraphics[width=.21\textwidth]{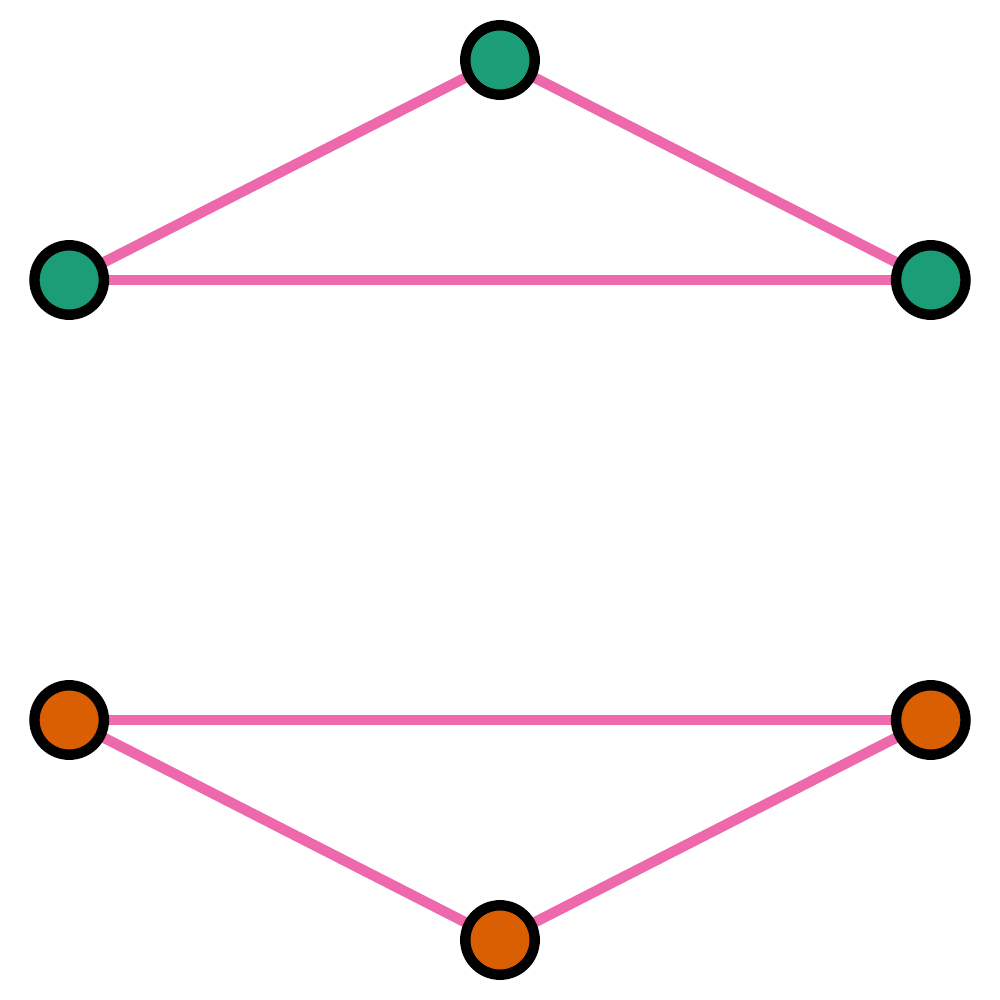}} & 
    \subfloat[$h=0$, \ $\hat h = 0$]{\includegraphics[width=.21\textwidth]{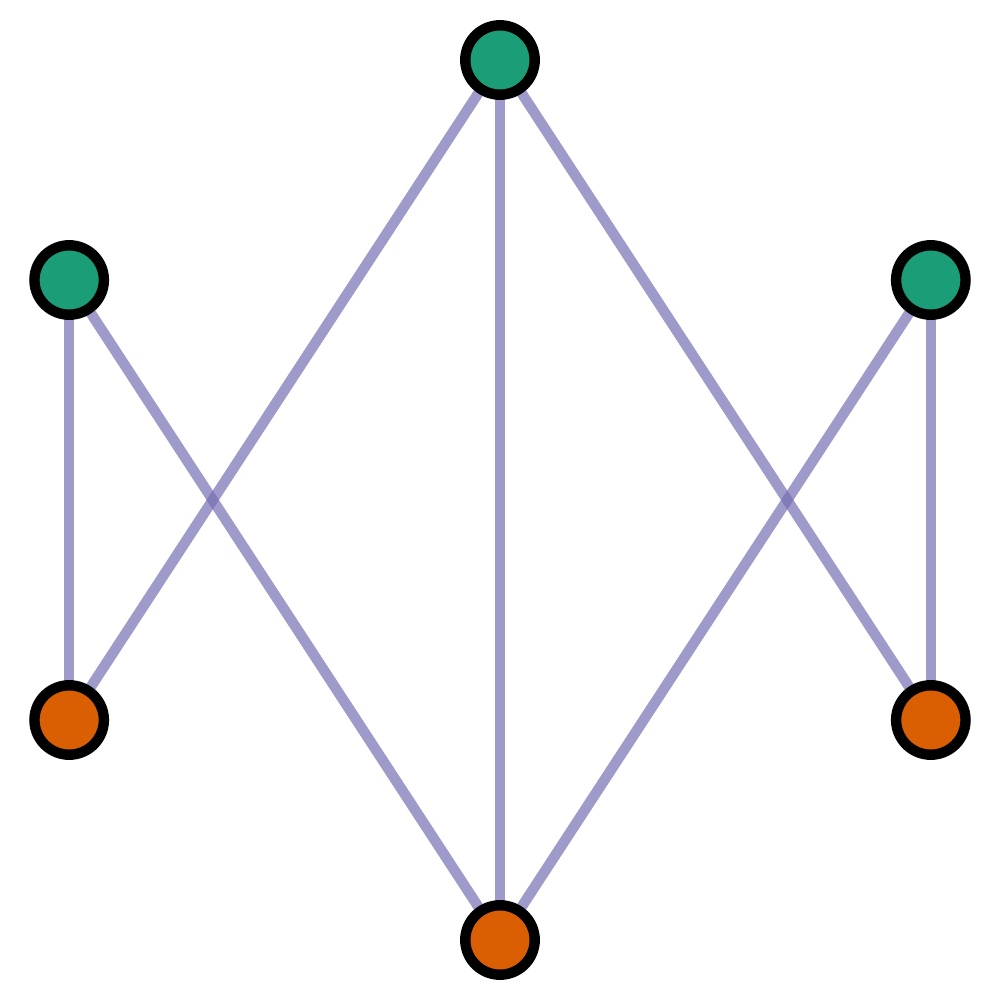}} &
    \subfloat[$h=.5$, \ $\hat h = 0$]{\includegraphics[width=.21\textwidth]{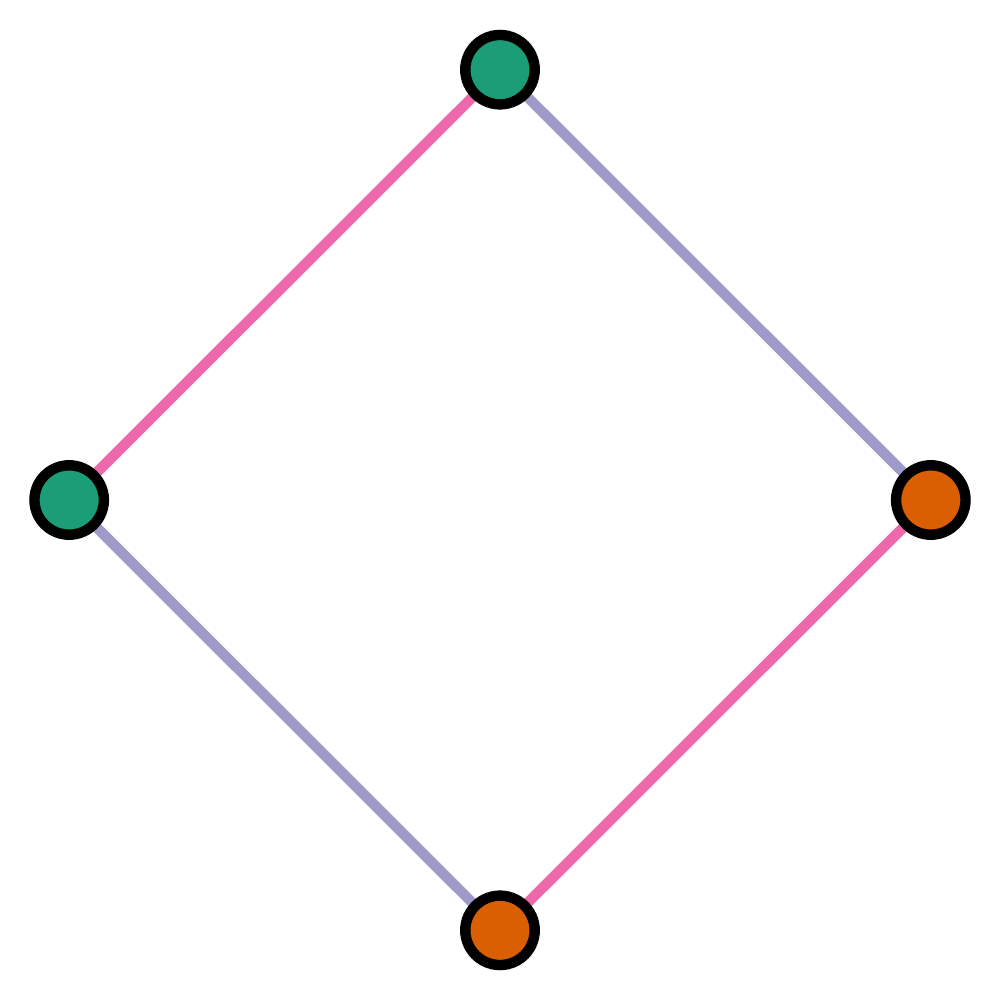}} & 
    \subfloat[$h=.33$, \ $\hat h = 0$]{\includegraphics[width=.21\textwidth]{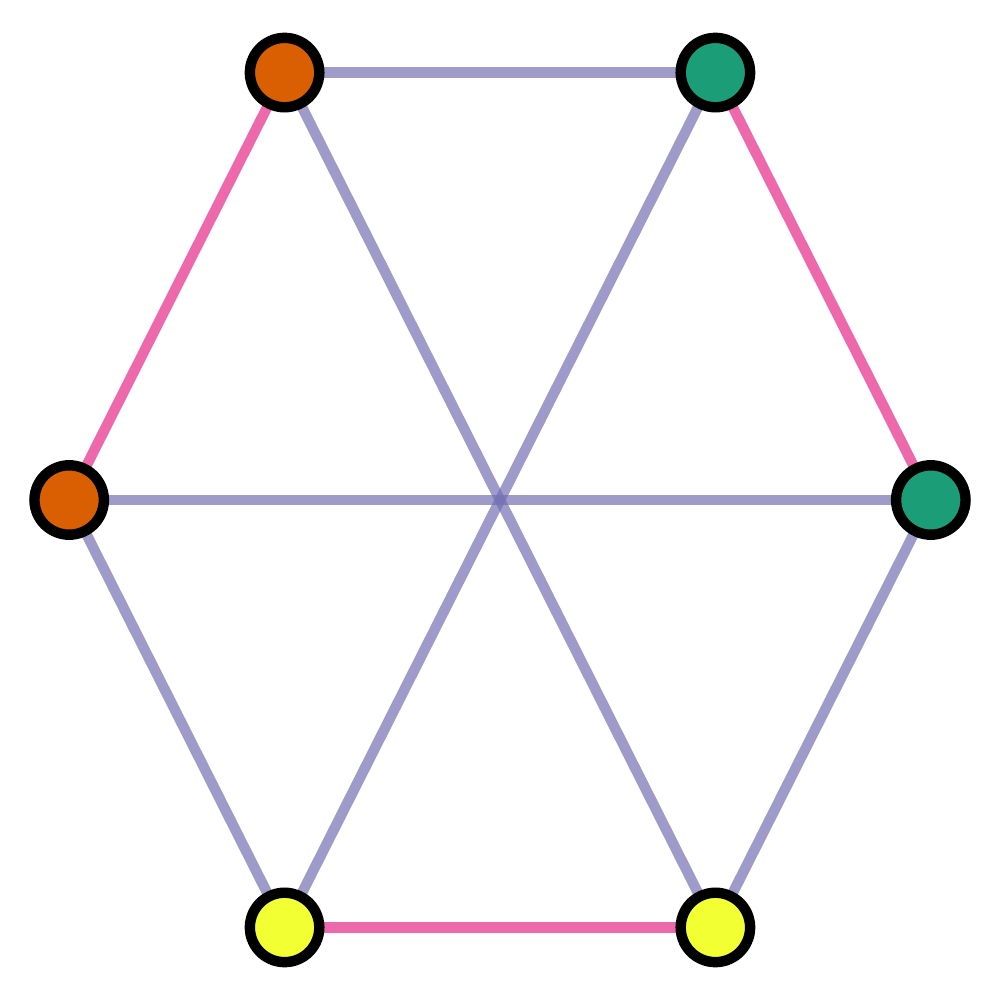}} \\
    & 
    \subfloat[$h=.53$, \ $\hat h = .07$]{\includegraphics[width=.21\textwidth]{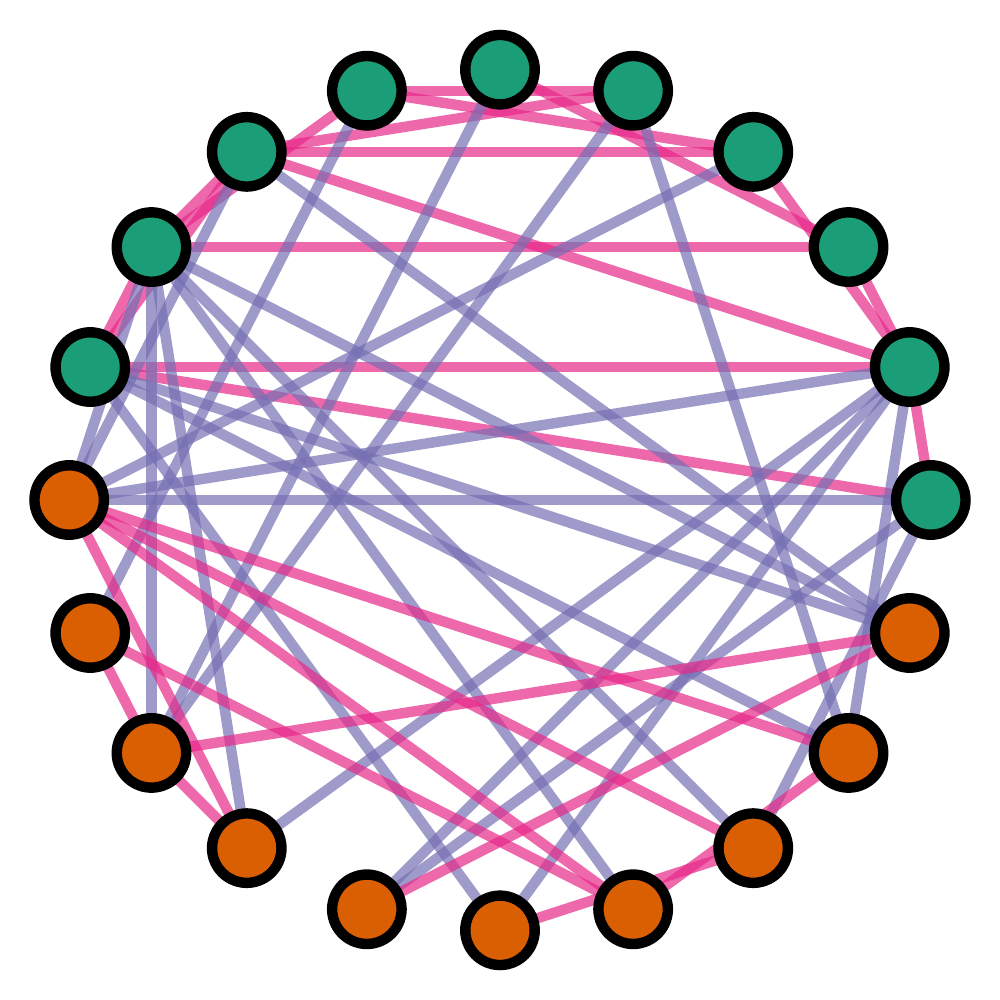}} &
    \subfloat[$h=.66$, \ $\hat h = .04$]{\includegraphics[width=.21\textwidth]{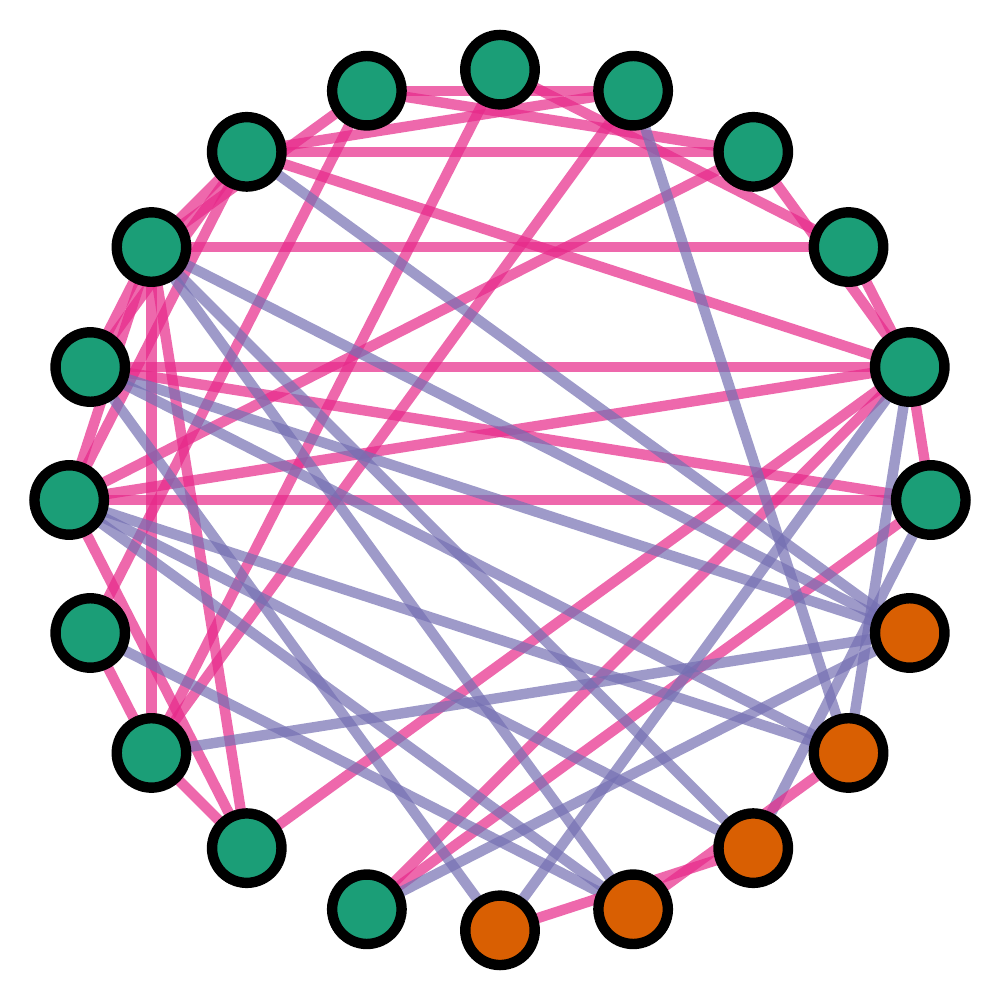}} 
    \end{tabular}
    \caption{Examples of graphs with different label-topology relationships and comparison of our measure $\hat h$ with the edge homophily ratio $h$. The node classes are labeled by color.  \textcolor[HTML]{e7298a}{Pink} edges link nodes of the same class, while \textcolor[HTML]{7570b3}{purple} edges link nodes of different classes. (a,b) Pure homophily and pure heterophily. Both measures equal $1$ in homophily and $0$ in heterophily.  (c,d) Graphs where each node is connected to one member of each class. Edge homophily depends on the number of classes, while our measure $\hat h$ does not. (e,f) Random Erd\H{o}s-R\'enyi graphs in which edges are independent of labels. Edge homophily is sensitive to class imbalance, while our measure $\hat h$ is not.}
    \label{fig:example_graphs}
\end{figure}

In Section~\ref{sec:new_measure}, we consider metrics that attempt to capture the level of class-label homophily in a graph in a single scalar quantity.
These metrics could be useful for practitioners who need to choose appropriate graph learning algorithms for some graph data that they have --- performance of algorithms heavily depends on the homophily of the graph.
Moreover, they are useful for choosing datasets to benchmark on, as we do in this paper.

A better representation of homophily may be given by the compatibility matrix, which consists of $C^2$ values instead of a single scalar value. Following previous work \cite{zhu2020beyond}, for a graph $G$ with $C$ node classes we define the $C \times C$ compatibility matrix $\mbf H$ by
\begin{equation}
    \mbf H_{kl} = \frac{|(u,v) \in E : k_u =k, \; k_v = l| }{|(u,v) \in E : k_u = k| }.
\end{equation}

This captures finer details of label-topology relationships in graphs than single scalar metrics capture. For classes $k$ and $l$, the entry $\mbf H_{kl}$ measures the proportion of edges from nodes of class $k$ that are connected to nodes of class $l$. A homophilous graph has high values of $\mbf H_{kk}$ for each class $k$.

\begin{figure*}
\centering
\begin{tabular}{ccc}
    \includegraphics[width=.3\textwidth]{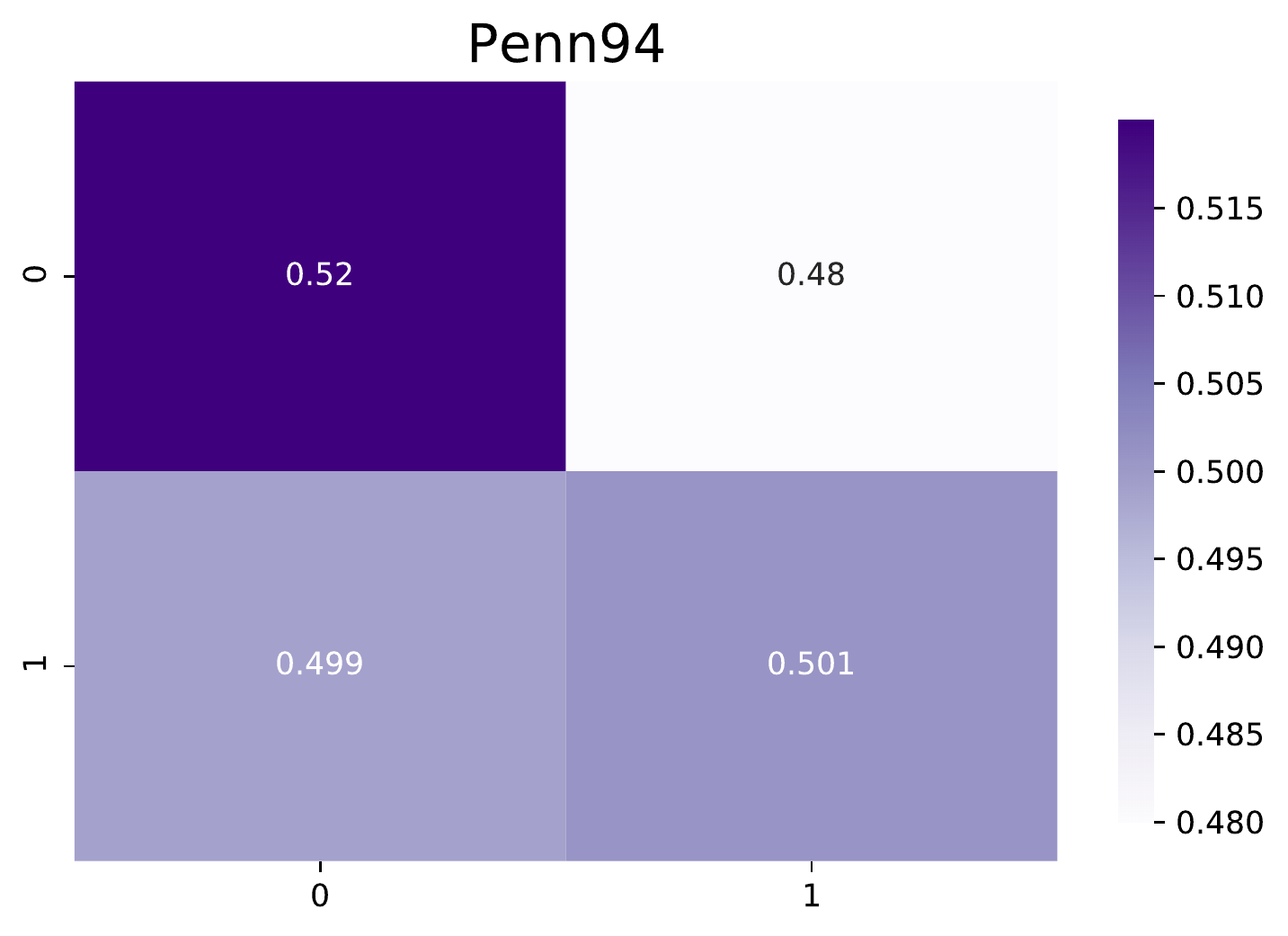} & 
    \includegraphics[width=.3\textwidth]{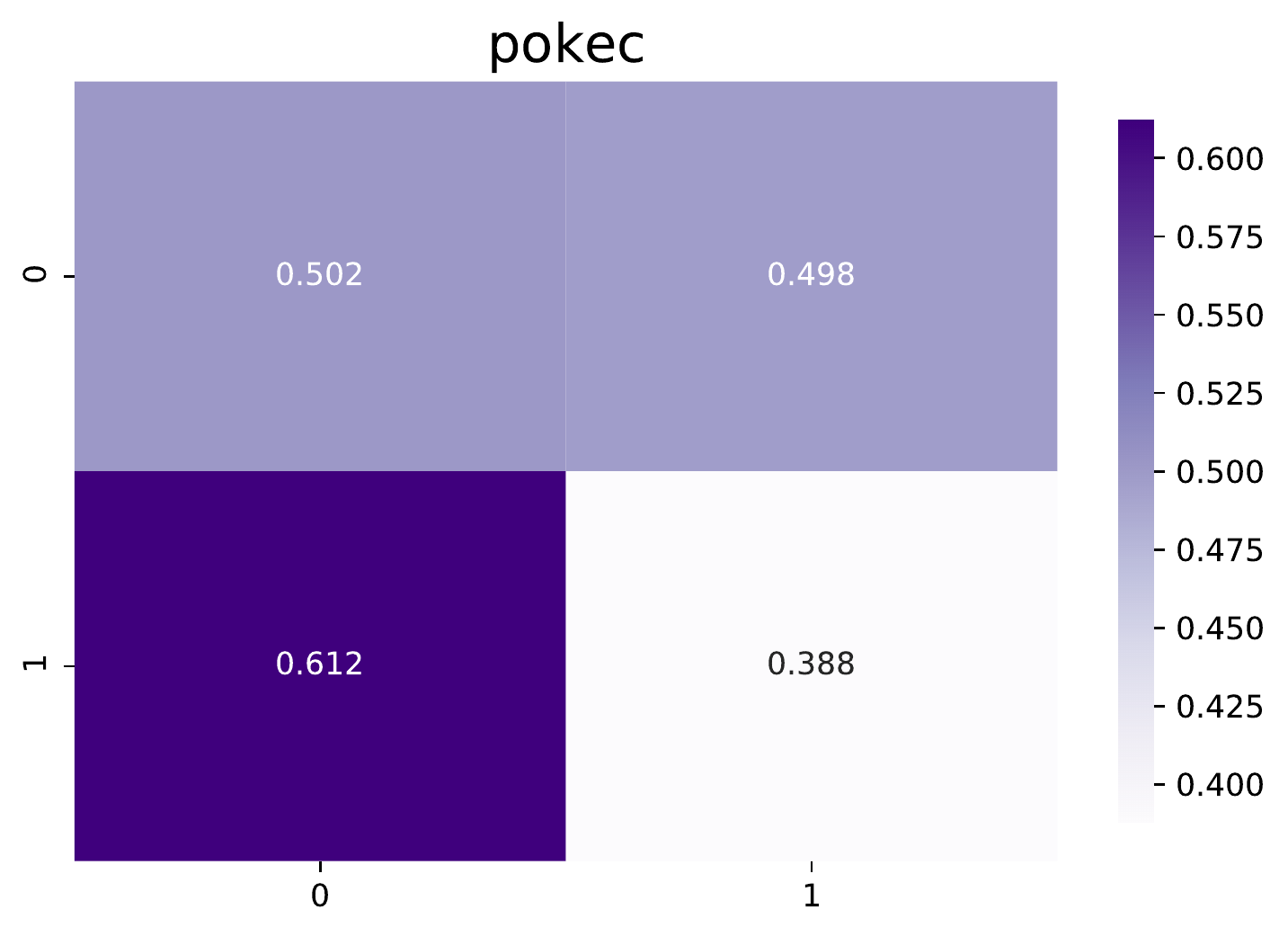} &
    \includegraphics[width=.3\textwidth]{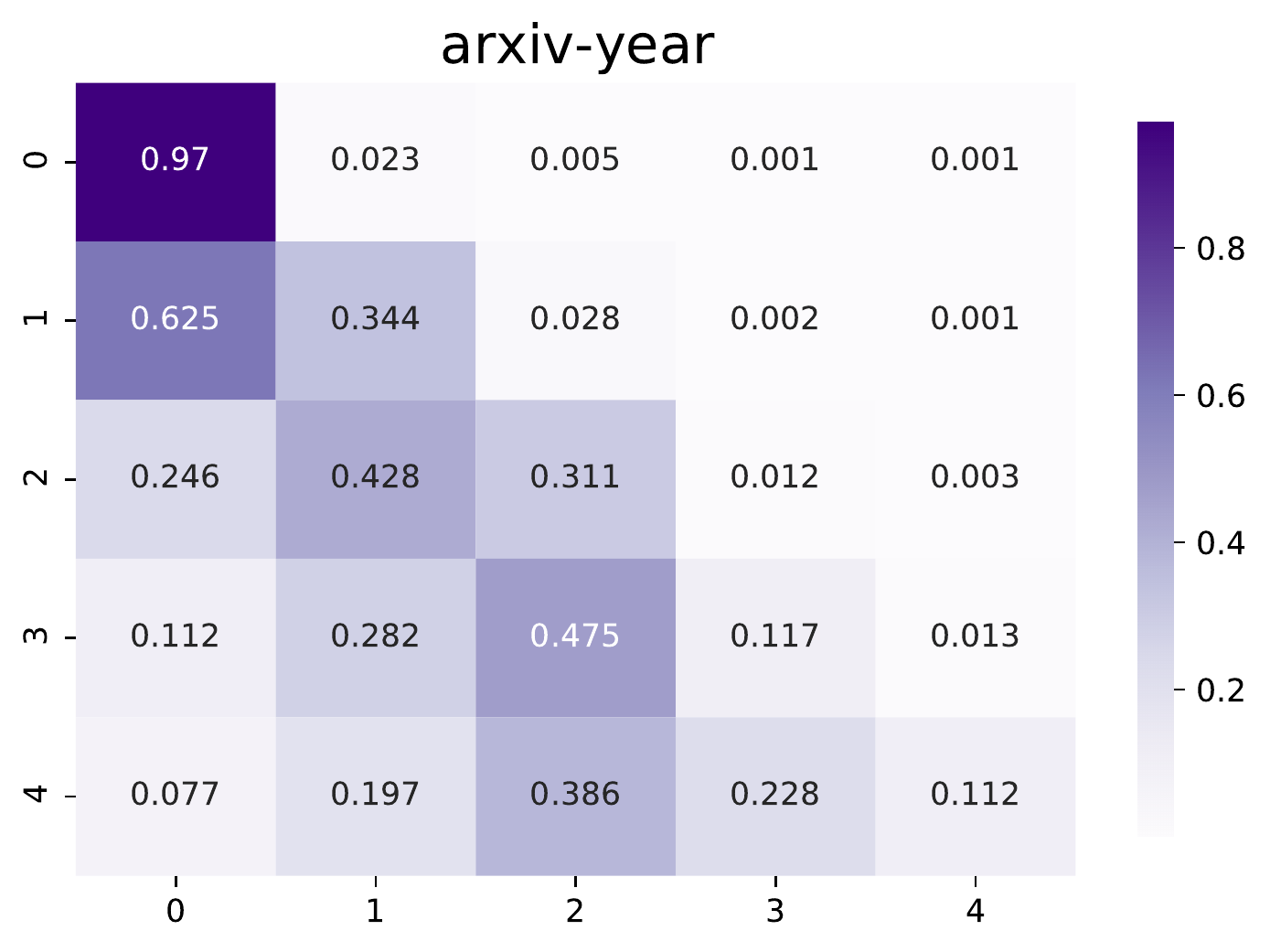} \\
    \includegraphics[width=.3\textwidth]{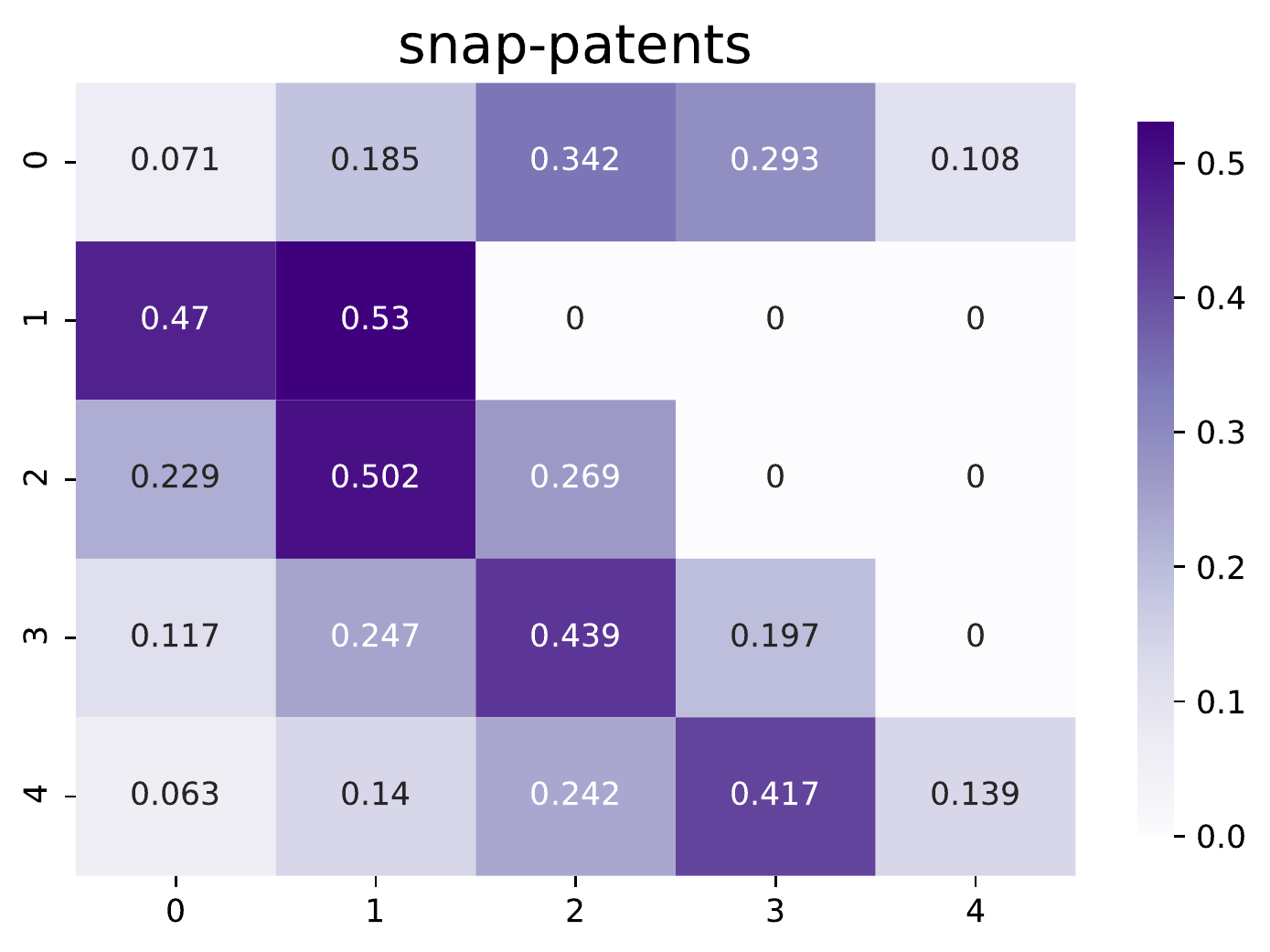} &
    \includegraphics[width=.3\textwidth]{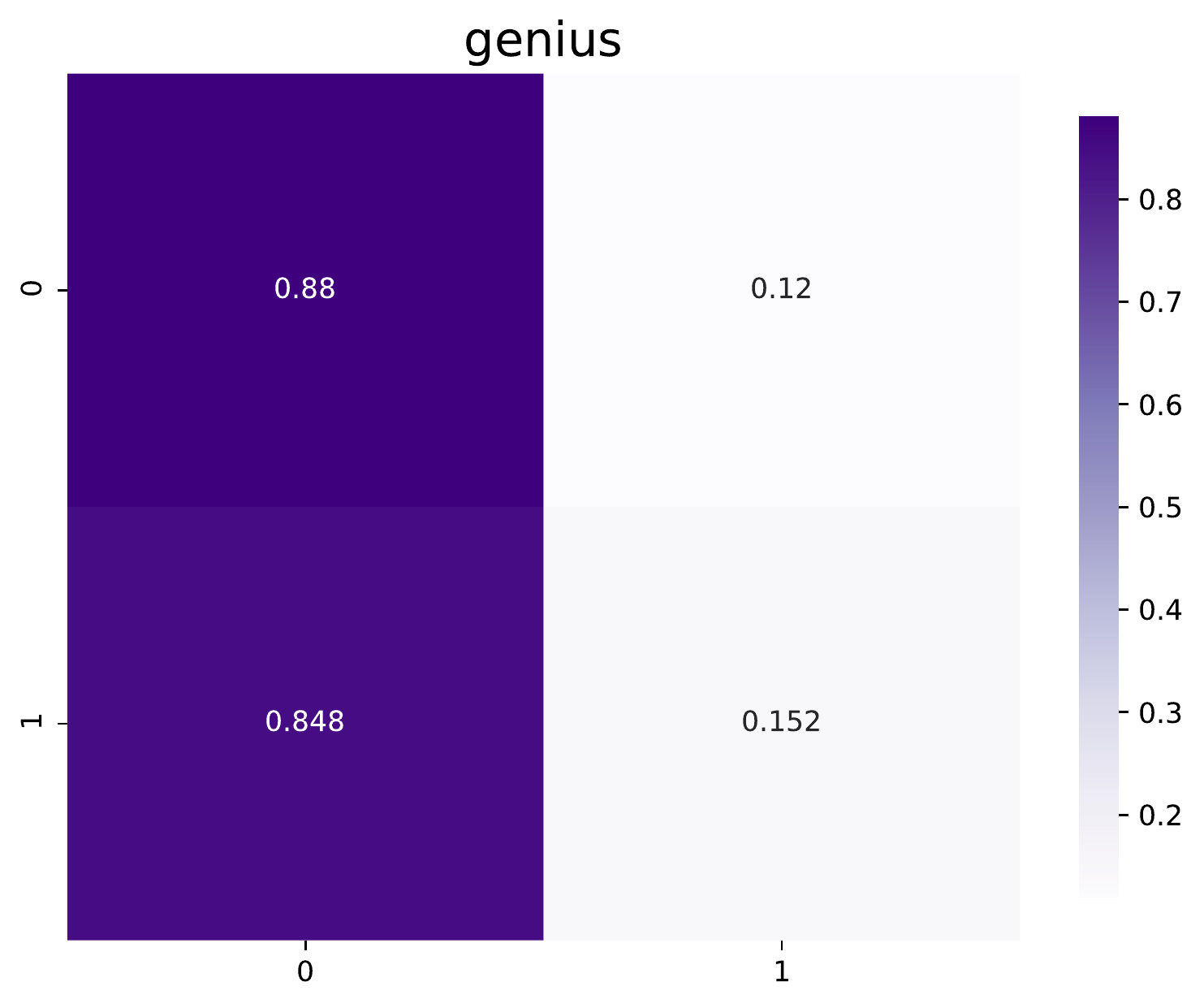} &
    \includegraphics[width=.3\textwidth]{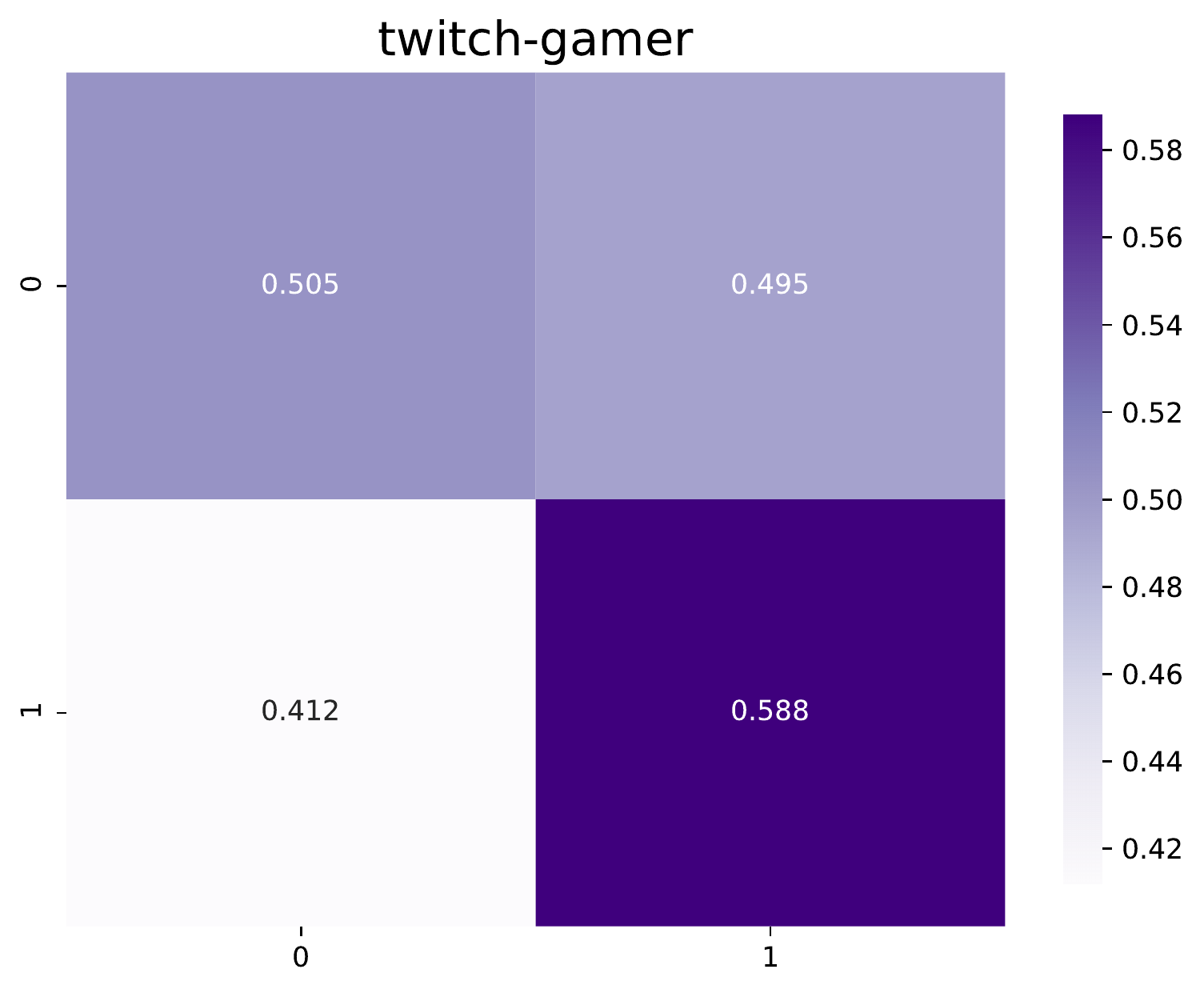} \\
    & \includegraphics[width=.3\textwidth]{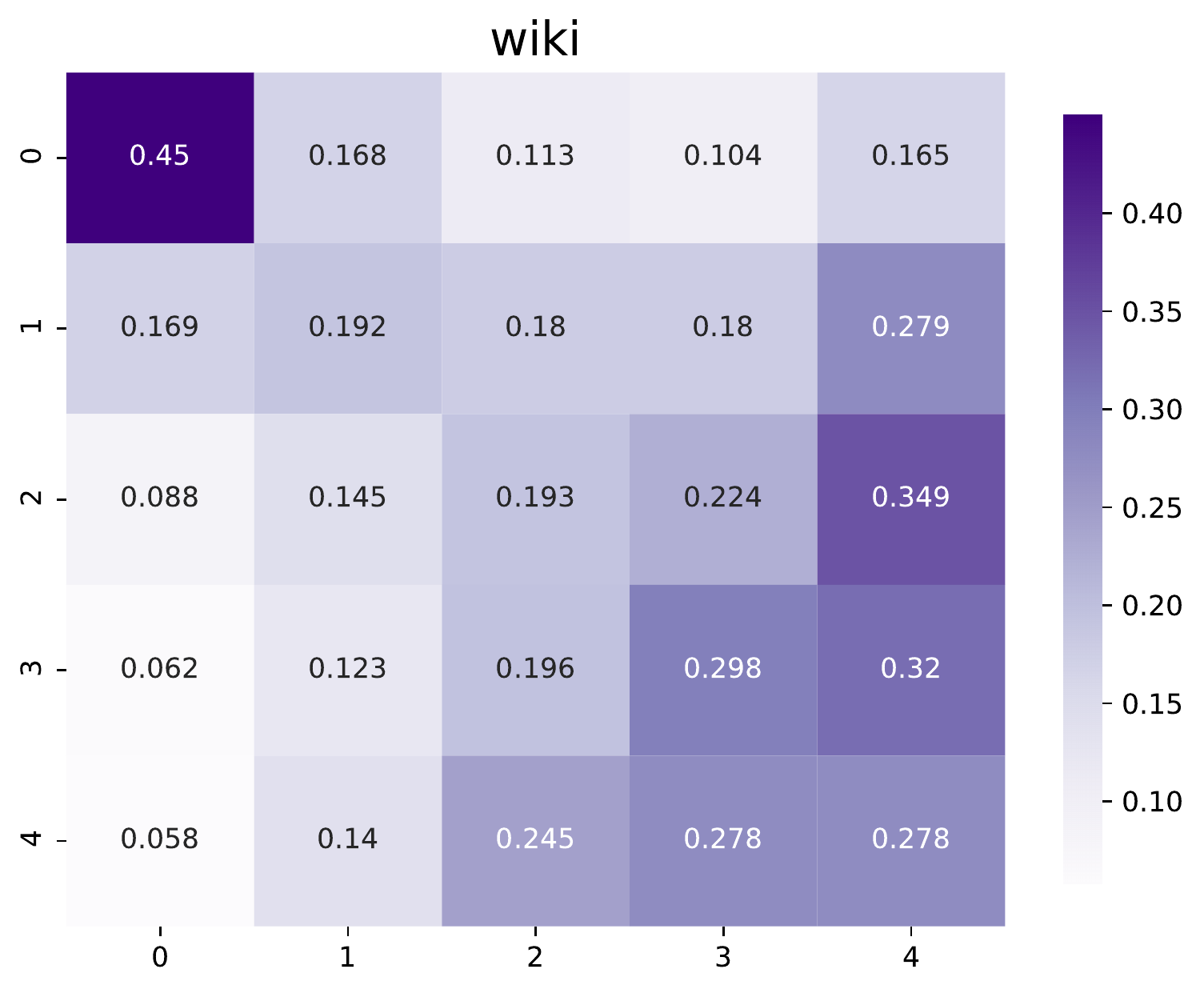}
\end{tabular}
\caption{Compatibility matrices of our proposed datasets. These datasets from a variety of different contexts exhibit a wide range of non-homophilous structures.}
\label{fig:compat_ours}
\end{figure*}

Compatibility matrices for our proposed datasets are shown in Figure \ref{fig:compat_ours}. As evidenced by the different patterns, the proposed datasets show interesting types of label-topology relationships besides homophily. For instance, the citation datasets arXiv-year and snap-patents have primarily lower-triangular structure, since most citations reference past work. In wiki, low view articles tend to link to each other, while mid-rank articles frequently link to highly viewed articles and vice versa. In Pokec, there is some heterophily, in that one gender has some preference for friends of another gender.

However, the compatibility matrix can be unwieldy for datasets with many classes. For the cases where a single scalar representation of homophily is desired, our introduced measure better captures the presence or absence of homophily than existing metrics. Figure~\ref{fig:example_graphs} compares our measure $\hat h$ to edge homophily on example graphs.

\subsection{Homophilous Data Statistics}\label{sec:homophilous_stats}

\begin{table}[ht]
    \centering
    \caption{Statistics for homophilic graph datasets. \# C is the number of node classes.}
    \label{tab:homophilic_stats}
    {\footnotesize
    \begin{tabular}{crrrrrr}
    \toprule
    Dataset & \# Nodes & \# Edges &   \# C &  Edge hom. & $\hat h$ (ours) \\
    \midrule
         Cora & 2,708 & 5,278 &  7 &  .81 & .766\\
         Citeseer & 3,327 & 4,552 &  6 &  .74 & .627\\
         Pubmed & 19,717 & 44,324 &  3 &  .80 & .664\\
         ogbn-arXiv & 169,343 & 1,166,243 &  40 &  .66 & .416\\
         ogbn-products & 2,449,029 & 61,859,140 &  47 &  .81 & .459 \\
         oeis & 226,282 & 761,687  & 5 &  .50 & .532 \\
    \bottomrule
    \end{tabular}
    }
\end{table}

\begin{figure*}[t]
\centering
\begin{tabular}{ccc}
    \includegraphics[width=.27\textwidth]{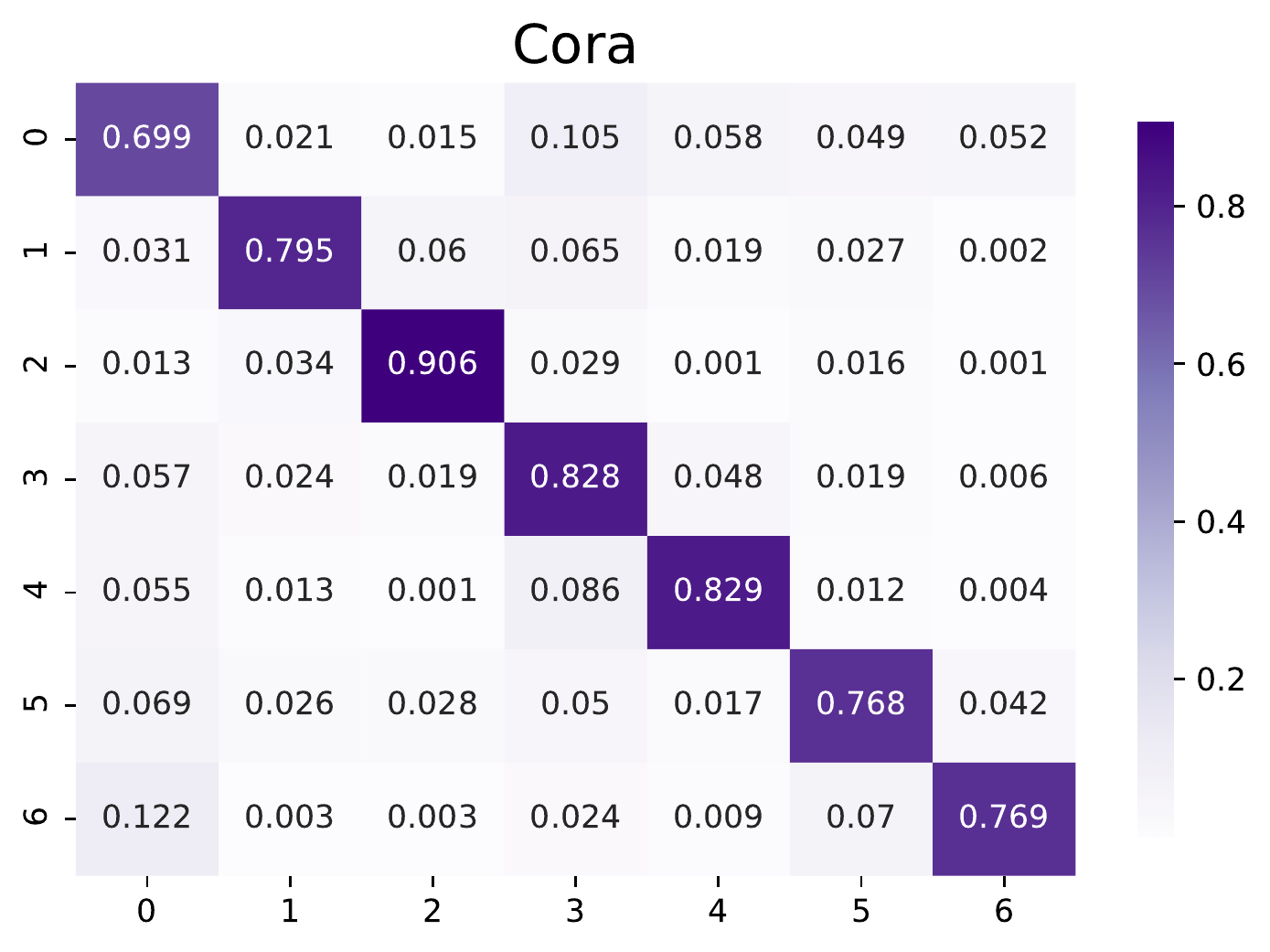} &
    \includegraphics[width=.27\textwidth]{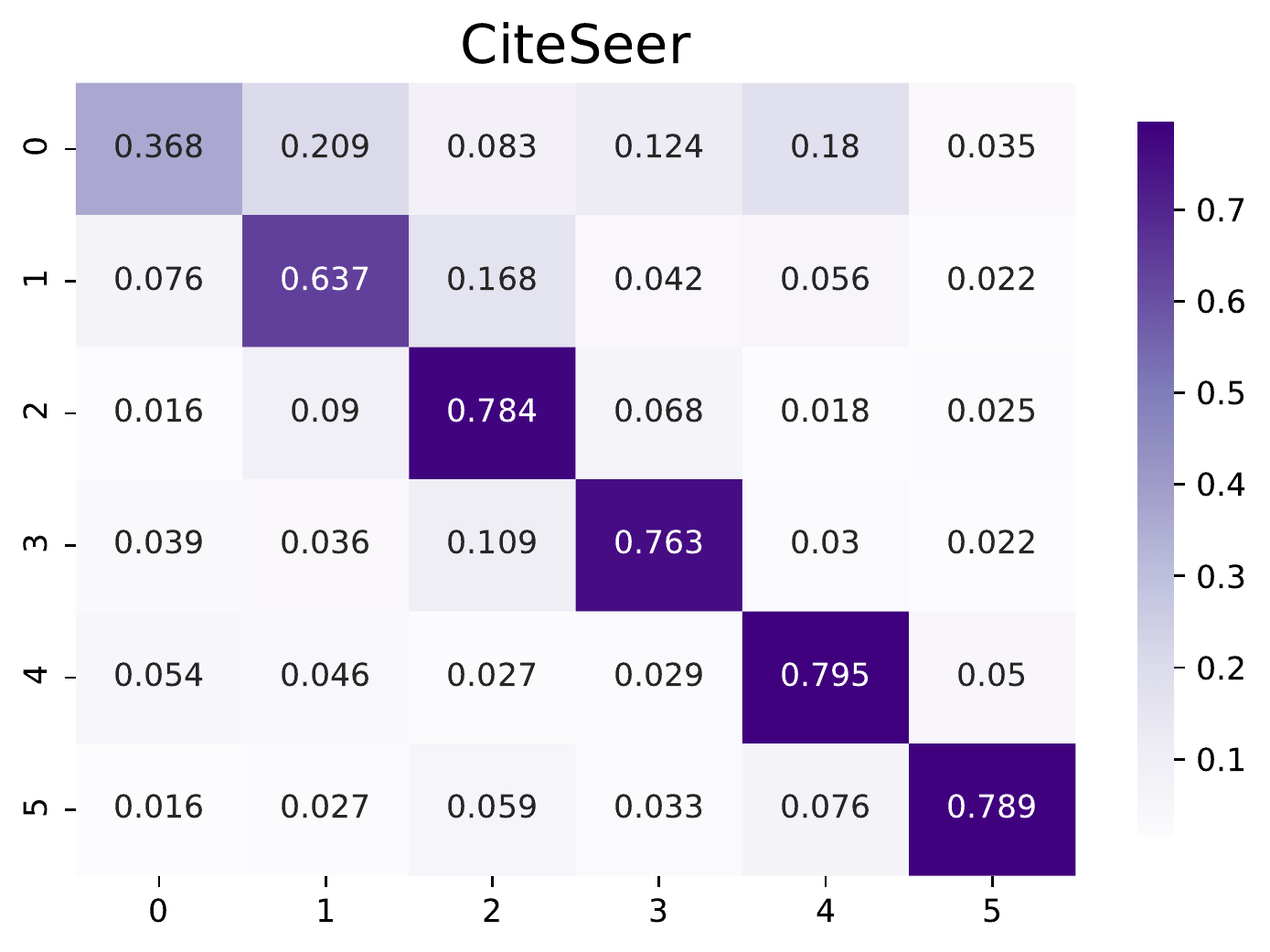} &
    \includegraphics[width=.27\textwidth]{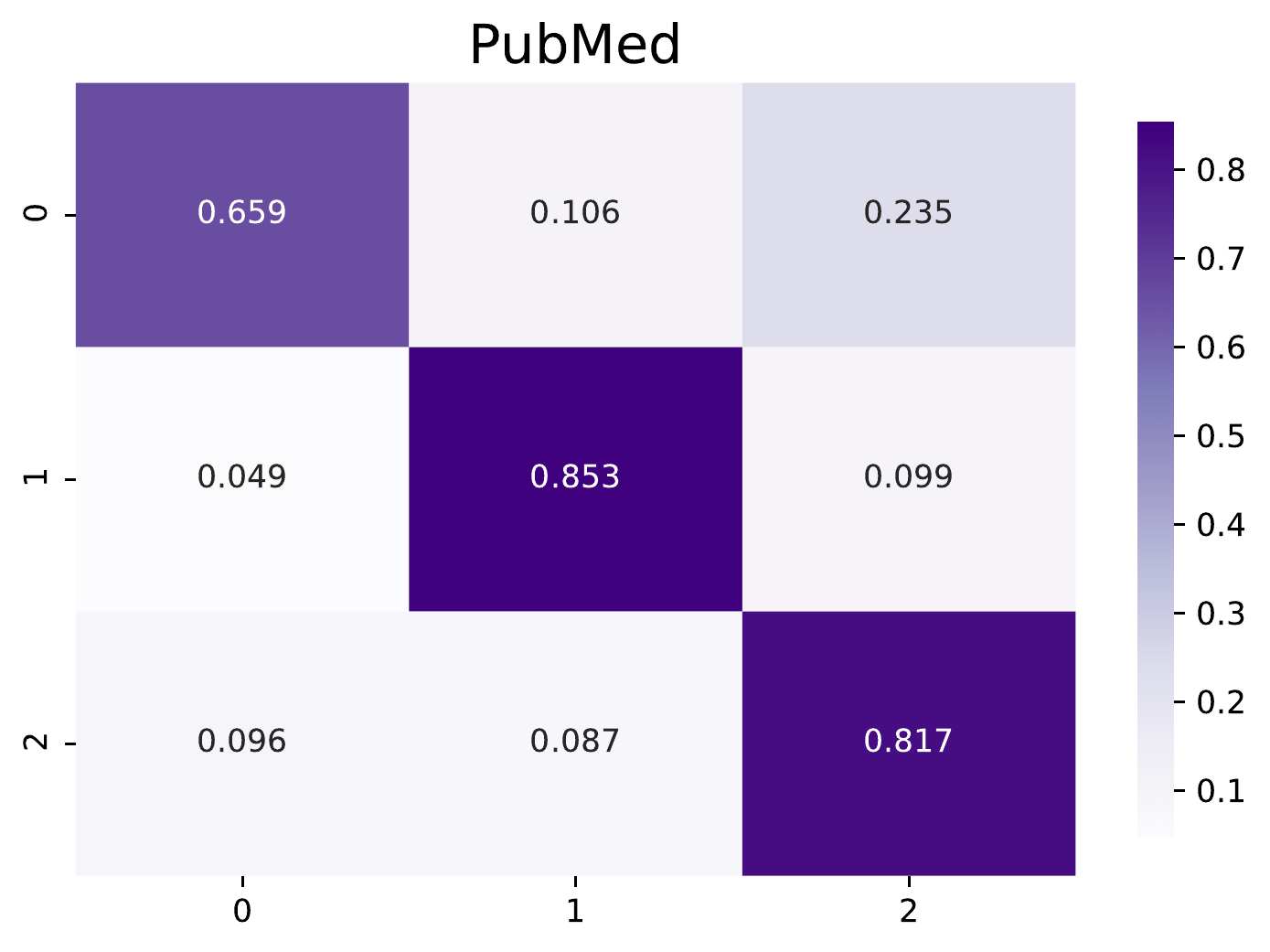} \\
    \includegraphics[width=.27\textwidth]{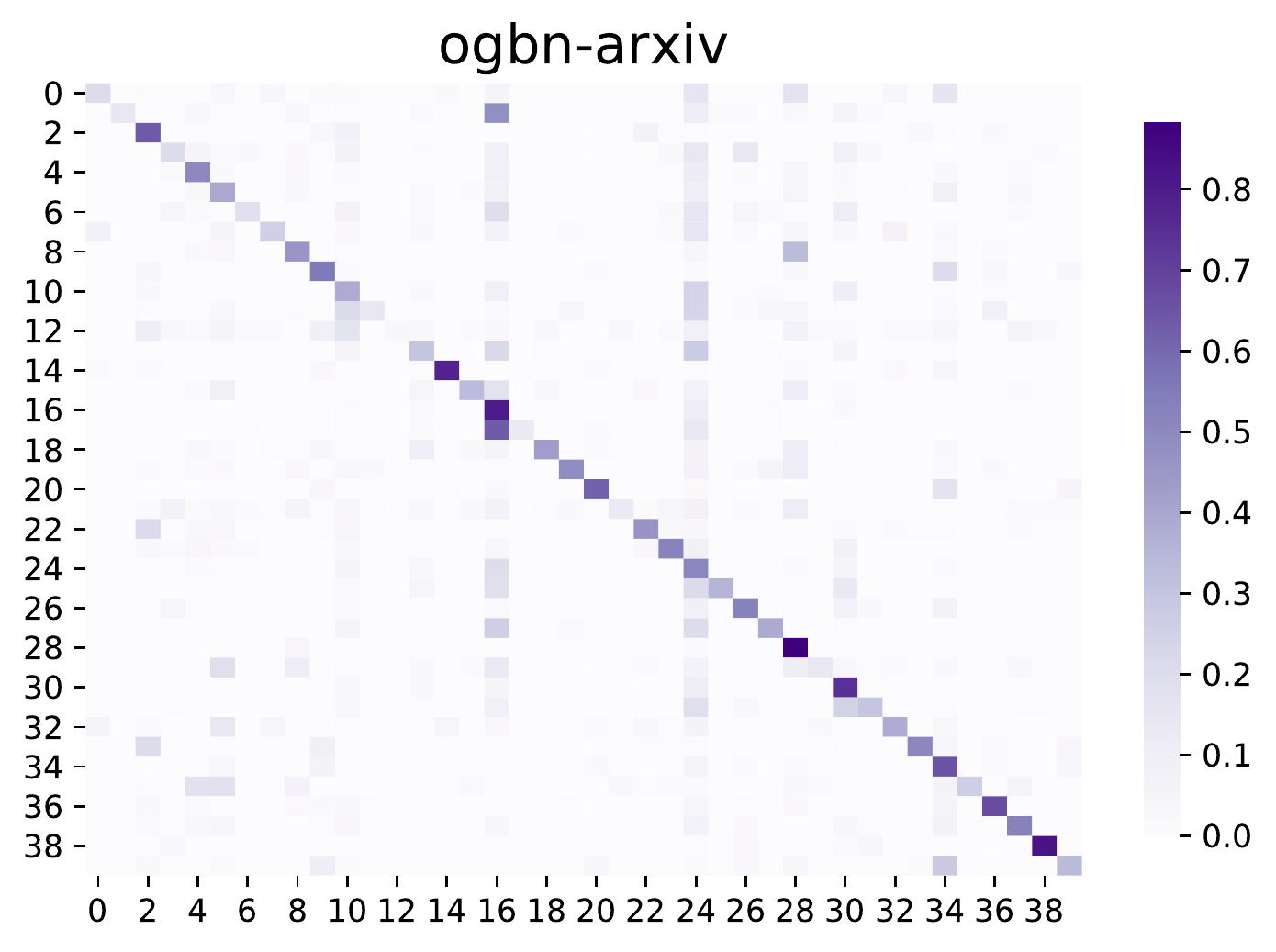} &
    \includegraphics[width=.27\textwidth]{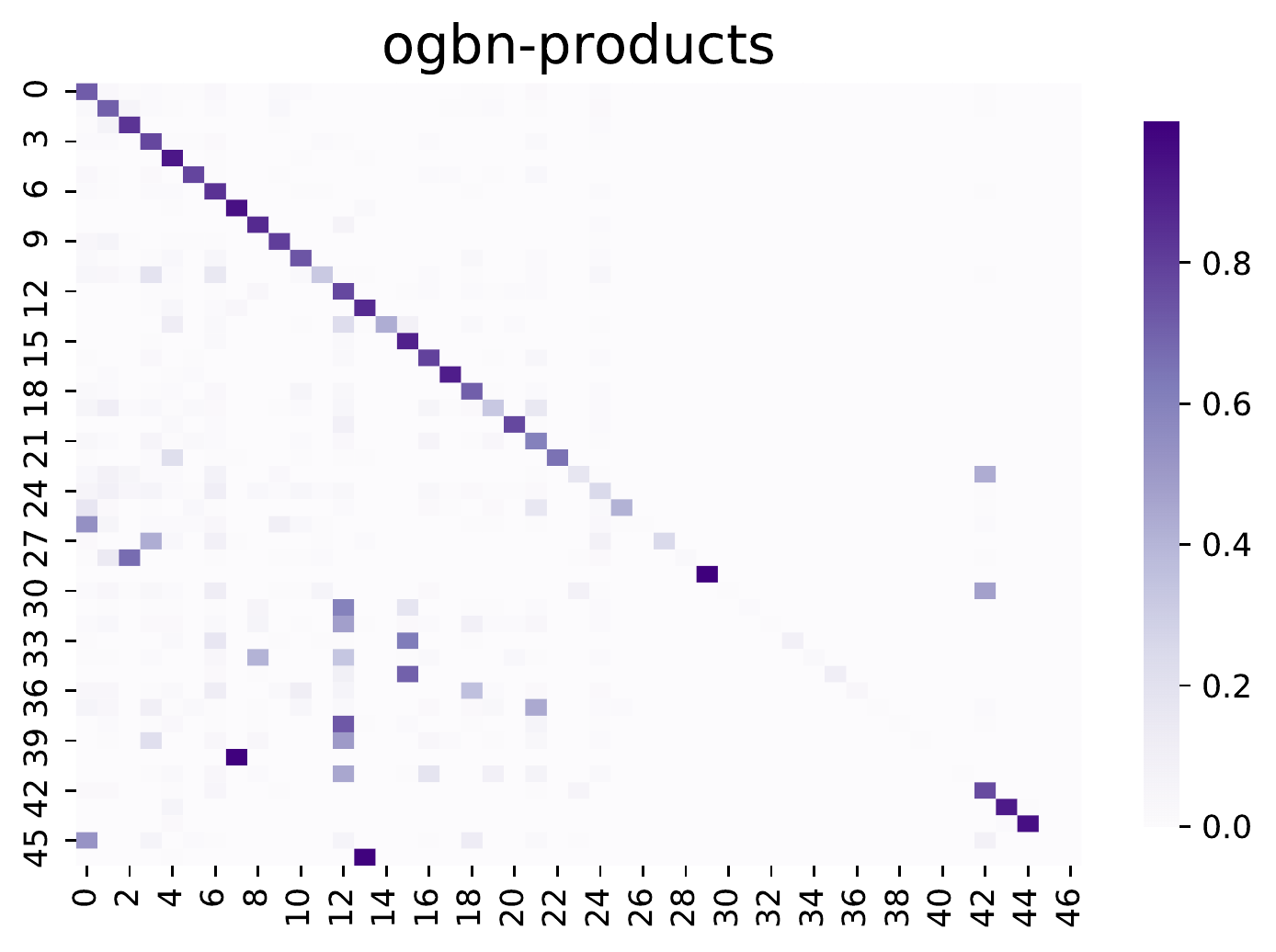} &
    \includegraphics[width=.27\textwidth]{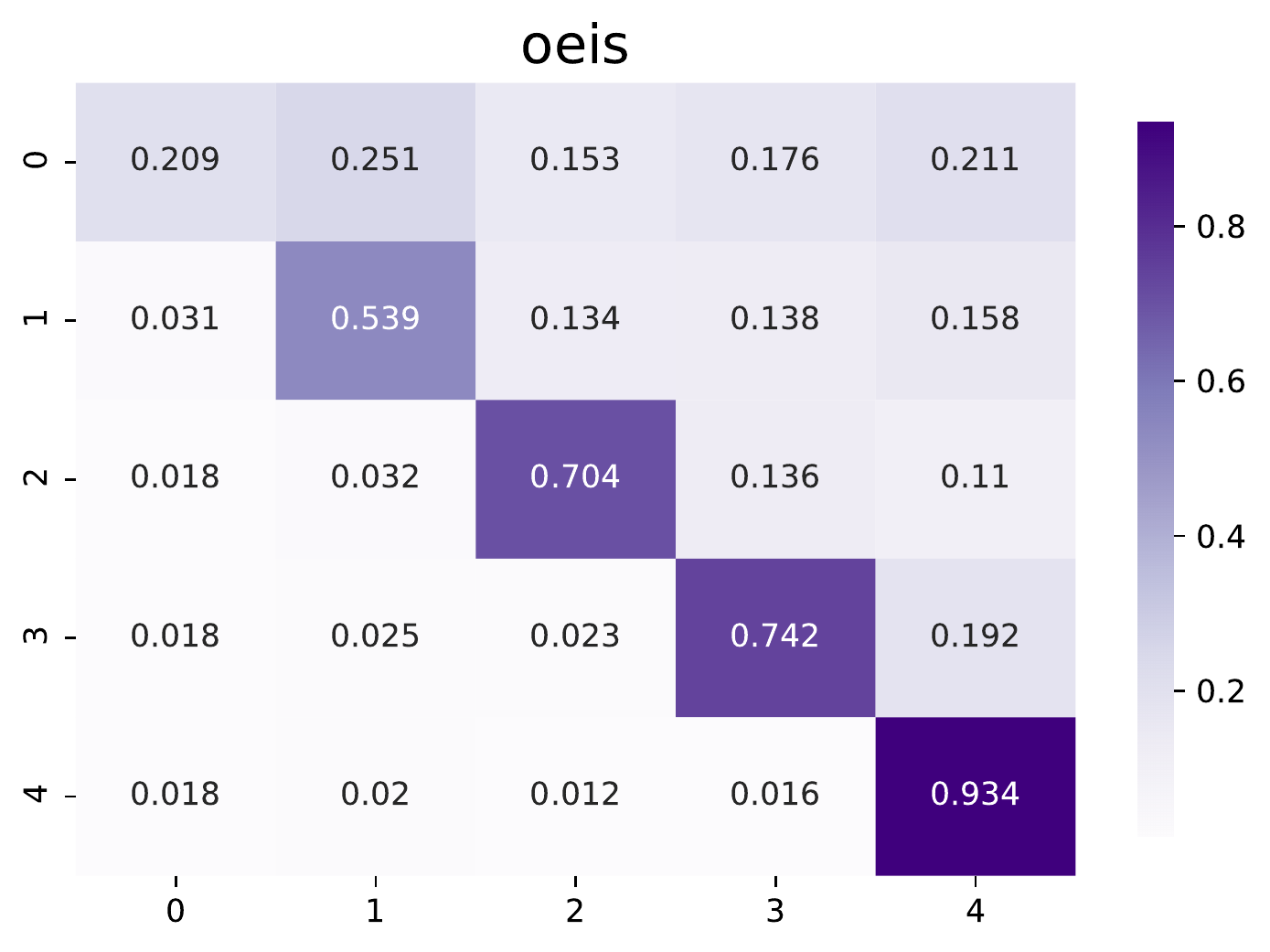}
\end{tabular}
\caption{Compatibility matrices of homophilic datasets. The diagonal dominance indicates strong homophily.}
\label{fig:compat_homophilic}
\end{figure*}

In contrast to the different compatibility matrix structures of our proposed non-homophilous datasets, much other graph data have primarily homophilous relationships, as can be seen in Figure~\ref{fig:compat_homophilic} and Table \ref{tab:homophilic_stats}. The Cora, CiteSeer, PubMed, ogbn-arXiv, and ogbn-products datasets are widely used as benchmarks for node classification \cite{yang2016revisiting, hu2020open}, and are highly homophilous, as can be seen by the diagonally dominant structure of the compatibility matrices and by the high edge homophily and $\hat h$.

We collected the oeis dataset displayed in the bottom right of Figure~\ref{fig:compat_homophilic}. The nodes are entries in the Online Encyclopedia of Integer Sequences \cite{sloane2007line}, and directed edges link an entry to any other entry that it cites. In analogy to arXiv-year and snap-patents, the node labels are the time of posting of the sequence. However, in this case the graph relationships are homophilous, even as we vary the number of distinct classes (time periods). This is in part due to differences between posting in this online encyclopedia and publication of academic papers or patents. For instance, there is less overhead to posting an entry in the OEIS, so users often post separate related entries and variants of these entries in rapid succession. Also, an entry in the encyclopedia often inspires other people to work on similar entries, which can be created in much less time than an academic follow-up work to a given paper. These related entries tend to cite each other, which contributes to homophilic relationships over time. Thus, the data here does not follow the special temporal citation structure of academic publications and patents.

\subsection{Previous Non-Homophilous Data}

\begin{figure*}[t]
\centering
\begin{tabular}{ccc}
    \includegraphics[width=.27\textwidth]{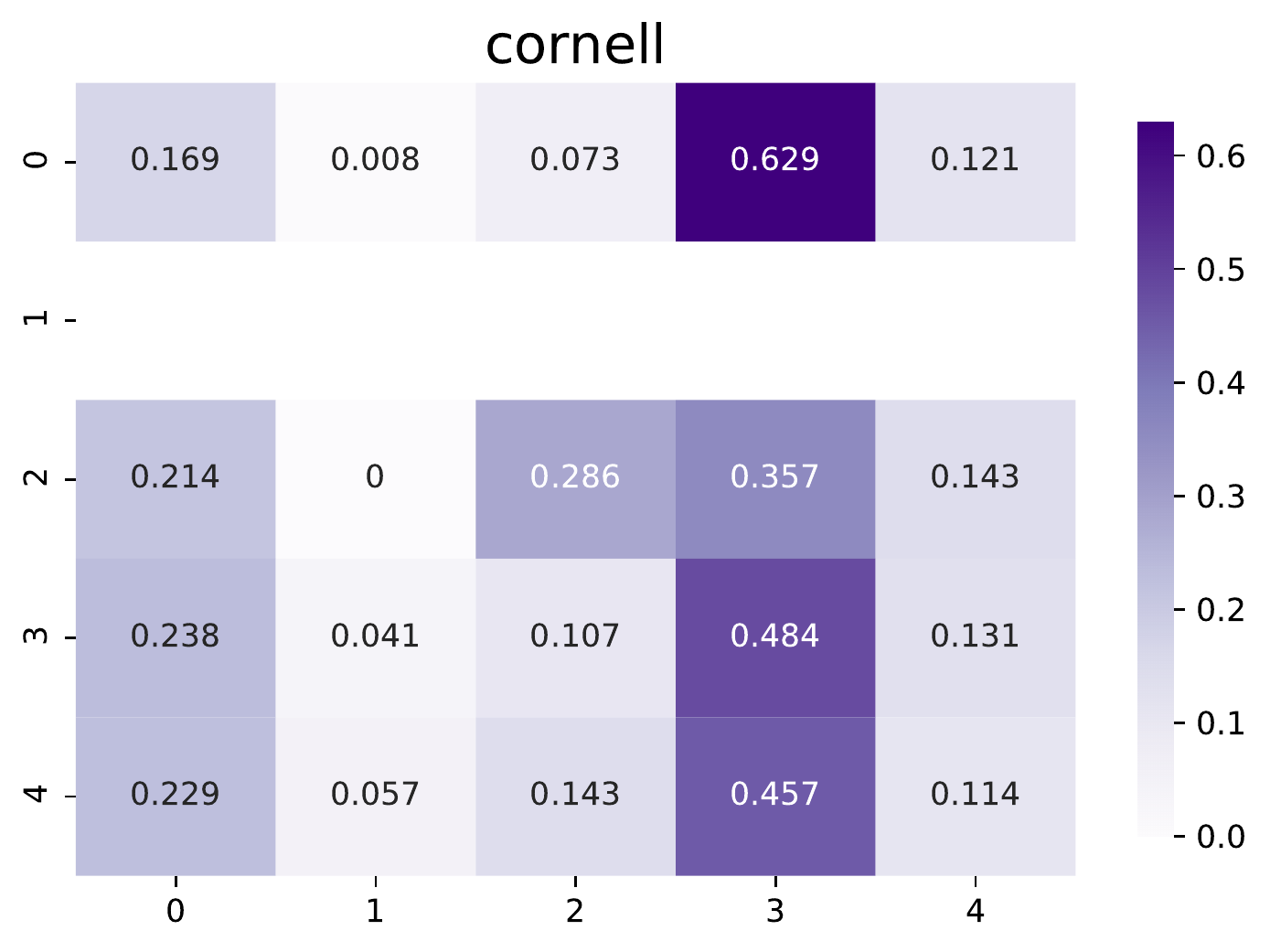} &
    \includegraphics[width=.27\textwidth]{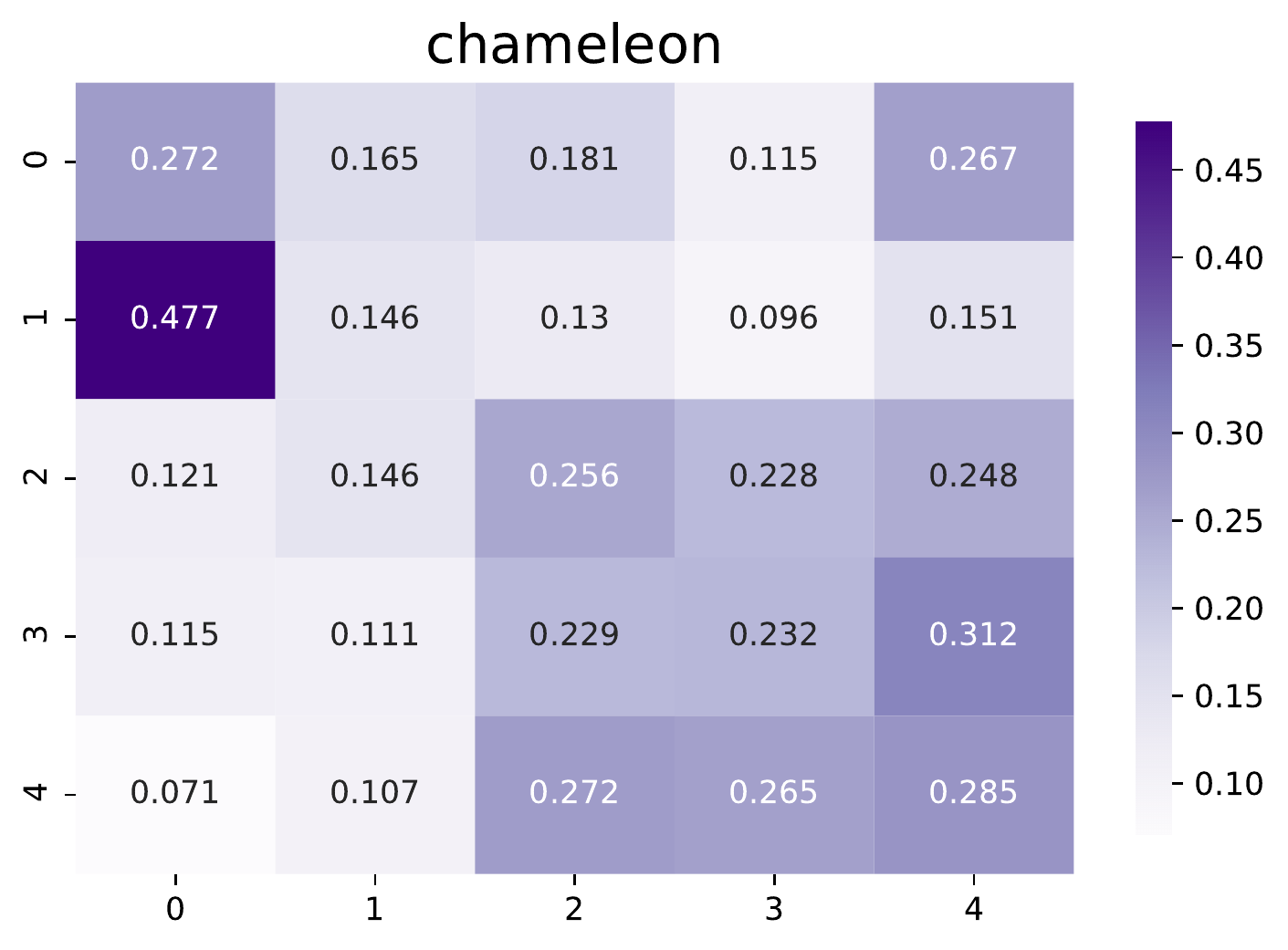} &
    \includegraphics[width=.27\textwidth]{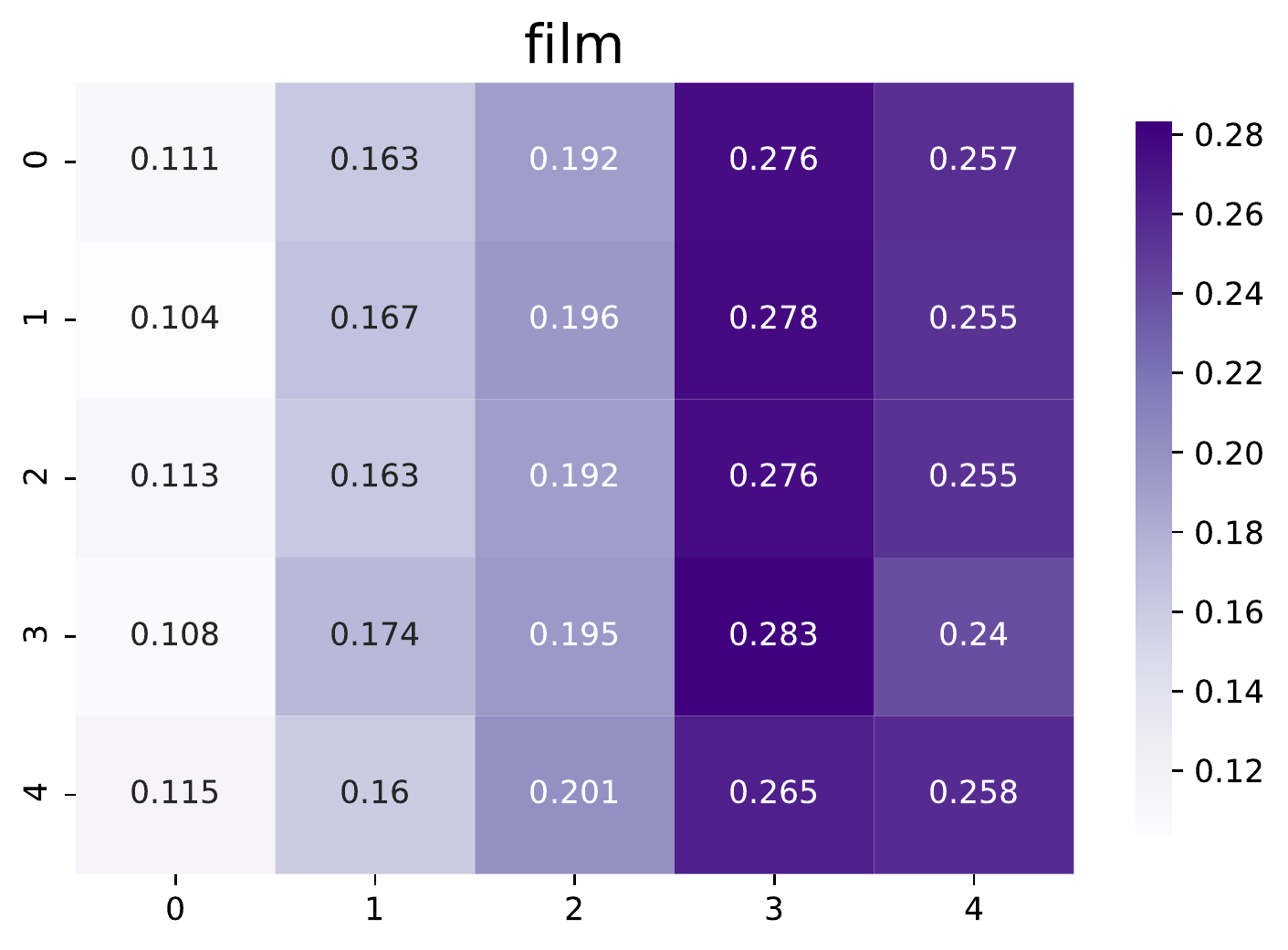} \\
    \includegraphics[width=.27\textwidth]{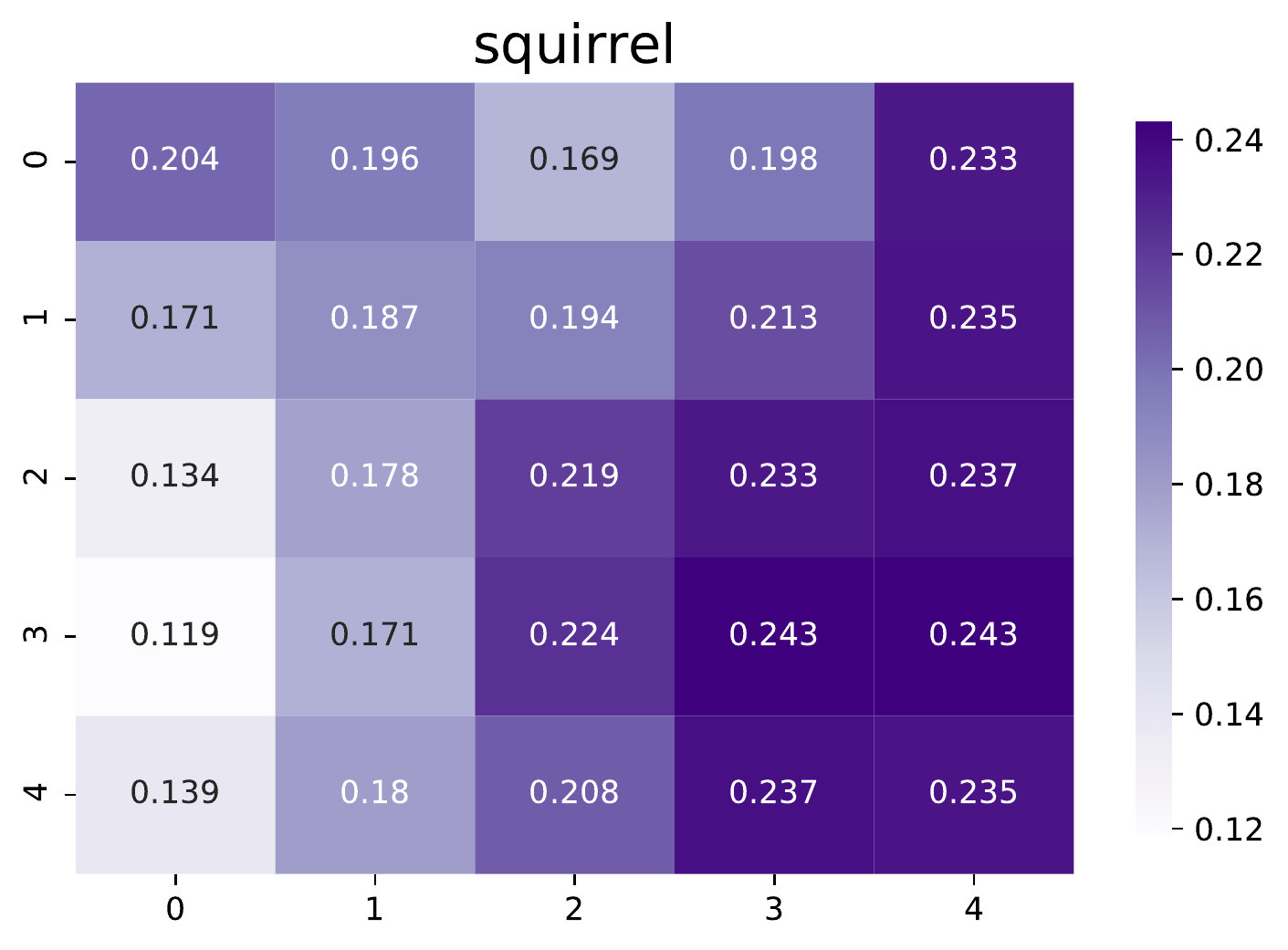} &
    \includegraphics[width=.27\textwidth]{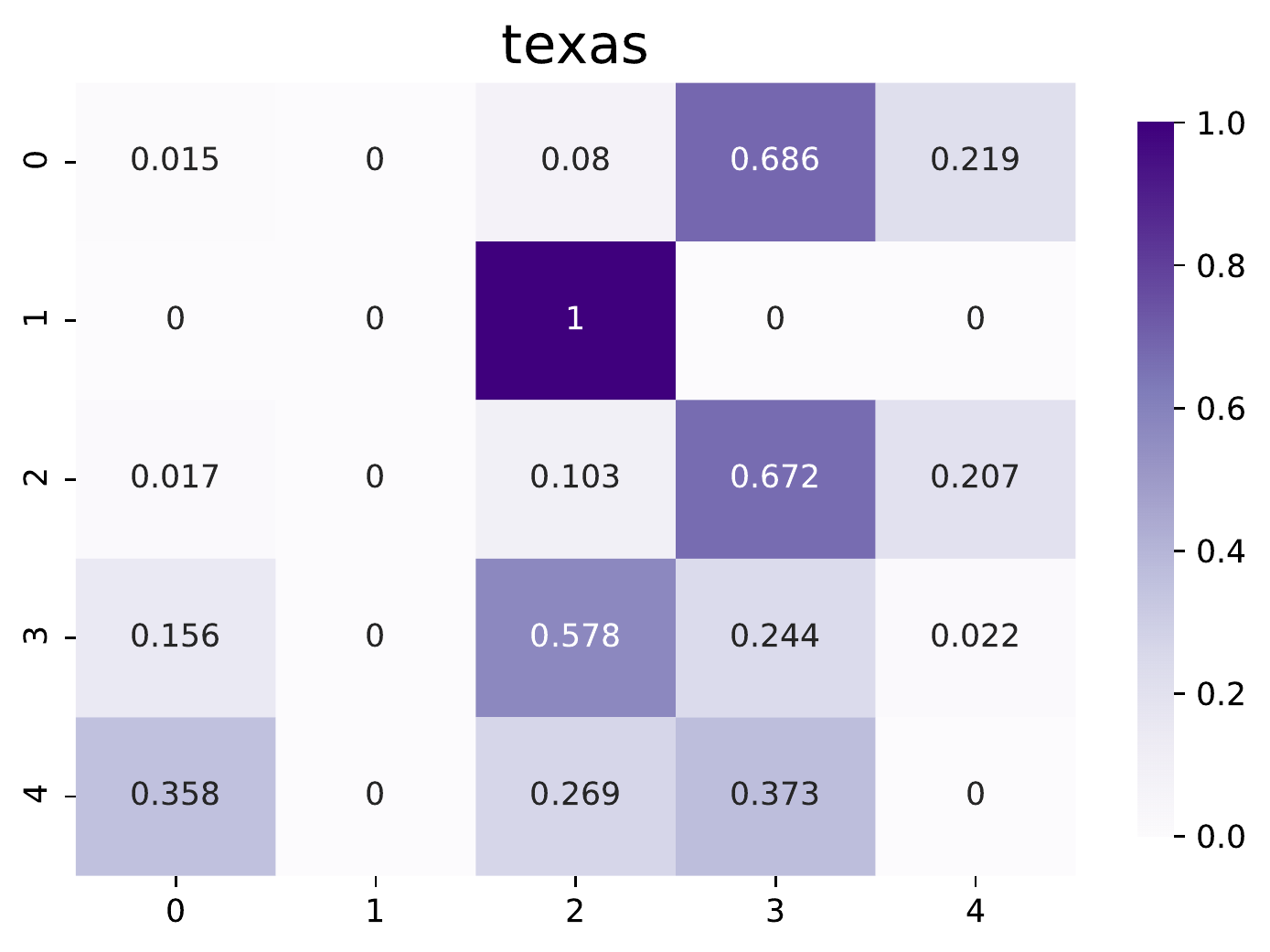}  &
    \includegraphics[width=.27\textwidth]{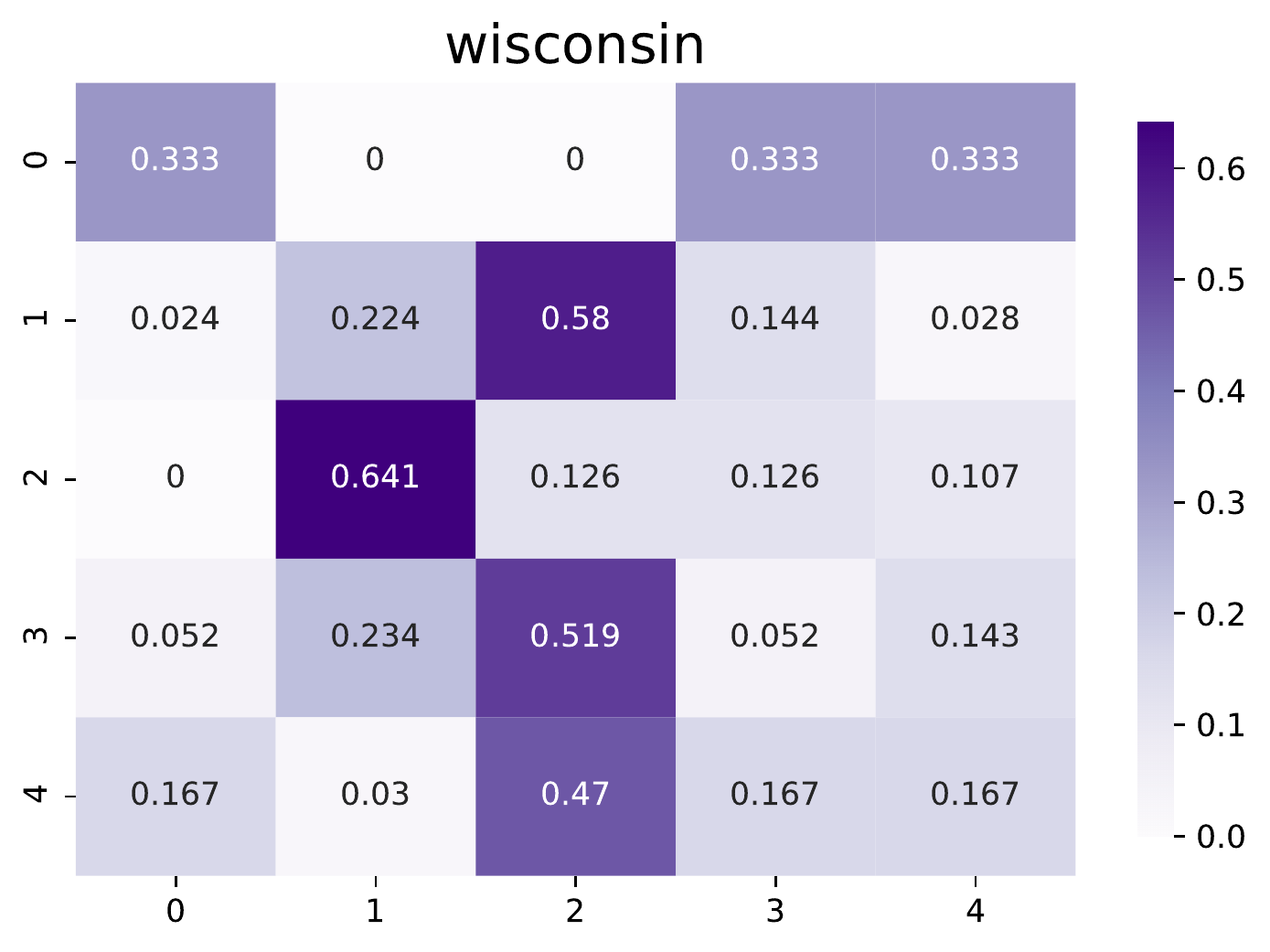}
\end{tabular}
\caption{Compatibility matrices of datasets in \citet{pei2019geom} (collected by \cite{rozemberczki2019multi, tang2009social, mitchell1997web}). The ``film'' dataset is also referred to as ``Actor''. Note that there are no edges leading out of the nodes of class 1 in the Cornell dataset, so there is an empty row in its matrix.}
\label{fig:compat_geom_gcn}
\end{figure*}

For the six datasets in \citet{pei2019geom} often used in evaluation of graph representation learning methods in non-homophilous regimes \cite{zhu2020beyond}, basic statistics are listed in Table \ref{tab:geom_gcn} and compatibility matrices are displayed in Figure \ref{fig:compat_geom_gcn}. We propose datasets that have up to orders of magnitude more nodes and edges and come from a wider range of contexts. There are several cases of class-imbalance in these previously used datasets, which may make the edge homophily misleading. As discussed in Appendix \ref{sec:measure}, our measure may be able to alleviate issues with edge homophily in measuring homophily of these datasets, and offers a way to distinguish between the Chameleon, Actor, and Squirrel datasets that all have similar edge homophily.

\subsection{General Non-homophilous Settings}

Different settings in which non-homophilous / disassortative relationships are prevalent have been identified in the literature, and many of these non-homophilous settings are represented by our proposed datasets. We list these general non-homophilous settings for reference, with our proposed datasets that belong to these settings in parentheses:
\begin{itemize}
    \item Gender relations in social or interaction networks  \cite{altenburger2018monophily, chin2019decoupled, jia2020residual} (Penn94, Pokec).
    \item Technological and internet relationships, such as in web page connections \cite{newman2003mixing, pei2019geom} (wiki).
    \item Malicious or fraudulent nodes, such as in auction networks \cite{chau2006detecting, pandit2007netprobe} (genius).
    \item Publication time in citation networks \cite{peel2017graph} (arXiv-year, snap-patents).
    \item Biological structures such as in food webs \cite{gatterbauer2014semi} and protein interactions \cite{newman2003mixing}.
    \item Specific online user attributes \cite{rozemberczki2021twitch} (twitch gamers) 
\end{itemize}
While not all example graph data from these contexts are non-homophilous, a diverse range are. In order to succeed in future applications in these contexts, it is of importance to develop methods that are able to handle non-homophilous structures.

\subsection{Class-Imbalance and Metrics}

\begin{figure}[ht!]
    \centering
    \includegraphics[width=.35\textwidth]{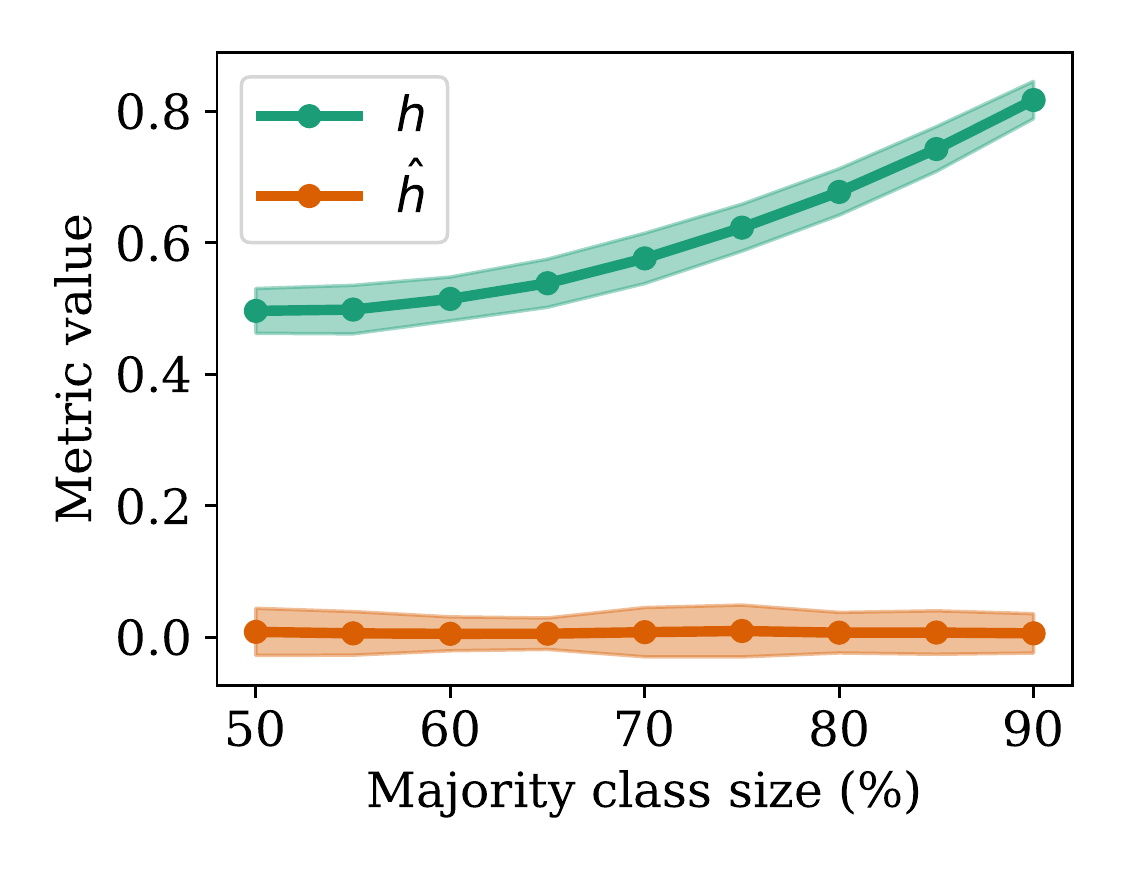}
    \caption{Comparison of edge homophily $h$ and our measure $\hat h$ on random class-imbalanced graph data with edges independent of node labels. Three standard deviations are shaded. Our measure is mostly constant as the classes become more imbalanced, while edge homophily increases.}
    \label{fig:metrics}
\end{figure}

In this section, we present experiments that demonstrate an instance in which our metric is not affected by imbalanced classes, while edge homophily is. We generate graphs in which node labels are independent of edges by randomly choosing node labels and generating graph edges by the Erd\H{o}s-R\'enyi random graph model \cite{erdHos1960evolution}. In particular, we fix the number of classes to two, the number of nodes to 100, and the probability of edge formation as .25 between every pair of nodes. Then we generate 100 samples of these random graphs, and compute the mean and standard deviation of both edge homophily $h$ and our measure $\hat h$. As seen in Figure~\ref{fig:metrics}, our measure $\hat h$ is constantly near zero as we increase the size of the majority class, while the edge homophily $h$ increases as the size of majority class increases.

\subsection{Degree Distributions of Proposed Non-homophilous datasets} 

\begin{figure}
    \centering
    \begin{tabular}{cc}
    \includegraphics[width=.35\textwidth]{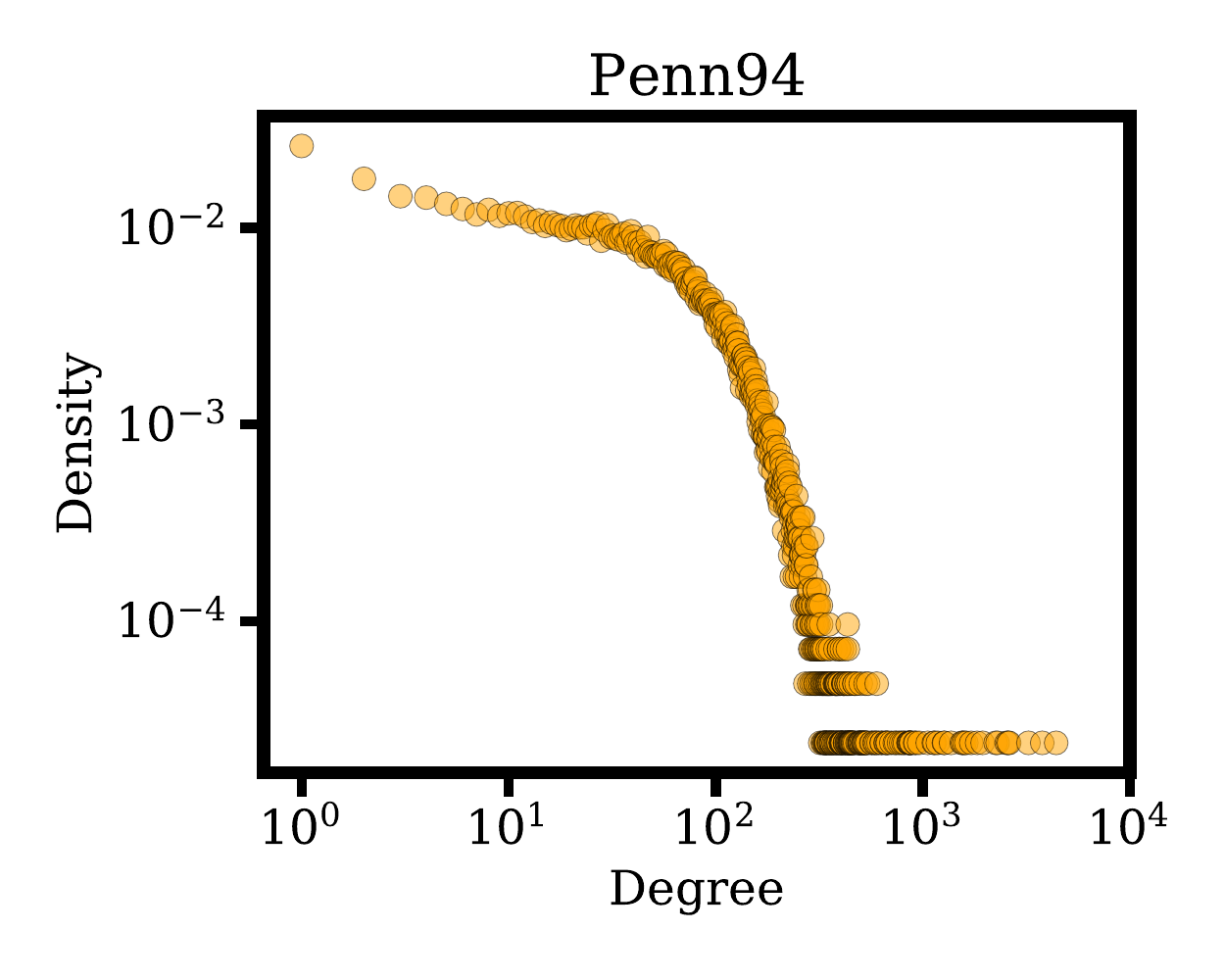} &
    \includegraphics[width=.35\textwidth]{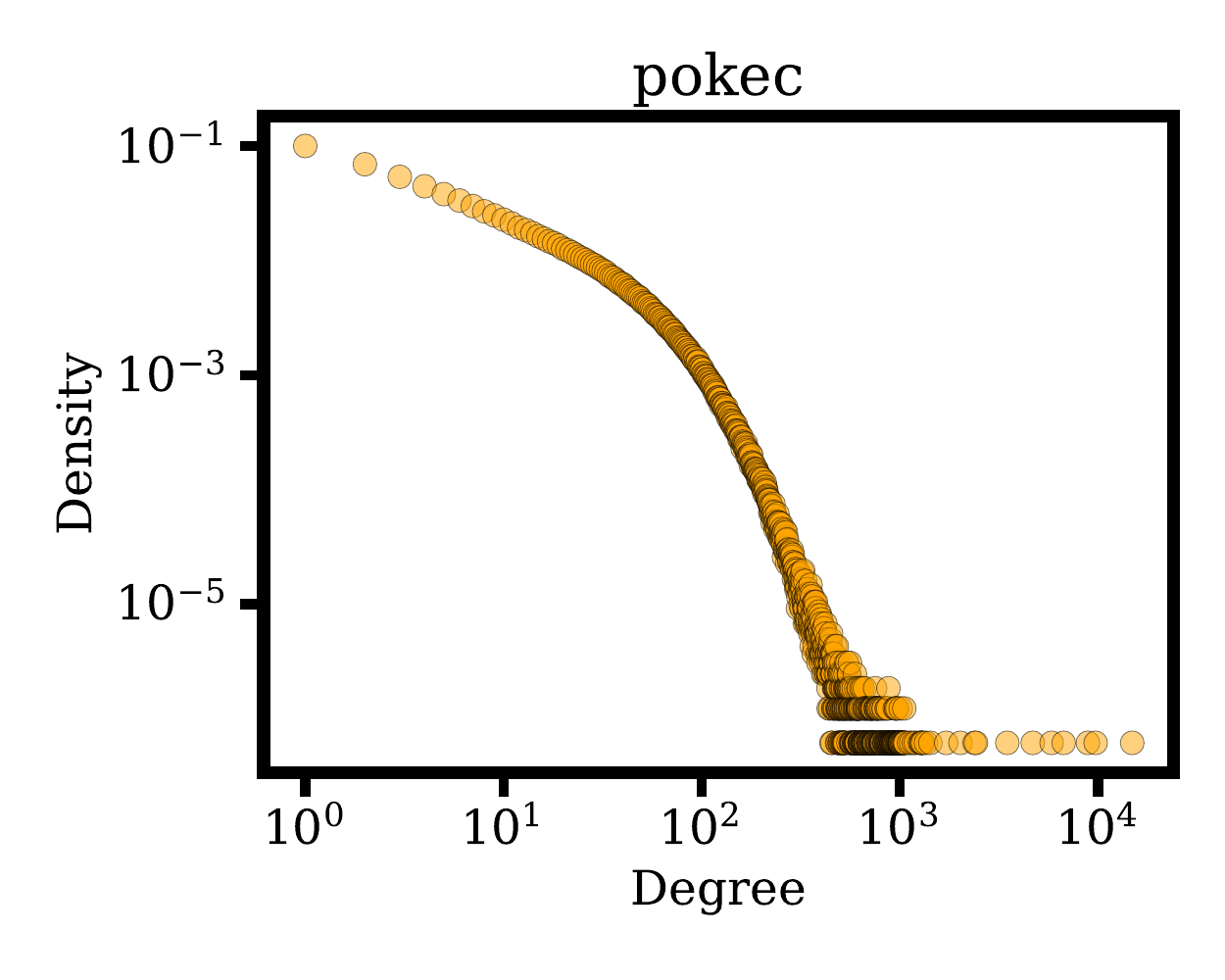} \\
    \includegraphics[width=.35\textwidth]{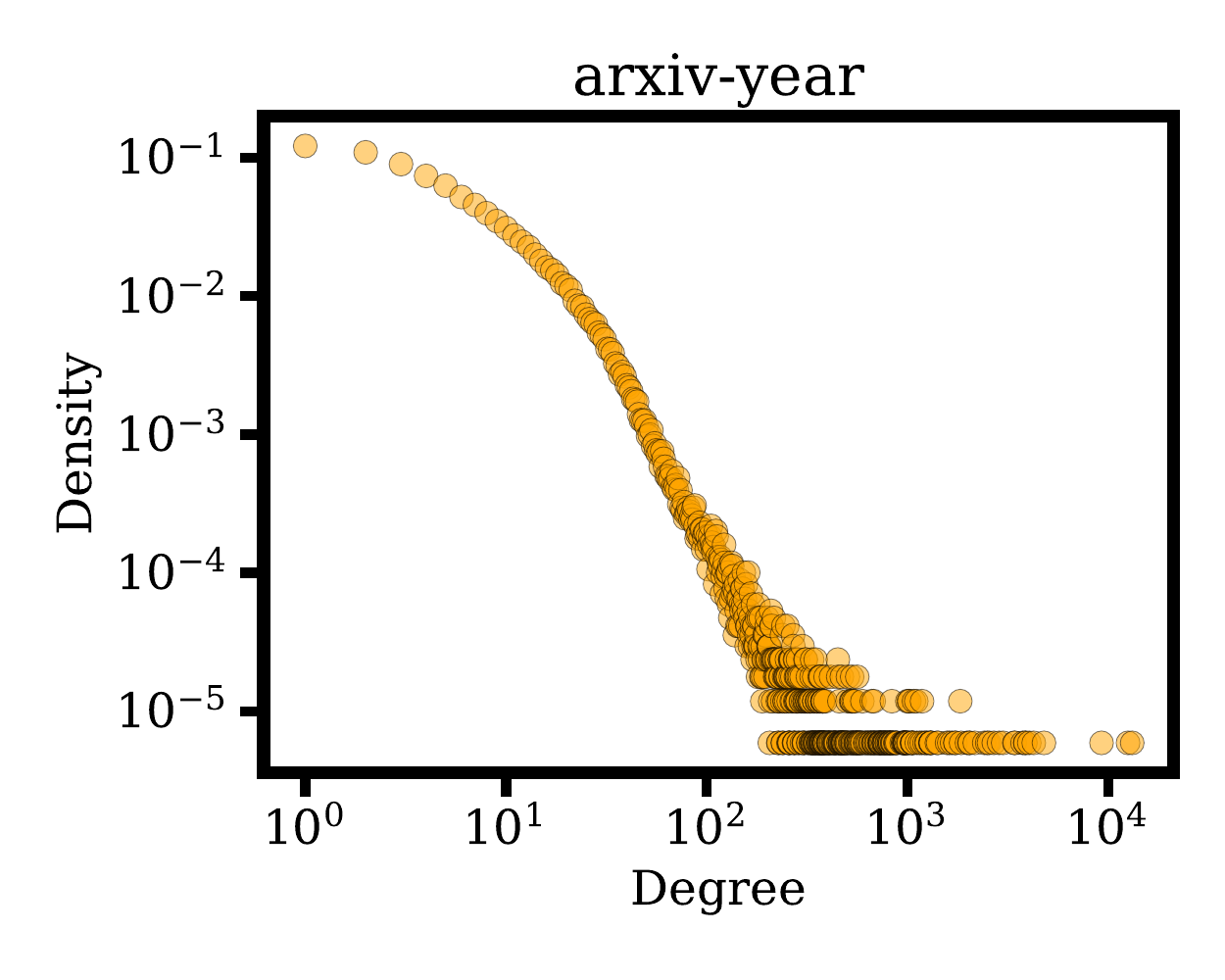} &
    \includegraphics[width=.35\textwidth]{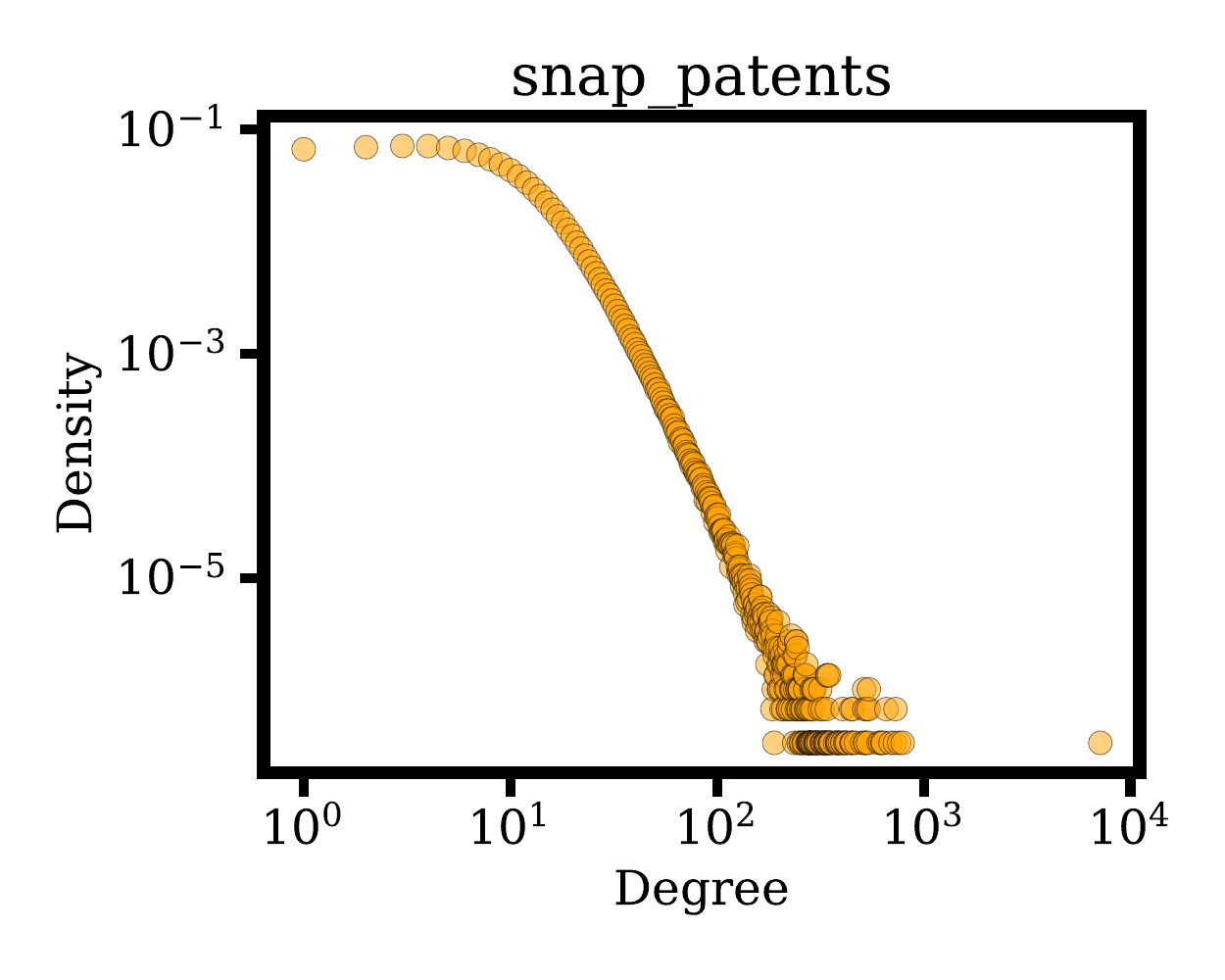} \\
    \includegraphics[width=.35\textwidth]{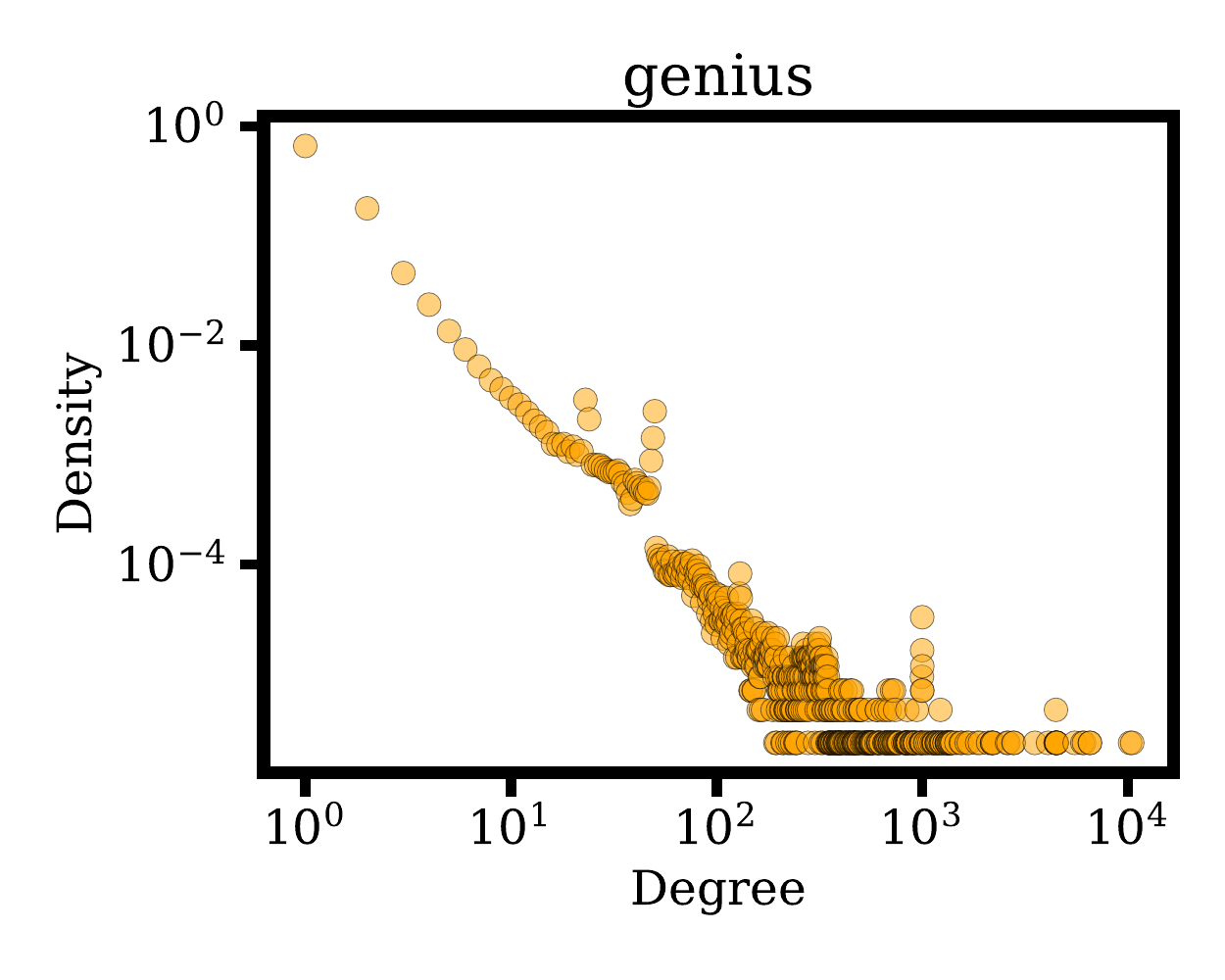} &
    \includegraphics[width=.35\textwidth]{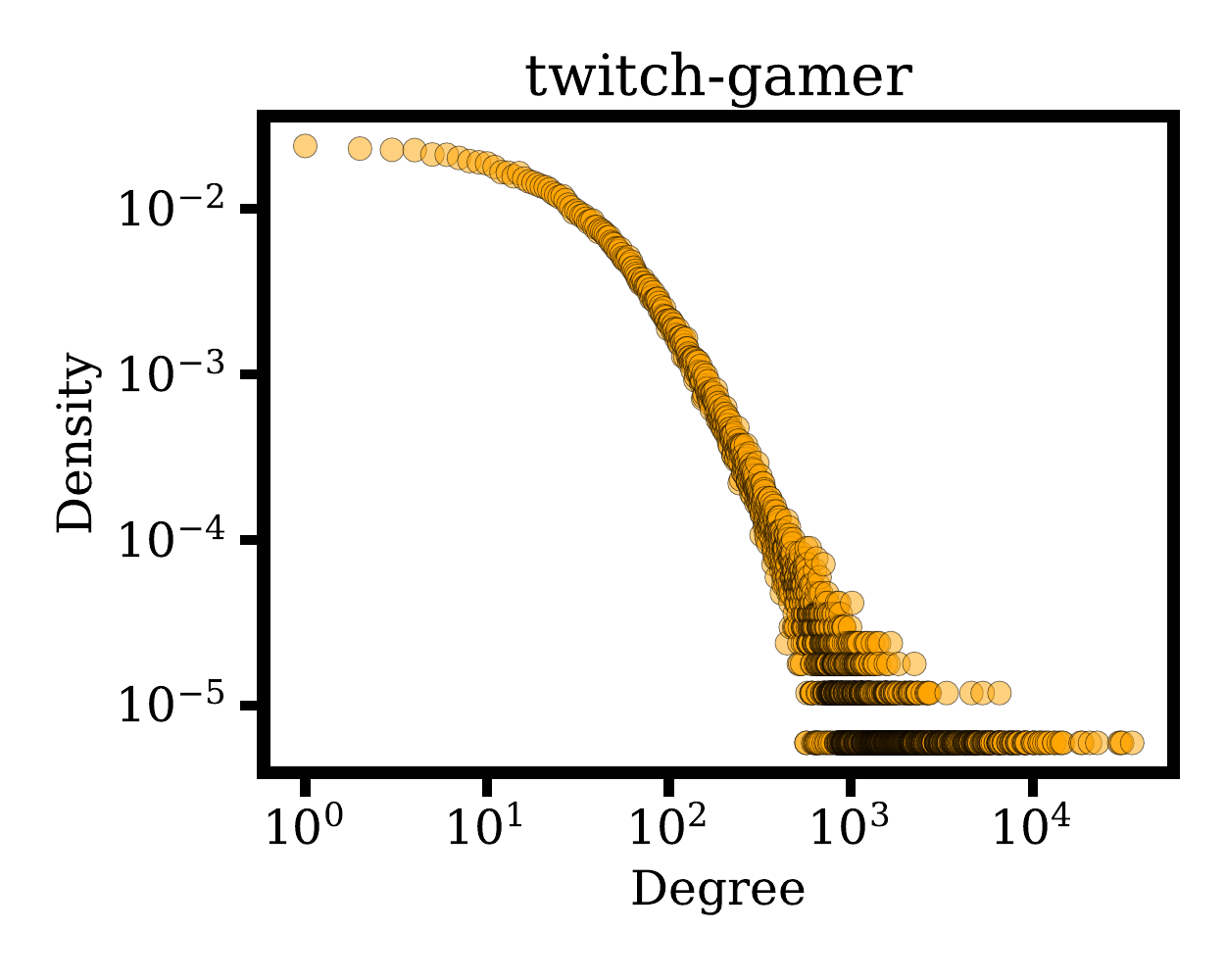} \\
    \includegraphics[width=.35\textwidth]{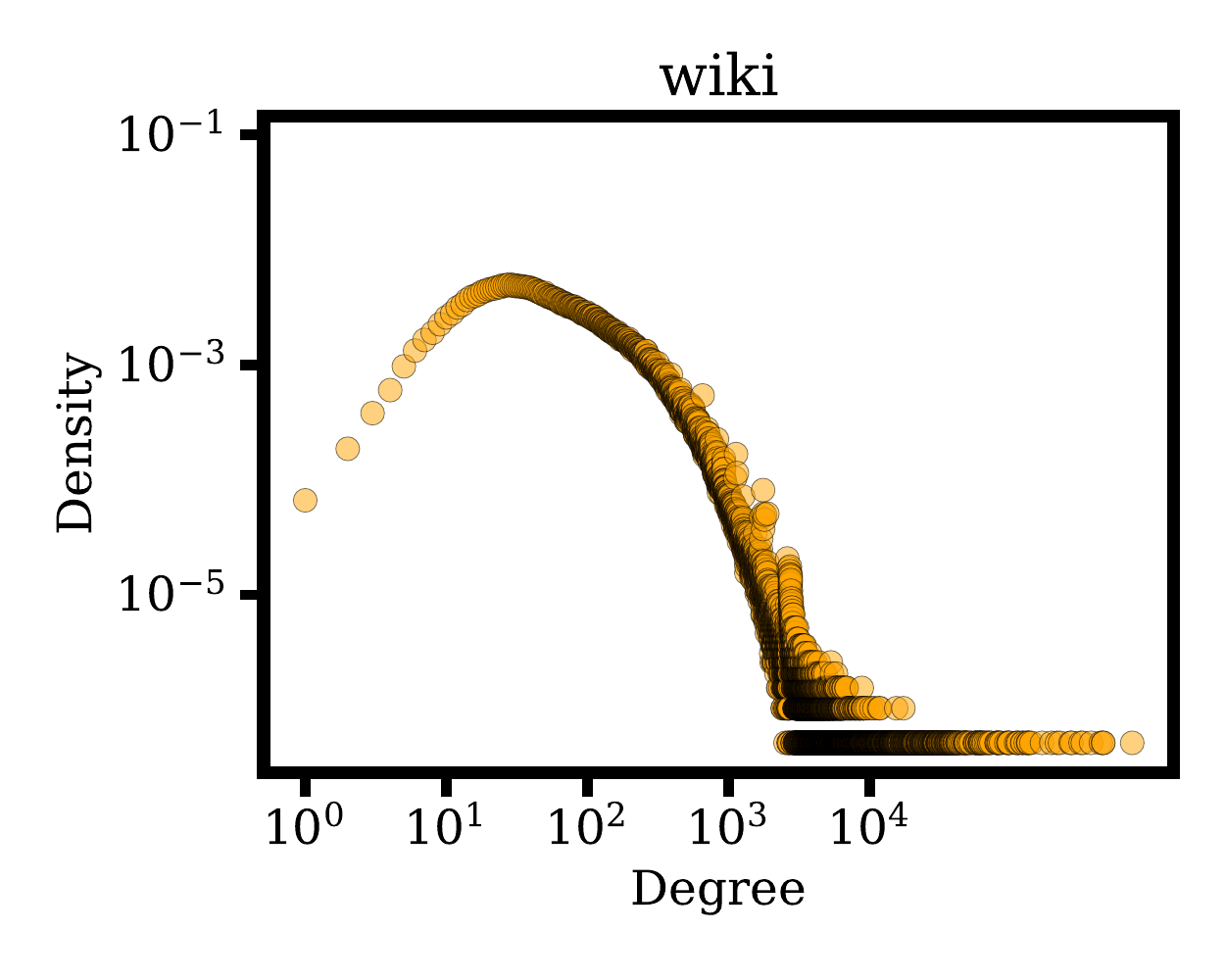} &
    \end{tabular}
    \caption{Degree distribution of our proposed non-homophilous datasets}
    \label{fig:degree_distr}
\end{figure}

Past work has found that the degree distribution can affect the performance of models on node classification~\cite{zhu2020beyond}. As a result, we provide degree distributions of all of our proposed non-homophilous datasets in Figure~\ref{fig:degree_distr}. The degree distributions are all heavy-tailed, as is typically expected in real-world graph data. The wiki distribution also has an interesting additional property, where the mode appears to be at a degree between 10 and 100.

\subsection{Two-hop homophily levels}

While the homophily measures are based on one-hop information, we may also consider properties of two-hop neighborhoods. Even though there are limitations to measuring higher-order homophily~\cite{veldt2021higher}, here we estimate an interesting quantity in two-hop neighborhoods. We use a node homophily measure, where the neighborhood of each node is defined to be the nodes of exactly two hops away:
\begin{equation}
    \frac{1}{|V|} \sum_{v \in V} \frac{\text{Number of exact two-hop neighbors of $v$ with same class as $v$}}{\text{Number of $v$'s exact two-hop neighbors}}.
\end{equation}
This is an expensive operation to compute, so we approximate the sum over a random sample of $k$ nodes, and then replace the normalization factor of $\frac{1}{|V|}$ by $\frac{1}{k}$. Note that for each node $v$ in the sample, the inner sum is still using the two-hop neighborhoods of the original graph.  Table~\ref{tab:two-hop} compares the computed one-hop and two-hop measures. The two-hop measure is mostly similar to or somewhat higher than the standard one-hop node homophily.
\begin{table}[ht]
    \centering
    \caption{Comparison of node homophily over one-hop neighborhoods and strict two-hop neighborhoods. The measure over two-hop neighborhoods is estimated with $k=500$ sampled nodes.}
    \label{tab:two-hop}
    \begin{tabular}{lll}
    \toprule 
         Dataset & Two-hop & One-hop  \\
         \midrule
         Penn94 & .474 & .483 \\
         pokec & .611 & .428 \\
         arXiv-year & .341 & .289\\
         snap-patents & .330 & .221\\
         genius & .749 & .508\\
         twitch-gamers & .513 & .556\\
         \bottomrule
    \end{tabular}
\end{table}

\section{Experimental Details}\label{sec:exp_details}

For gradient-based optimization, we use the AdamW optimizer \cite{kingma2014adam, loshchilov2018decoupled} with weight decay .001  and learning rate $.01$ by default, unless we tune the optimizer for a particular method (as noted below in \ref{sec:hparams}).  Hyperparameter tuning is conducted using grid search for most methods. Tuning for C\&S is done as in the original paper \cite{huang2021combining}, which uses Optuna \cite{optuna} for Bayesian hyperparameter optimization. All graphs are treated as undirected besides arXiv-year and snap-patents, in which the directed nature of the edges capture useful temporal information; however, we find that label propagation and C\&S (which builds on label propagation) perform better with undirected graphs in these cases, so we keep the graphs as undirected for these methods. 

We implement our methods and run experiments in PyTorch~\cite{paszke2019pytorch} (3-clause BSD license), and make heavy use of the PyTorch-Geometric library~\cite{fey2019fast} (MIT license) for graph representation learning. For full-batch training, simple methods are run on a NVIDIA 2080 Ti with 11 GB GPU memory. In cases where the NVIDIA 2080Ti did not provide enough memory, we re-ran experiments on a NVIDIA Titan RTX with 24 GB GPU memory, reporting (M) in Table \ref{tab:results} if the GPU memory was still insufficient. For minibatch training on wiki, we also make use of NVIDIA RTX 3090 GPUs with 24~GB GPU memory. Each experiment was run on one GPU at a time.
 
\subsection{Full-batch Hyperparameters}\label{sec:hparams}

Experimental results for full-batch training are reported on the hyperparameter settings below, where we choose the settings that achieve the highest performance on the validation set.
We choose hyperparameter grids that do not necessarily give optimal performance, but hopefully cover enough regimes so that each model is reasonably evaluated on each dataset. Unless otherwise stated, each GNN has dropout of .5 \cite{srivastava2014dropout} and BatchNorm \cite{ioffe2015batch} in each layer. The hyperparameter grids for the different methods are:

\begin{itemize}
    \item MLP: hidden dimension $\in \{16, 32, 64, 128, 256\}$, number of layers $\in \{2,3\}$. We use ReLU activations.
    \item Label propagation: $\alpha \in \{.01, .1, .25, .5, .75, .9, .99\}$. We use 50 propagation iterations.
    \item LINK: weight decay $\in \{.001, .01, .1\}$.
    \item SGC: weight decay $\in \{.001, .01, .1\}$.
    \item C\&S: Normalized adjacency matrix A\textsubscript{1}, A\textsubscript{2} $\in \{D^{-\frac{1}{2}}AD^{-\frac{1}{2}}, D^{-1}A, AD^{-1} \}$ for the residual propagation and label propagation, where $A$ is the adjacency matrix of the graph and $D$ is the diagonal degree matrix; $\alpha_1, \alpha_2 \in (0.0, 1.0)$ for the two propagations. Both Autoscale and FDiff-scale were used for all experiments, and scale $\in (0.1, 10.0)$ was searched in FDiff-scale settings. The base predictor is chosen as the best MLP model for each dataset.
    \item GCN: lr $\in \{.1, .01, .001\}$, hidden dimension $\in \{4, 8, 16, 32, 64\}$, except for snap-patents and pokec, where we omit hidden dimension = 64. Each activation is a ReLU. 2 layers were used for all experiments.
    \item GAT: lr $\in \{.1, .01, .001\}$. For snap-patents and pokec: hidden channels $\in \{4, 8, 12\}$ and gat heads $\in \{2, 4\}$. For all other datasets: hidden channels $\in \{4, 8, 12, 32\}$ and gat heads $\in \{2, 4, 8\}$. We use the ELU activation \cite{clevert2015fast}. 2 layers were used for all experiments. 
    \item GCNJK: Identical for GCN, also including JK Type $\in \{\text{cat}, \text{max} \}$.
    \item GATJK: Identical for GAT, also including JK Type $\in \{\text{cat}, \text{max} \}$. 
    \item APPNP: MLP hidden dimension $\in \{16, 32, 64, 128, 256\}$, learning rate $\in \{.01, .05, .002\}$, $\alpha \in \{.1, .2, .5, .9\}$.  We truncate the series at the $K=10$th power of the adjacency.
    \item H\textsubscript{2}GCN: hidden dimension $\in \{8, 16, 32, 64\}$, number of layers $\in \{1,2\}$, dropout $\in \{0, .5\}$. The architecture follows Section 3.2 of \cite{zhu2020beyond}.
    \item MixHop: hidden dimension $\in \{8, 16, 32\}$, number of layers $\in \{2, 3\}$. Each layer has uses the 0th, 1st, and 2nd powers of the adjacency and has ReLU activations. The last layer is a linear projection layer, instead of the attention output mechanism in \cite{abu2019mixhop}.
    \item GPR-GNN: The basic setup and grid is the same as that of APPNP. We use their Personalized PageRank weight initialization.
    \item GCNII: number of layers $\in \{2,8,16,32,64\}$, strength of the initial residual connection $\alpha_{\ell} \in \{0.1,0.2,0.5\}$, hyperparameter used to compute the strength of the identity mapping $\lambda \in \{0.5,1.0,1.5\}$.
    \item LINKX: We use take $\mrm{MLP}_\A$ and $\mrm{MLP}_\X$ to be one layer networks, i.e. linear mappings of size $d \times n$ and $d \times D$, respectively. The hidden dimension is taken to be $d \in \{16, 32, 128, 256\}$, and the number of layers of $\mrm{MLP}_f$ is in $\{1, 2, 3\}$. 
\end{itemize}

\subsection{Minibatching hyperparameters}

The setup for minibatching is similar to the setup of full-batch training as above, with some differences that we note here. For all GNNs, we fix the hidden dimension to 128, which is a common hidden dimension used in Cluster-GCN~\cite{chiang2019cluster} and GraphSAINT~\cite{zeng2019graphsaint}. We use concatenation jumping knowledge connections \cite{xu2018representation} for GCNJK. For GCNJK and MixHop, our hyperparameter grid only chooses a number of layers $L \in \{2, 4\}$, along with the hyperparameters for the minibatching methods that we give below. 

\begin{itemize}
    \item MLP, LINK, and LINKX use the same hyperparameter grids as in the full-batch setting, and they are trained with standard minibatching with a batch size of $n/10$. We train with one such batch in each of 500 epochs.
    \item All Cluster-GCN based experiments partition the graph into 200 parts and process a number of parts in $ \{1, 5\}$ at once. We train over all partitions in each of 500 epochs.
    \item All GraphSAINT-Node based experiments have a budget of nodes in $\{5{,}000, 10{,}000\}$. We train over five subgraphs for each of 500 epochs.
    \item All GraphSAINT-RandWalk based experiments use a random walk length of the same size as the number of layers $L$ of the GNN, and use a number of roots in $\{5{,}000, 10{,}000\}$. We train over five subgraphs for each of 500 epochs.
\end{itemize}

These hyperparameter settings are in the same range as those used in the original papers \cite{chiang2019cluster, zeng2019graphsaint} for datasets that have similar node counts to our proposed datasets. For a total number of nodes $n$ in the full graph, note that the MLP, LINK, and LINKX minibatching method processes about $50n$ total nodes across all batches in training. The Cluster-GCN method processes $500n$ nodes. When the number of nodes is 10,000, the GraphSAINT-Node method approximately processes between $601n$ nodes (Penn94) and $8.5n$ nodes (snap-patents). The GraphSAINT-RandWalk method with 10,000 root nodes processes a few times more nodes (usually between two to five times more) than the GraphSAINT-Node method, as more nodes are sampled from the random walks (and we take the number of layers $L$ to be 2 or 4).

\section{Further Experiments}

\subsection{Benefits of Separate Embeddings} \label{sec:linkx_ablation}

\begin{table}[ht]
    \caption{Comparing separately embedding $\A$ and $\X$ (as in LINKX) with direct concatenation. LINKX outperforms the concatenation based model.}
    \centering
    \label{tab:linkx_ablation}
    {\scriptsize
    \begin{tabular}{lllllllll}
    \toprule
	 & Penn94 &  pokec &   arXiv-year & snap-patents & genius & twitch-gamers \\
    \midrule
	MLP & $73.61 \std{0.40}$  & $62.37\std{0.02}$ & $36.70\std{0.21}$ & $31.34\std{0.05}$ & $86.68 \std{0.09}$ & $60.92\std{0.07}$ \\    
     LINK  & $80.79 \std{0.49}$ & $80.54\std{0.03}$  &  $53.97\std{0.18}$ &  $60.39\std{0.07}$ & $73.56\std{0.14}$ & $64.85\std{0.21}$ \\    
     \midrule
	 MLP$([\A; \X])$ & $84.64\std{0.33}$ & $81.74\std{0.15}$  &  $54.15\std{0.20}$ & $59.12\std{0.29}$ & \bestcell $91.61\std{0.05}$ & $64.89\std{0.18}$ \\
	 LINKX &  \bestcell $84.71 \std{0.52}$ & \bestcell $82.04\std{0.07}$  &  \bestcell $56.00 \std{1.34}$ & \bestcell $61.95 \std{0.12}$ &  $90.77 \std{0.27}$ & \bestcell $66.06\std{0.19}$\\
    \bottomrule
    \end{tabular}
    }
\end{table}

Here, we give more evidence to justify separately embedding $\A$ and $\X$ in LINKX. Recall from Section~\ref{sec:linkx_design} that there are computational benefits to separately embedding them.
Table~\ref{tab:linkx_ablation} give results for concatenating $\A$ and $\X$ so that they are jointly embedded as $\mrm{MLP}([\A; \X])$. For this concatenation model, we search over the same hyperparameter grid as that of $\mrm{MLP}_f$ in LINKX. We see that LINKX generally outperforms the concatenation model (besides on the genius dataset). Moreover, LINKX always outperforms both MLP and LINK, while the concatenation model does not do as well as LINK on snap-patents and is within a standard deviation on twitch-gamers.

\subsection{Minibatching experiments} \label{sec:appendix_minibatch_exp}

\begin{table}[ht]
    \centering
    \caption{Comparison of different methods on the arXiv graphs with (homophilous) ogbn-arXiv subject labels and (non-homophilous) arXiv-year publication year labels. Percent relative error of minibatch methods against corresponding full batch methods are in parentheses.}
    {\small
    \begin{tabular}{lllll}
    \toprule
         & ogbn-arXiv Coarse & arXiv-year Coarse \\
         \midrule
         GCNJK-Full & $70.33 $  & $44.18 $ \\
         GCNJK-Cluster & $68.28 \ (2.9\%) $ & $42.88 \  (3.0\%)$\\
         GCNJK-SAINT-Node & $67.69 \ (3.7\%)$ & $41.48 \ (6.1\%)$ \\
         GCNJK-SAINT-RW & $69.57 \ (1.1\%)$ & $43.58 \ (1.4\%)$ \\
         \hdashline
         MixHop-Full &  $72.26 $ & $48.14 $\\
         MixHop-Cluster & $70.16 \ (3.0\%)$ & $46.90 \ (2.8\%)$\\
         MixHop-SAINT-Node & $67.26 \ (7.1\%)$ & $41.68 \ (14.7\%)$\\
         MixHop-SAINT-RW & $71.54 \ (1.0\%)$ & $45.94 \ (5.0\%)$ \\
         \bottomrule
         \toprule
         &  ogbn-arXiv Fine \hspace{5pt}  &  arXiv-year Fine \hspace{10pt}   \\
         \midrule
         GCNJK-Full  & $70.33$  & $44.18$  \\
         GCNJK-Cluster  &  $68.01 \ (3.3\%)$ & $42.49 \ (3.8\%)$  \\
         GCNJK-SAINT-Node   & $66.18 \ (5.9\%)$ & $39.88 \ (9.7\%)$ \\
         GCNJK-SAINT-RW  & $68.96 \ (1.9\%)$ & $42.47 \ (3.9\%)$ \\
         \hdashline
         MixHop-Full  & $72.26$ & $48.14$   \\
         MixHop-Cluster  & $69.37 \ (3.5\%)$ & $46.60 \ (3.2\%)$ \\
         MixHop-SAINT-Node  & $64.71 \ (10.1\%)$ & $39.17 \ (18.7\%)$ \\
         MixHop-SAINT-RW  & $70.48 \ (2.0\%)$ & $43.29 \ (10.1\%)$  \\
         \bottomrule
    \end{tabular}
    }
    \label{tab:arxiv_compare}
\end{table}

\myparagraph{Homophilous vs Non-Homophilous.} To see differences in graph minibatching between homophilous and non-homophilous settings, we setup experiments using the same graph with homophilous labels and then non-homophilous labels; we take the ogbn-arXiv graph and train GNNs with minibatching with on the original ogbn-arXiv subject area labels (that are homophilous, see Table~\ref{tab:homophilic_stats}), and also train GNNs with minibatching on the arXiv-year labels that we defined in this work (that are non-homophilous, see Table~\ref{tab:data_stats}). For the minibatching methods, we use similar hyperparameters as in Appendix~\ref{sec:appendix_minibatch_exp} (Cluster-GCN uses 200 partitions and processes 5 parts at once, GraphSAINT-Node has a 10,000 node budget and GraphSAINT-RandWalk has 10,000 roots), which we call the ``coarse'' setting. In the so-called ``fine'' setting, we use smaller subgraph minibatches (Cluster-GCN uses 750 partitions and processes 5 parts at once, GraphSAINT-Node has a 2,500 node budget and GraphSAINT-RandWalk has 2,500 roots). Both the GCNJK and MixHop architectures are fixed to have two layers and 128 hidden dimensions.

Table~\ref{tab:arxiv_compare} shows the results of these arXiv experiments on two different label sets. We see that performance degradation as measured by percent relative error in test accuracy is more substantial in the non-homophilous arXiv-year experiments for the GraphSAINT methods, while it is mostly comparable for the Cluster-GCN methods.

Graph minibatching could perform poorly in non-homophilous settings for a variety of reasons. 
It has been shown both theoretically and empirically that higher-order information from more than one-hop neighbors are important for classification in certain non-homophilous settings \cite{zhu2020beyond, altenburger2018monophily}. One reason why graph minibatching may have issues here is that it is difficult to preserve higher-order neighborhood structure when minibatching on graphs. For instance, suppose we minibatch a graph with $n$ nodes by taking an induced subgraph on $n/2$ randomly selected nodes. Then for a node $u$ in the subgraph, each one hop path $u \to v$ has probability $1/2$ of being in the subgraph, whereas each two-hop path $u \to w \to v$ has probability $1/4$ of being in the subgraph, since both $w$ and $v$ must be sampled.

Finally, even if the percent relative error of minibatching methods is similar in homophilous vs. non-homophilous graphs, the performance degradation would generally be more detrimental in non-homophilous graphs. This is because the gap between GNNs and methods that do not use the graph topology like MLPs are lower in many non-homophilous settings. Since MLPs do not face much performance degradation when trained with simple i.i.d. node minibatching, the gap between GNNs and MLPs is even lower in minibatched settings. Indeed, we see that MLPs perform on par with or outperform many GNNs in the minibatched setting in Table~\ref{tab:mini_results}.

\begin{table}[ht]
    \vspace{-5pt}
    \centering
	\caption{Minibatching results when using GCN as a base model. The setup is the same as in Section~\ref{sec:mini_results}. LINKX results are provided as reference.}
	\label{tab:gcn_results}
    {\tiny
    \begin{tabular}{lllllllll}
    \toprule
	 & Penn94 & pokec $\dagger$ & arXiv-year & snap-patents $\dagger$ & genius  & twitch-gamers $\dagger$ & wiki  $\dagger$ \\
    \midrule
	GCN-Cluster &  70.24$\pm$0.24 & 67.33$\pm$0.21 & 45.80$\pm$0.22 & 38.53$\pm$0.08 & 82.12$\pm$0.40 & 60.71$\pm$0.15 & (T) \\
	GCN-SAINT-Node & 69.41$\pm$0.55 & 59.91$\pm$0.08 & 44.92$\pm$0.23 & 27.16$\pm$0.16 & 80.95$\pm$0.09 & 59.03$\pm$0.19 & 42.59$\pm$0.09 \\
	GCN-SAINT-RW & 69.79$\pm$0.31 & 62.50$\pm$0.18 & 48.07$\pm$0.21 & 32.75$\pm$0.08 & 80.98$\pm$0.25 & 59.52$\pm$0.08 & 44.22$\pm$0.18 \\
	 \midrule
	LINKX Minibatch & \bestcell 84.50$\pm$0.65 & \bestcell 81.27$\pm$0.38 & \bestcell 53.74$\pm$0.27 & \bestcell 60.27$\pm$0.29 & \bestcell 85.81$\pm$0.10 & \bestcell 65.84$\pm$0.19 & \bestcell 59.80$\pm$0.41 \\
	 \bottomrule
    \end{tabular}
    }
\end{table}

\myparagraph{GCN results.} In Section~\ref{sec:mini_results}, we use GCNJK and MixHop as base models for minibatching evaluation. For completeness, we also include results for using GCN as a base model in Table~\ref{sec:mini_results}. The results are qualitatively the same, though GCN generally performs worse than the other two GNNs.

\subsection{Experiments on Prior Datasets}\label{sec:prior_datasets}

Although the \citet{pei2019geom} datasets suffer issues with evaluation of graph learning methods as discussed in the main paper, we test LINKX on these datasets for comparison. We use the 10 fixed splits of \cite{pei2019geom}, directly reporting results from papers where they also use the official splits, and re-running methods when this was not the case. In particular, GPR-GNN used larger 60-20-20 random splits, while GCNII did not evaluate on Actor and Squirrel. Thus, we re-run GPR-GNN and GCNII using the hyperparameters of Section \ref{sec:hparams}.

\begin{table}[ht]
    \vspace{-5pt}
    \centering
	\caption{Comparison of non-homophilous methods on datasets of  \citet{pei2019geom} (collected by \cite{rozemberczki2019multi, tang2009social, mitchell1997web}). $\dagger$ represents re-run result, if unofficial dataset splits were used in their method paper, or  the paper did not evaluate their method on the specific dataset. Best results up to a standard deviation are highlighted. Geom-GCN and GCNII did not report standard deviation information. } 
	\label{tab:gcn_results}
    {\tiny
    \begin{tabular}{lllllll}
    \toprule
	 & Texas & Wisconsin & Actor & Squirrel & Chameleon & Cornell  \\
	 \# Nodes & 183 & 251 & 7,600 & 5,201 & 2,277 & 183  \\
    \midrule
	 H$_2$GCN-1 & \bestcell $84.86 \std{6.77}$ & \bestcell $ 86.67 \std{4.69} $ & \bestcell $ 35.86 \std{1.03} $ & $ 36.42 \std{1.89} $ & $ 57.11 \std{1.58} $ & \bestcell $ 82.16 \std{4.80} $ \\
	 H$_2$GCN-2 & $ \bestcell 82.16 \std{5.28}$ & \bestcell $ 85.88 \std{4.22} $ & \bestcell $ 35.62 \std{1.30} $ & $ 37.90 \std{2.02} $ & $ 59.39 \std{1.98} $ & \bestcell $82.16 \std{6.00} $  \\
	 MixHop & $ 77.84 \std{7.73} $ & $ 75.88 \std{4.90} $ & $ 32.22 \std{2.34} $ & $ 43.80 \std{1.48} $ & $ 60.50 \std{2.53} $ & $ 73.51 \std{6.34} $\\ 
	 GPR-GNN & $76.22 \std{10.19} \dagger$  & $75.69 \std{6.59} \dagger $ & $33.12 \std{0.57} \dagger $ & $54.35 \std{0.87} \dagger $ & $62.85 \std{2.90} \dagger$ & $68.65 \std{9.86} \dagger $ \\ 
	 GCNII & $77.84$ & $81.57$ & $34.36 \std{0.77} \dagger $  & $56.63 \std{1.17} \dagger $ &  $62.481$ & $76.46$ \\
	 Geom-GCN-I & $57.58$ & $58.24$ & $29.09$ & $33.32$ & $60.31$ & $56.76$\\ 
	 Geom-GCN-P & $67.57$ & $64.12$ & $31.63$ & $38.14$ & $60.90$ & $60.81$ \\
	 Geom-GCN-S & $59.73$ & $56.67$ & $30.30$ & $36.24$ & $59.96$ & $55.68$\\ 
	 \midrule
	LINKX & $74.60 \std{8.37}$ & $75.49 \std{5.72}$ & \bestcell $36.10 \std{1.55}$ & \bestcell $61.81 \std{1.80}$ & \bestcell $68.42 \std{1.38}$ &  $77.84 \std{5.81}$ \\
	 \bottomrule
    \end{tabular}
    }
\end{table}

For these experiments, we used the following hyper-parameter grid for LINKX: MLP$_\A \in \{1, 2\}$, MLP$_\X \in \{1,2 \}$, hidden channels $\in \{64, 128, 256, 512\}$, number of MLP$_f$ layers $\in \{1,2,3,4\}$ learning rate in $\{0.05, 0.01, 0.002\}$ and dropout  $\in \{0.0, 0.5\}$. 

LINKX performs well on datasets with thousands of nodes, despite being primarily a scalable method, only falling short on the tiniest of datasets. Overall, evaluation on these datasets is very noisy, highly dependent on hyperparameter grid selection, and few conclusions can be drawn on the relative performance of different methods in non-homophily more broadly. In particular, note that methods that do well on these datasets do not necessarily do well on our large, non-homophilous datasets. For example, while MixHop is the best non-homophilous GNN on the datasets we present in this paper, it does poorly on the datasets of \citet{pei2019geom}; in contrast, H$_2$GCN achieves excellent performance here, but its design choices make it run out of memory on even medium-sized datasets.

\subsection{Experiments on Homophilous datasets} 

Though LINKX was not designed to do well on small homophilous datasets, we also report LINKX on the homophilous datasets of Cora, Citeseer and Pubmed \cite{yang2016revisiting}, with the standard 48/32/20 training, validation, and test proportions. We use the same hyper-parameter grid for LINKX as in Section \ref{sec:prior_datasets}.

\begin{table}[ht]
    \vspace{-5pt}
    \centering
	\caption{Comparison of on homophilous datasets. Results other than LINKX reported from \cite{zhu2020beyond}. Best three results per dataset are highlighted.  } 
\label{tab:homophilous_results}
    {\small
    \begin{tabular}{lllll}
    \toprule
	 & CiteSeer & PubMed & Cora\\
	 \# Nodes & 3327 & 19717 & 2708\\
    \midrule
	 H$_2$GCN-1 & \bestcell $77.07 \std{1.64} $ & \bestcell $89.40 \std{0.34} $ &  $86.92 \std{1.37}$ \\
	 H$_2$GCN-2 &  \bestcell $76.88 \std{1.77}$ & \bestcell $89.59 \std{0.33}$ & \bestcell $87.81 \std{1.35} $\\
	 MixHop &  $76.26 \std{1.33}$ &  $85.31 \std{0.61}$ & \bestcell $87.61 \std{0.85} $\\ 
	 GCN & \bestcell $76.68 \std{1.64} $ & $87.38 \std{0.66} $ & \bestcell $87.28 \std{1.26}$ \\ 
	 GAT &  $75.46 \std{1.72}$ & $84.68 \std{0.44} $ & $82.68 \std{1.80} $\\

	 \midrule
	LINKX & $73.19 \std{0.99}$ & \bestcell $87.86 \std{0.77} $ & $84.64 \std{1.13}$  \\
	 \bottomrule
    \end{tabular}
    }
\end{table}

\section{Further dataset details}\label{sec:further_dataset}

\subsection{Licenses}

In this section, we note the licenses of the datasets we collect: 

\begin{itemize}
    \item wiki: Wikipedia is licensed under Creative Commons Attribution-ShareAlike 3.0 Unported License and the GNU Free Documentation License, unversioned, with no invariant sections, front-cover texts, or back-cover texts.
    \item Penn94: The Facebook 100 datasets are available online (https://archive.org/details/oxford-2005-facebook-matrix), and to the best of our knowledge were not released with a license, though the corresponding paper has an arXiv non-exclusive license to distribute. Upon release, there were privacy concerns \cite{zimmer11Facebook}, as the data release may not have respected certain privacy settings, and the initial release of the data included unique identifiers. We use a version that does not include the unique identifiers, but of course may still be subject to deanonymization attacks. While we do not add any sensitive information to the dataset, we acknowledge that deanonymization is possible, though our work does not directly contribute to deanonymization risks.
    \item Pokec: We retrieved the data from SNAP; the original source of the data is \cite{pokec}. To the best of our knowledge, the data was not released with a license. This is a social network, so there may be privacy concerns with the user data. Still, we include it as a suitable benchmark that has been previously used, as the dataset is large and has interesting feature information. We only provide numerical values and do not provide any of the raw text in the dataset.
    \item arxiv-year: The dataset is licensed under ODC-BY. We originally downloaded the data from the Open Graph Benchmark \cite{hu2020open}, and the data is a subset of the Microsoft Academic Graph \cite{Wang2020Microsoft}
    \item snap-patents: The data was originally publically released by NBER in 2001 in a working paper by \cite{NBERw8498}. To the best of our knowledge, the dataset was not released with a license. 
    \item genius: The dataset is open-sourced by the authors with no attached license. It was originally introduced in a conference paper \cite{lim2021expertise}. While this is a social network and thus may face privacy concerns, we believe that the task of predicting undesired nodes can be very beneficial to society, so we benchmark methods on it. We only provide numerical values, and omit raw text information that has been previously released in the dataset.
    \item twitch-gamer: The dataset is open-sourced by the authors in a repository with the MIT license. It was originally introduced in an academic paper \cite{mcpherson2001birds}. This is also a social network that may face privacy concerns, but the prediction task of detecting situationally undesired nodes may be socially beneficial, so we benchmark methods on it. There is no raw text in the dataset.
\end{itemize}

\subsection{Dataset Properties}\label{sec:dataset_properties}

\myparagraph{Penn94} \cite{traud2012social} is a friendship network from the Facebook 100 networks of university students from 2005, where nodes represent students. Each node is labeled with the reported gender of the user. The node features are major, second major/minor, dorm/house, year, and high school. 

\myparagraph{Pokec} \cite{snapnets} is the friendship graph of a Slovak online social network, where nodes are users and edges are directed friendship relations. Nodes are labeled with reported gender. We derive node features from profile information, such as geographical region, registration time, and age.

\myparagraph{arXiv-year} \cite{hu2020open} is the ogbn-arXiv network with different labels. Our contribution is to set the class labels to be  the year that the paper is posted, instead of paper subject area. The nodes are arXiv papers, and directed edges connect a paper to other papers that it cites. The node features are averaged word2vec token features of both the title and abstract of the paper. The five classes are chosen by partitioning the posting dates so that class ratios are approximately balanced.
    
\myparagraph{snap-patents} \cite{leskovec2005graphs, snapnets} is a dataset of utility patents in the US. Each node is a patent, and edges connect patents that cite each other. Node features are derived from patent metadata. 
Our contribution is to set the task to predict the time at which a patent was granted, resulting in five classes.

\myparagraph{genius} \cite{lim2021expertise} is a subset of the social network on genius.com --- a site for crowdsourced annotations of song lyrics. Nodes are users, and edges connect users that follow each other on the site. This social network has not been used for node classification in the literature, so we define the task of predicting certain marks on the accounts. About 20\% of users in the dataset are marked ``gone'' on the site, which appears to often include spam users. Thus, we predict whether nodes are marked. The node features are user usage attributes like the Genius assigned expertise score, counts of contributions, and roles held by the user.

\myparagraph{twitch-gamers} \cite{rozemberczki2021twitch} is a connected undirected graph of relationships between accounts on the streaming platform Twitch. Each node is a Twitch account, and edges exist between accounts that are mutual followers. The node features include number of views, creation and update dates, language, life time, and whether the account is dead. The binary classification task is to predict whether the channel has explicit content.

\myparagraph{wiki} is a dataset of Wikipedia articles, where nodes represent pages and edges represent links between them. We collect this new dataset, with a process that we describe further in Appendix~\ref{sec:wiki}. Node features are constructed using averaged title and abstract GloVe embeddings \cite{pennington-etal-2014-glove}. Labels represent total page views over 60 days, which are partitioned into quintiles to make five classes.

\subsection{Wiki collection details}\label{sec:wiki}

Here, we detail the process of crawling and cleaning the wiki dataset. We generated the graph using a breadth-first search, where we started from the Wikipedia page on Hilbert Spaces, then proceeded to visit all its neighbors, and so forth. For each Wikipedia page visited, we used the MediaWiki web service API to get a list of pages linked to by the given page — this forms the directed edges of the graph. In each API query, we also received the number of page views per day in the past 60 days. We crawled these articles throughout April and May of 2021. To convert this into discrete labels, we used an even quantile function, which set view boundaries to make the number of nodes in each class as even as possible. The output of this function formed the labels of the graph. 

For the node features, we formed 300 dimensional Wikipedia Glove vectors \cite{pennington-etal-2014-glove} for each word in the title and abstract, then averaged the word vectors in the title and abstract, thus resulting in 600 dimensional feature vectors for each node. For words not found in the Glove dictionary, we used the zero vector. This procedure was modeled on the construction of the ogbn-arxiv dataset by \cite{hu2020open}. In particular, we avoided a one-hot vector because of the vast dimensionality that would be required. Finally, to clean the dataset, we pruned edges if either of the nodes that it spanned where not in our subset of wikipedia articles. 
Ultimately, we decided to stop collection at approximately one third of English wikipedia due to limitations of computational resources and time. Already, it is not possible to run full batch experiments on the wiki dataset, while requiring over 80GB of CPU RAM for some minibatching techniques. As such, we believe that the full wikipedia dataset would have been too large and unwieldy to use as an evaluation dataset for many research labs, and our dataset is at a good size that may hopefully provide utility to many researchers.

\end{document}